\documentclass{article}

\usepackage{microtype}
\usepackage{graphicx}
\usepackage{subfigure}
\usepackage{booktabs} % for professional tables

\usepackage{hyperref}

\usepackage[accepted]{icml2025}

\usepackage{amsmath}
\usepackage{amssymb}
\usepackage{mathtools}
\usepackage{amsthm}

\usepackage[capitalize,noabbrev]{cleveref}

\theoremstyle{plain}

\theoremstyle{definition}

\theoremstyle{remark}

\usepackage[textsize=tiny]{todonotes}

\definecolor{MILAPurple}{RGB}{109,0,109}  
\definecolor{airforceblue}{rgb}{0.36, 0.54, 0.66}
\definecolor{amaranth}{rgb}{0.9, 0.17, 0.31}
\definecolor{darkraspberry}{rgb}{0.53, 0.15, 0.34}
\definecolor{cadmiumred}{rgb}{0.89, 0.0, 0.13}

\usepackage{lipsum}
\newcommand{\noise}{\ensuremath{\varepsilon}}
\newcommand{\tmap}{\ensuremath{\mathcal{T}}}  
\usepackage{fontawesome5}
\usepackage{tikz}
\usepackage{tikz-3dplot} % Add the tikz-3dplot package
\usetikzlibrary{mindmap, calc, 3d,decorations.text,shapes.arrows,positioning,fit,backgrounds, matrix, decorations.pathmorphing,fadings,patterns,shadows,shadows.blur}
\icmltitlerunning{The Butterfly Effect}

\begin{document}

\twocolumn[
\icmltitle{The Butterfly Effect: Neural Network Training Trajectories Are \\ Highly Sensitive to Initial Conditions}



\icmlsetsymbol{equal}{*}
\icmlsetsymbol{intern}{\triangleright}

\begin{icmlauthorlist}
\icmlauthor{Devin Kwok}{equal,mcgill,mila}
\icmlauthor{Gül Sena Altıntaş}{equal,uoft,vector}%\thanks{Work primarily conducted while visiting Mila - Quebec AI Institute.}
\icmlauthor{Colin Raffel}{uoft,vector}
\icmlauthor{David Rolnick}{mcgill,mila}
\end{icmlauthorlist}

\icmlaffiliation{mcgill}{School of Computer Science, McGill University, Montreal, Canada}
\icmlaffiliation{mila}{Mila -- Quebec AI Institute, Montreal, Canada}
\icmlaffiliation{uoft}{University of Toronto, Vector Canada}
\icmlaffiliation{vector}{Vector Institute, Toronto, Canada}

\icmlaffiliation{mcgill}{McGill University}
\icmlaffiliation{mila}{Mila -- Quebec AI Institute}
\icmlaffiliation{uoft}{University of Toronto}
\icmlaffiliation{vector}{Vector Institute}

\icmlcorrespondingauthor{Gül Sena Altıntaş}{gsaltintas@cs.toronto.edu}
\icmlcorrespondingauthor{Devin Kwok}{devin.kwok@mail.mcgill.ca}

\icmlkeywords{Machine Learning, ICML}

\vskip 0.3in
]

\printAffiliationsAndNotice{\icmlEqualContribution} % otherwise use the standard text.

%%% re-written slightly for camera-ready
\begin{abstract}
Neural network training is inherently sensitive to initialization and the randomness induced by stochastic gradient descent.
However, it is unclear to what extent such effects lead to meaningfully different networks, either in terms of the models' weights or the underlying functions that were learned.
In this work, we show that during the initial ``chaotic'' phase of training, even extremely small perturbations reliably causes otherwise identical training trajectories to diverge---an effect that diminishes rapidly over training time.
We quantify this divergence through (i) $L^2$ distance between parameters, (ii) the loss barrier when interpolating between networks, (iii) $L^2$ and barrier between parameters after permutation alignment, and (iv) representational similarity between intermediate activations; revealing how perturbations across different hyperparameter or fine-tuning settings drive training trajectories toward distinct loss minima.
Our findings provide insights into neural network training stability, with practical implications for fine-tuning, model merging, and diversity of model ensembles.
\footnote{Our code is available at \url{https://github.com/gsaltintas/lmc}}
\end{abstract}

%%%%%%%%%%%%%%%%%%%%%%%%%%%%%%%%%%%%%%%%%%%%%%%%%%%%
%%%%%%%%%%%%%%%%% Butterfly Figure %%%%%%%%%%%%%%%%%
%%%%%%%%%%%%%%%%%%%%%%%%%%%%%%%%%%%%%%%%%%%%%%%%%%%%
% Define relative measurements
\newcommand{\vertspace}{2.25}    % Vertical spacing unit
\newcommand{\horizspace}{1}   % Horizontal spacing unit
\newcommand{\nodesize}{4pt}   % Node size
\newcommand{\nodesep}{2pt}    % Node inner separation

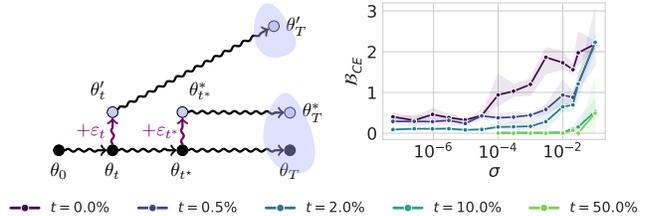
\begin{figure}
\vskip0.1in
    \centering
    \resizebox{!}{0.3\linewidth}{
    
    % \resizebox{0.4\linewidth}{!}{
    \begin{tikzpicture}[every node/.style={font={\normalsize}},]
    \tikzset{every node/.style={
        draw, 
        circle, 
        fill=blue!20, 
        inner sep=\nodesep,
        minimum size=\nodesize
    }}
    
    % Define coordinates relative to base units
    \coordinate (top) at (0, \vertspace);
    \coordinate (middle) at (0, 3/5*\vertspace);
    \coordinate (bottom) at (0.5*\horizspace, 0);
    
    % Place nodes using relative coordinates
    \node[fill=black] (Mpre) at ([xshift=0 cm]bottom) {};
    \node[fill=black] (MA0) at ([xshift=\horizspace cm]bottom) {};
    \node[fill=black] (MB0) at ([xshift=1.3*\horizspace cm]MA0) {};
      \draw[->, very thick, decorate, decoration={snake, amplitude=.25mm, segment length=2mm, post length=1mm}] (Mpre) -- (MA0) node[midway, right, fill=none, draw=none] {}; % {spawn for $k$ steps};
    \node[] (MperturbedA) at ([yshift=0.7\horizspace cm]MA0) {};
    \node[] (MperturbedB) at ([yshift=0.7\horizspace cm]MB0) {};
    \node[] (MAt) at ([xshift=3*\horizspace cm, yshift=1.6*\horizspace cm] MperturbedA) {};
    \node[fill=black] (Mt) at ([xshift=2*\horizspace cm] MB0) {};
    \node[] (MBt) at ([yshift=0.7*\horizspace cm] Mt) {};

      \draw[->, very thick, decorate, decoration={snake, amplitude=.25mm, segment length=2mm, post length=1mm}, MILAPurple] (MA0) -- (MperturbedA) node[midway, shift={(-4mm,0mm)}, fill=none, draw=none, MILAPurple] {$+\noise_{t}$};
      \draw[->, very thick, decorate, decoration={snake, amplitude=.25mm, segment length=2mm, post length=1mm}, MILAPurple] (MB0) -- (MperturbedB) node[midway, shift={(-4mm,0mm)}, fill=none, draw=none, MILAPurple] {$+\noise_{t^\ast}$};
    
      \draw[->, very thick, decorate, decoration={snake, amplitude=.25mm, segment length=2mm, post length=1mm}] (MperturbedA) -- (MAt);
      \draw[->, very thick, decorate, decoration={snake, amplitude=.25mm, segment length=2mm, post length=1mm}] (MA0) -- (MB0);
      \draw[->, very thick, decorate, decoration={snake, amplitude=.25mm, segment length=2mm, post length=1mm}] (MperturbedB) -- (MBt);
      \draw[->, very thick, decorate, decoration={snake, amplitude=.25mm, segment length=2mm, post length=1mm}] (MB0) -- (Mt);

    %% add labels
    \tikzset{every node/.style={draw=none, minimum size=2pt, font={\normalsize}}}

      \node[draw=none, below] at (Mpre.south) {$\theta_0$};
      \node[draw=none, below] at (MA0.south) {$\theta_t$};
      \node[draw=none, below] at (MB0.south) {$\theta_{t^\star}$};
      \node[draw=none, above left] at (MperturbedA.north) {$\theta_{t}^\prime$};
      \node[draw=none, above right] at (MperturbedB.north) {$\theta^\ast_{t^\ast}$};
      \node[draw=none, right] at (MAt.east) {$\theta^\prime_T$};
      \node[draw=none, right] at (MBt.east) {$\theta^\ast_T$};
      \node[draw=none, below] at (Mt.south) {$\theta_T$};
      
    %%% generate basins
     % Automatically generate an irregular fuzzy shaded area around the node
    % Draw fuzzy shaded region around MAt
    
% Draw fuzzy shaded region around MAt
\begin{scope}
    \fill[blue!30, opacity=0.4] 
        plot[smooth cycle, tension=1]
        coordinates {
            ([xshift=-0.2cm, yshift=0.4cm] MAt)
            ([xshift=0.3cm, yshift=0cm] MAt)
            ([xshift=0.3cm, yshift=-0.5cm] MAt)
            ([xshift=-0.3cm, yshift=-0.4cm] MAt)
        };
        
    \fill[blue!30, opacity=0.4] 
        plot[smooth cycle, tension=1]
        coordinates {
            ([xshift=-0.3cm, yshift=0.2cm] MBt)
            ([xshift=0.2cm, yshift=0.1cm] MBt)
            ([xshift=0.4cm, yshift=-0.3cm] Mt)
            % ([xshift=0.1cm, yshift=-0.2cm] MBt)
            ([xshift=-0.4cm, yshift=-0.1cm] Mt)
        };
\end{scope}
\end{tikzpicture}
}
    \includegraphics[height=0.3\linewidth]{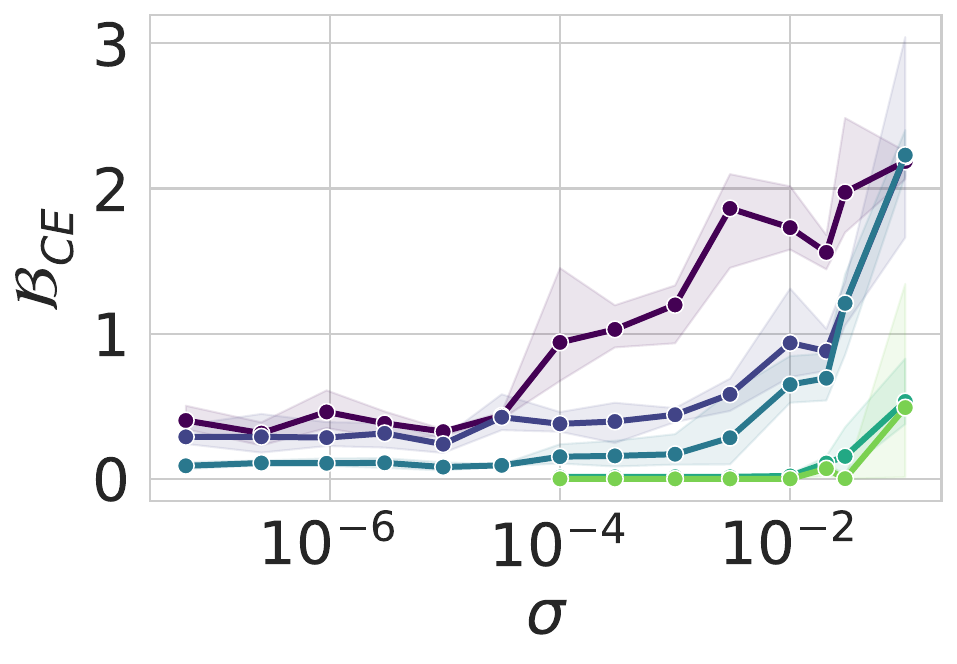}
% \vskip-0.05in
\centerline{
    \includegraphics[width=1.05\linewidth]{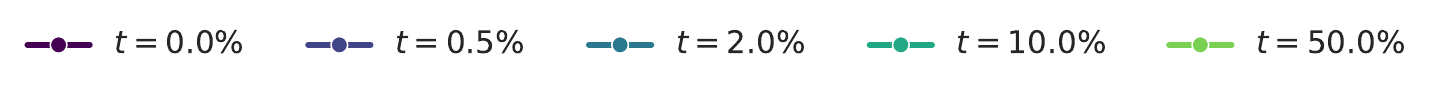}
}
\vskip-0.1in
    \caption{
    \label{fig:barrier-perturb}
    \textbf{Left}: illustration of the ``butterfly effect'': a network $\theta_0$ is trained until time $t$ and perturbed by $\noise$ early ($\noise_t$) or later ($\noise_{t^\ast}$) in training. Both copies are trained deterministically until $T$ and their divergence is measured (purple loss basins).
    \textbf{Right}: barriers (training cross-entropy loss) at $T$ versus perturbation magnitude ($\sigma = 1$ is the network's initialization scale). A perturbation of \emph{as little as one weight} (leftmost points) reliably causes divergence when applied early, but not when applied at later $t$ (colors).
    }
\vskip -0.2in
\end{figure}

\section{Introduction} \label{sec:intro}

Neural network training is known to be unstable in the sense that noise can disrupt convergence to a particular minimum \citep{frankle2020linear, wu2018sgd}.
This is true even when considering solutions that perform equally well, since symmetries and connected minima in the loss landscape give rise to many different ways for a neural network to parameterize identical or similar functions.

%%% re-written for camera-ready

Although training instability affects convergence in general \citep{iyer2023maximal, jastrzebski2020break}, it also prevents different runs of the same network from consistently reaching one particular solution, which has practical implications for model merging~\citep{singh2020model,ainsworth2022git} and ensembling.
Prior work has categorized training into chaotic (early) and stable (late) phases \citep{fort2020deep, frankle2020linear}, but it is not clear if instability is more a product of noise (e.g. batch noise, data augmentations, GPU indeterminacy), a network's current state (e.g. random vs. pre-trained), or the training procedure itself (e.g. optimizer and hyperparameter selection).
A thorough understanding of instability should disentangle these factors, since their influence may vary over training time and between different settings.
Furthermore, depending on the task, not all of these factors can be controlled--- when fine-tuning a pre-trained model one can vary hyperparameters but not the initial weights, for instance, while the opposite may be true when pre-training the same model.

These limitations motivate us to study training stability absent the effects of noise.
Drawing from the dynamical systems perspective, we consider how much a \emph{deterministic} training map diverges when a controlled perturbation is applied to its initial conditions (i.e.~the starting weights of a network).
By selecting initial weights from models trained or fine-tuned for varying durations, we build up a picture of where in the loss landscape training trajectories tend to diverge, and how sensitive trajectories are to perturbations in these regions.
Crucially, our approach can quantify stability more precisely and for a wider range of models---from randomly initialized to pre-trained---than was possible in prior works, which only measured instability to training noise.
Our contributions are as follows:

\begin{enumerate}
\vspace{-0.15in}
    \item We show that a tiny perturbation of \emph{as little as a single weight} early in training causes two otherwise identically initialized and trained networks to diverge---the \emph{butterfly effect} (\cref{fig:butterfly-time}).
    \item Conversely, even networks that are stable to training noise diverge under larger perturbations. This points towards the possibility of using perturbations during training or fine-tuning to increase model diversity and ensembling performance (\cref{fig:butterfly-cka}).
    \item We find that stability improves under settings including wider/shallower networks and increased learning rate warm-up (among others), and these settings can be combined to further decrease, but not eliminate, instability near initialization (\cref{fig:butterfly-warmup-arch-combo}).
    \item While pre-trained networks are orders of magnitude more stable than randomly initialized networks, stability varies greatly between tasks and remarkably, more pre-training of language models can actually \emph{reduce} stability in some cases (\cref{fig:fine-tuning}).
    \item Contrary to a dynamical systems perspective, $L^2$ and barriers~\citep{frankle2018lottery} do not grow exponentially over training (\cref{fig:butterfly-l2-lyapunov}), and although $L^2$ and barriers are strongly correlated in some cases, this is not true generally (\cref{fig:l2-barriers}). 
\vspace{-0.1in}
\end{enumerate}

\section{Related Work}\label{sec:related_work}

\emph{Stability and optimization.} 
Many works have studied the stability of optimization by asking if neural networks converge to well-generalizing minima \citep{cohenGradientDescentNeural2022, jastrzebski2020break, wu2018sgd} or converge at all \citep{iyer2023maximal, jacotNeuralTangentKernel2020, sohldickstein2024boundaryneuralnetworktrainability}.
However, these works do not consider stability relative to any particular training trajectory.

We focus on the question of whether training from an initial point tends to follow the same trajectory to a specific ``loss basin'', i.e.\ a linearly connected, low-loss region of the loss landscape.
Although narrower in scope, this question is highly relevant for practical contexts such as when using pre-trained models or conducting fine-tuning.
In order to merge or ensemble models in these contexts, one may not merely want to find a good solution, but may, for example, want to converge towards or away from a particular pre-existing solution in order to improve merge-ability or model diversity, respectively.

\emph{Dynamical system stability.} Neural network training has been studied as a stochastic process \citep{smith2017bayesian, tehConsistencyFluctuationsStochastic2015,redman2024identifying} and as a dynamical system \citep{wu2018sgd, jastrzebski2020break, cohenGradientDescentNeural2022}.
We take the latter view and analyze how small perturbations to a network's initial weights evolve over training.
This has two advantages:
first, we can differentiate between instability due to noise vs. instability inherent to training itself (in the same way that deterministic dynamical systems exhibit chaos); and second, we can model simpler and larger (exponential) instabilities, which if present, should dominate over stochastic effects and make the stochastic perspective unnecessary \citep{wu2018sgd}.

Prior works \citep{wu2018sgd, jastrzebski2020break, cohenGradientDescentNeural2022} have considered dynamical systems stability in terms of whether the $L^2$ distance between a network's weights and a fixed minimum will grow over time.
Our work differs in that we evaluate stability relative to a moving trajectory, and we also want to know if networks are diverging in function (not only in weights).
To measure functional divergence, we use barriers \citep{frankle2020linear}, barriers accounting for neural network permutation symmetries \citep{entezari2021role, ainsworth2022git}, and representational similarity via Angular CKA \citep{williams2021generalized}. Although these quantities are not amenable to approximation by linear dynamical systems, they better capture the practical differences between networks.

\emph{Linear mode connectivity.}
Barriers \citep{frankle2020linear, neyshabur2020being} are the maximum increase in loss on a linear path between two networks.
Networks with barriers below some noise threshold are said to exhibit \emph{linear mode connectivity} (LMC), which among other useful properties is a necessary condition for the loss landscape to be locally convex \citep[Definition 3.1]{neyshabur2020being}. 
Thus, non-zero barriers indicate that two networks belong to different convex loss basins \citep{goodfellow2014qualitatively, huang2017snapshot, yunis2022on}.\footnote{Following \citet{entezari2021role, neyshabur2020being}, we make the extra assumptions needed to assume the converse, i.e. that networks with zero barrier are in the same loss basin.}

Note that we do not consider the more general notion of (non-linear) mode connectivity for several reasons.
First, despite being non-convex, neural network training is often understood in terms of convex optimization.
In convex regions of the loss landscape however, convex optimization behaves exactly as expected.
Similarly, the dynamical systems perspective often takes a linear or quadratic approximation of the loss landscape, which again holds precisely in convex regions.
Practically, merging models by weight averaging requires linearly connected networks, which is guaranteed if the networks are from the same convex region.
Finally, both theoretical \citet{simsek2021geometry, lin2024exploring} and empirical \citep{draxlerEssentiallyNoBarriers2018, garipovLossSurfacesMode2018, sonthalia2024deepneuralnetworksolutions} works have suggested that most or all minima may be trivially connected by non-linear paths.

\emph{Spawning experiments.} \citet{frankle2020linear, frankle2020the} and \citet{fort2020deep} consider if training is stable to random batch order and data augmentations (training noise) by spawning pairs of networks from the same parent, and measuring barriers between them after training.
They find that training becomes stable to training noise after an early period of instability.
\citet{altintas2023disentangling} shows that reducing early training variability by lowering learning rates, increasing batch sizes, and adding learning rate warm-up further increase stability to training noise, and \citet{singhLandscapingLinearMode} relates stability in barriers to the loss landscape geometry.
These findings align with observations that after an initial chaotic phase, SGD training trajectories can be approximated by a linear kernel \citep{fort2020deep}.

We adopt the same parent-child spawning experiments \citep{frankle2020linear,frankle2020the,fort2020deep}, but we eliminate all training noise so as to isolate the effects of perturbation to specific training times.
This not only lets us precisely identify when and how much perturbation causes instability, but also allows us to do so on both randomly initialized (chaotic) and pre-trained (stable) networks.

\emph{Model averaging.}
Weight averaging can merge zero-barrier networks from the same training trajectory \citep{izmailov2018averaging}, different runs \citep{utans1996weight,wortsman2021learning}, or different tasks \cite{mirzadehLinearModeConnectivity2020} to improve inference speed and even performance.
Weight averaging is the basis for more sophisticated model merging strategies \citep{ilharcoTaskArithmetic2022}, including those using permutations to align diverging \citep{wang2019federated} or unrelated \citep{singh2020model} models.
Our work indicates the conditions in which models are stable with respect to barriers, and thus amenable to weight averaging.

\emph{Permutation symmetries.} 
Recent works have shown that independently initialized networks can converge to linearly connected basins after accounting for permutation symmetries, or different ways to order the neurons in a network \citep{entezari2021role, ainsworth2022git, benzing2022random}.
The permutations aligning two networks are most unstable early in training \citep{sharma2024simultaneous}, suggesting a connection with training instability.
While our work compares identical rather than randomly initialized networks, we apply weight and activation matching algorithms in the same manner as \citet{ainsworth2022git} to determine (1) if training instability causes permutations between networks, and (2) if undoing these permutations returns diverging trajectories to the same loss basin.

\emph{Representational similarity.}
Representational similarity compares the intermediate (hidden) outputs of two networks, and can detect functional differences even when two networks have identical performance.
Although many methods exist, we use Angular CKA \citep{williams2021generalized} which we explain and justify in detail in \cref{ap:sec:cka}.
In general, representational similarity methods are invariant to symmetries of a network's outputs, but not of its weights.
Thus, dissimilar representations indicate greater diversity between networks (which can improve ensembling performance), but similar representations do not guarantee that two networks are in the same loss basin (due to weight symmetries), and thus is not sufficient for weight averaging to succeed.

\emph{Fine-tuning stability.}
Pre-trained models are generally stable to training noise and converge to the same basin during fine-tuning \citep{neyshabur2020being}.
However, more recent work has found that this is not always true, as \citet{juneja2023linear} discovered that training noise causes fine-tuning of language models to converge to different basins.
While detrimental to model merging, this kind of instability can improve model diversity and thus ensemble performance \citep{lubanaMechanisticModeConnectivity2023, sadrtdinov2024stay}, even if a single basin has equivalent diversity \citep{lionHowGoodSingle2023}.

Our method enables us to find the threshold between stability and instability for any fine-tuning setting---even ones stable to training noise---since we can increase our perturbations to any scale necessary for inducing instability in a given network.
Using our method, we identify differences in stability between language and vision models, specifically studying the fine-tuning dynamics of ResNets \citep{He_2016_CVPR}, ViT \citep{dosovitskiy2021vit}, BERT \citep{devlinBERTPretrainingDeep2019}, and OLMo \citep{groeneveldOLMoAcceleratingScience2024}.

\section{Methods}\label{sec:background}

In this section, we define our notion of training stability, describe the framing for our perturbation experiments, and finally, define the functional dissimilarity scores and other quantities we evaluate (barriers, barriers modulo permutation, and $L^2$ divergence over training time).

\subsection{Training Instability}

Consider training as the iterative application of a stochastic training map $\tmap: \Theta \to \Theta$ to the initial parameters $\theta_0 \in \Theta$ of a neural network, so that the network's parameters after training are
\begin{align}
\theta_T = \tmap^T(\theta_0; \xi) &= \underbrace{\tmap \circ \tmap \circ ... \circ \tmap}_{T \text{ times}}(\theta_0; \xi_1, ..., \xi_T),
\label{eq:training-map}
\end{align}
where $\xi = (\xi_1, \dots, \xi_T)$ accounts for all of the stochastic factors influencing training, such as batch sampling, data augmentation, and hardware-induced non-determinism.
Unless specified, we treat $\theta$ as a vector concatenation of a network's parameters, and when writing $\tmap$ we omit the training or test data if it can be inferred from context.

We are interested in the degree to which training is stable, in the sense that a small perturbation to a network's initial weights does not significantly change the network after training.
To describe stability on a continuum, we evaluate how far $\theta_T$ and $\theta'_T$ have diverged after training according to various notions of similarity (\cref{sec:background:similarity}).
We choose $T$ so that networks converge to a similar level of training and test performance (see details in \cref{ap:sec:training}).

If we fix a particular $\xi$, $\tmap^T$ describes a dynamical system whose outcome depends only on the initial parameters $\theta_0$.
This perspective has numerous advantages.
First, we can separate the effects of training noise and isolate instability to the action of $\tmap$.
As dynamical systems can diverge at exponential rates, dominating over stochastic effects \citep{wu2018sgd}, our deterministic experiments also lower bound the instability of regular stochastic training---i.e.
\begin{align*}
    \mathbb{E} \left[ d \left( \tmap(\theta, \xi), \tmap(\theta + \noise, \xi') \right) \right]
    \geq 
    \mathbb{E} \left[ d \left( \tmap(\theta, \xi), \tmap(\theta + \noise, \xi) \right) \right]
\end{align*}
for independently sampled noise $\xi$ and $\xi'$ and a similarity measure $d$.

\subsection{Spawn-And-Perturb Experiment}

Our experiment adapts the parent-child spawning experiment introduced by \citet{frankle2020linear} to the notion of stability introduced above. The procedure is as follows:
\begin{enumerate}
    \item Choose an initial state $\theta_0$ for a network.
    \item Train the network until the perturbation time $t$, giving $\theta_t = \tmap^t(\theta_0 \,;\, \xi_{1:t})$.
    \item Make two copies of the network, and perturb one by adding $\noise$ noise with magnitude $\sigma$ to get $\theta'_t = \theta_t + \sigma \noise$.\footnote{For interpretability, $\|\noise\|_2^2$ is normalized to match the expected scale of the network at initialization---see \cref{ap:sec:perturb-all} for details.}
    \item Train both original ($\theta_t$) and perturbed ($\theta'_t$) copies with identical training noise to get $\theta_T = \tmap^{T-t}(\theta_t \,;\, \xi_{t:T})$ and $\theta'_T = \tmap^{T-t}(\theta'_t \,;\, \xi_{t:T})$.
    \item Measure the resulting instability via $d\left(\theta_T, \theta'_T \right)$, where $d: \Theta \to \mathbb{R}^+$ is a dissimilarity score.
\end{enumerate}
By controlling $\theta_0$ and $t$, we can explore the stability of different points $\theta_t$ in the loss landscape.
More specifically, we select between different trajectories by randomly initializing $\theta_0$ or setting it to a pre-trained checkpoint from another task.
We then vary the perturbation time $t$ to examine how stability evolves during training.
Changing $\tmap$ (by choosing different model architectures, optimizers, hyperparameters, or training tasks) enables comparisons between different loss landscapes.

To quantify instability, we record the rate at which a dissimilarity score $d(\theta_T, \theta'_T)$ increases relative to the perturbation magnitude $\sigma$.
By sampling perturbations of different sizes and directions, we can estimate the size and shape of the local region around $\theta_t$ where $\tmap$ does not tend to cause divergence in terms of $d$.

Our experiment differs from \citet{frankle2020linear} and \citet{fort2020deep} in that they use independent training noise starting at $t$, instead of a single perturbation, to induce instability.
While this reflects the stability of ordinary training, our experiments have two key advantages: we can isolate our instability analysis to specific parts of the training trajectory from $t$ to $T$, and we can also apply much smaller or larger perturbations than training noise to quantify instability over a broader scale.
In \cref{fig:perturb-frankle}, we verify that instability in our method---i.e.~to single perturbations---implies instability in the methods of \citet{frankle2020linear, fort2020deep}---i.e.~to training noise.
%%%%%%%%%%%%%%%%%%%%%%%%%%%%%%%%%%%%%%%%%%%%%%%%%%%%%%%%%%%%%%%%%%%%%%%%%%%%%%%%%%%%%%

\subsection{Perturbations}

The stability of a dynamical system around a given point is direction-dependent.
We take this into consideration by sampling perturbations using two different methods, which give either a narrow or broad distribution of directions.
As described in \cref{ap:sec:perturb-all}, all noise samples are also normalized to a fixed $L^2$ relative to the network's initialization scale.

\subparagraph{Batch perturbation.}
Batch perturbations (Eq. \ref{eq:perturb-batch}) measure stability along the directions most likely to be explored during training, by simulating a single independently sampled optimization step: 
\begin{align}
\hat\noise_\text{Batch} &= \frac1n \sum_{i=1}^{b} \nabla \ell(x_i, y_i; \theta_t), \qquad x_i, y_i \sim \mathcal{D}, \label{eq:perturb-batch}
\end{align}
where $\nabla\ell$ is the gradient of the loss function $\theta_t$ the network weights, and $(x_i, y_i)$ are $b$ examples sampled from a minibatch of the training dataset $\mathcal{D}$. 
Ignoring factors like momentum, this is equivalent to taking an extra training step at time $t$, which is rescaled by the perturbation magnitude $\sigma$ instead of the learning rate.

\subparagraph{Gaussian perturbation.}
To measure stability in all directions generally, we use Gaussian perturbations (Eq. \ref{eq:perturb-gauss}), which are scaled versions of the the network's distribution at random initialization. For networks initialized with a Kaiming/He normal distribution \citep{he2015delving}, Gaussian perturbations are sampled as
\begin{align}
    \hat\noise_\text{Gaussian} = \left[ \noise_{i}^{(l)} \right]
    , \qquad
     \noise_{i}^{(l)} &\sim \mathcal{N} \left( 0, \frac{2}{n_{l-1}} \right)
    \label{eq:perturb-gauss}
\end{align}
where $\noise_{i}^{(l)}$ is the perturbation for the $i$th weight of layer $l$, $n_{l-1}$ is the number of inputs from the preceding layer (commonly called \emph{fan-in}), and $\mathcal{N}(0, s)$ is the normal distribution with mean 0 and standard deviation $s$.

Although this is not strictly uniform in all directions (when considering all of a network's weights as a single vector), we choose to match the scale of the network's initialization to ensure that perturbations do not disproportionally affect some layers more than others.
Since biases and normalization weights have constant initialization, we do not perturb them in our main experiments.\footnote{\cref{fig:butterfly-norm} shows results from perturbing biases and normalization weights.}

\subsection{Evaluating Functional Similarity}

\label{sec:background:similarity}

We use four methods to evaluate the functional similarity of networks in our spawn-and-perturb experiment: (1) $L^2$ distance in weight space $\lVert \theta_T - \theta'_T \rVert_2$, (2) the loss barrier in \cref{eq:barrier}, (3) the loss barrier after accounting for permutation symmetries using the weight matching algorithm from \cite{ainsworth2022git}, and (4) the representational similarity of intermediate layers measured via Angular CKA \citep{williams2021generalized}.

\subparagraph{$L^2$ divergence.}

In a linear dynamical system, the $L^2$ divergence $\|\theta_T - \theta'_T\|_2$ can diverge exponentially over time at a rate of $\|\noise\|_2 e^{\lambda t}$, where $\lambda$ is curvature dependent (see \cref{ap:sec:lyapunov} for derivation).
To determine whether this linear approximation holds for neural network training, we measure $L^2$ distance between parameter vectors over the course of training to look for exponential growth and to determine whether $L^2$ divergences are proportional to the perturbation magnitude $\sigma$.

\subparagraph{Barriers.}
Barriers measure the maximum increase in loss or error along the linear path between weights \citep{frankle2020linear, neyshabur2020being}.
We measure the training loss barrier as
\begin{align}
\label{eq:barrier}
\sup_{\alpha \in (0,1)} &\ell(\alpha\theta_T + (1\!-\!\alpha)\theta'_T) \!- \!\alpha\ell(\theta_T) \!- \!(1\!-\!\alpha)\ell(\theta'_T),
\end{align}
where $\alpha$ interpolates between the networks and $\ell$ is the loss function for the training data.
Since our work is concerned with the shape of the loss landscape in which training occurs, we report the cross-entropy loss barrier for training data ($\mathcal{B}_\mathrm{ce}$) and after accounting for permutations ($\mathcal{B}^{WM}_\mathrm{ce}$) throughout the main text. 
\cref{ap:sec:barrier} describes how we compute barriers in more detail.

%%%%%%%%%% Main Fig
%%%%% STANDARD SETTING W. BATCH NOISE FULL SIZE
\begin{figure*}[ht]
\vskip 0.1in
\begin{center}
\centerline{
    \includegraphics[height=0.23\linewidth]{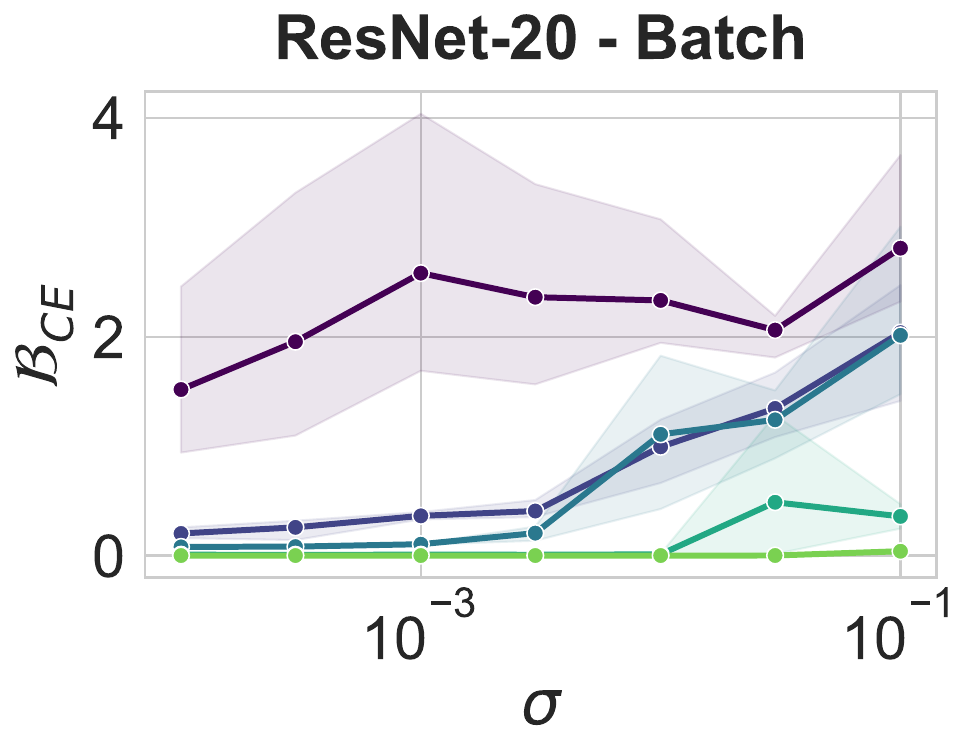}
    \includegraphics[height=0.23\linewidth]{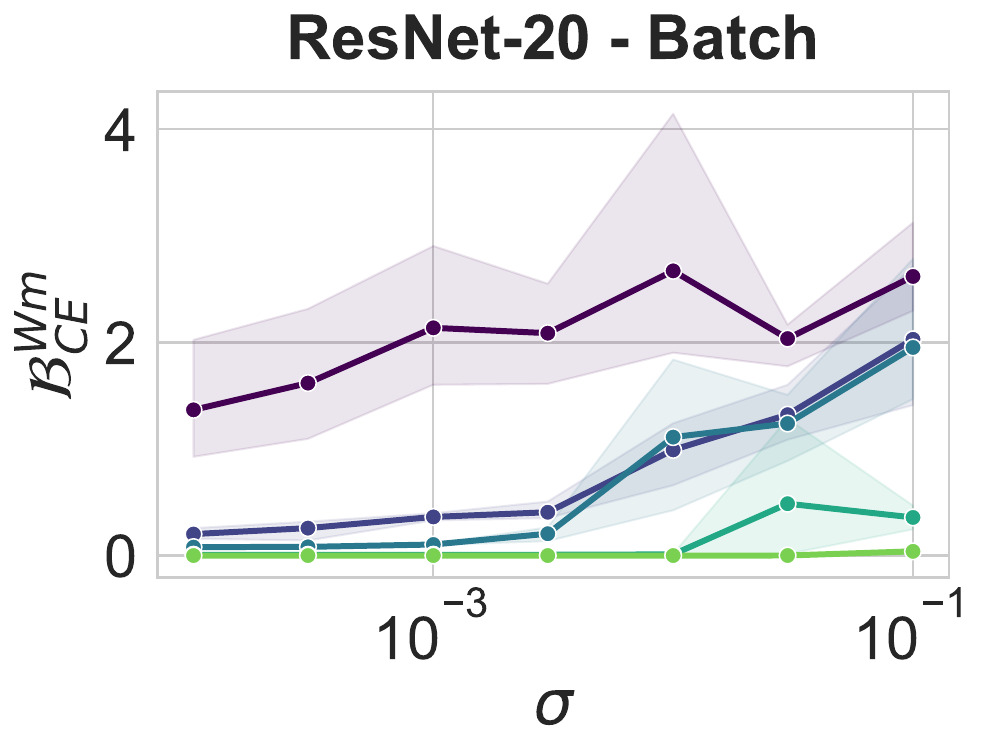}
    \includegraphics[height=0.23\linewidth]{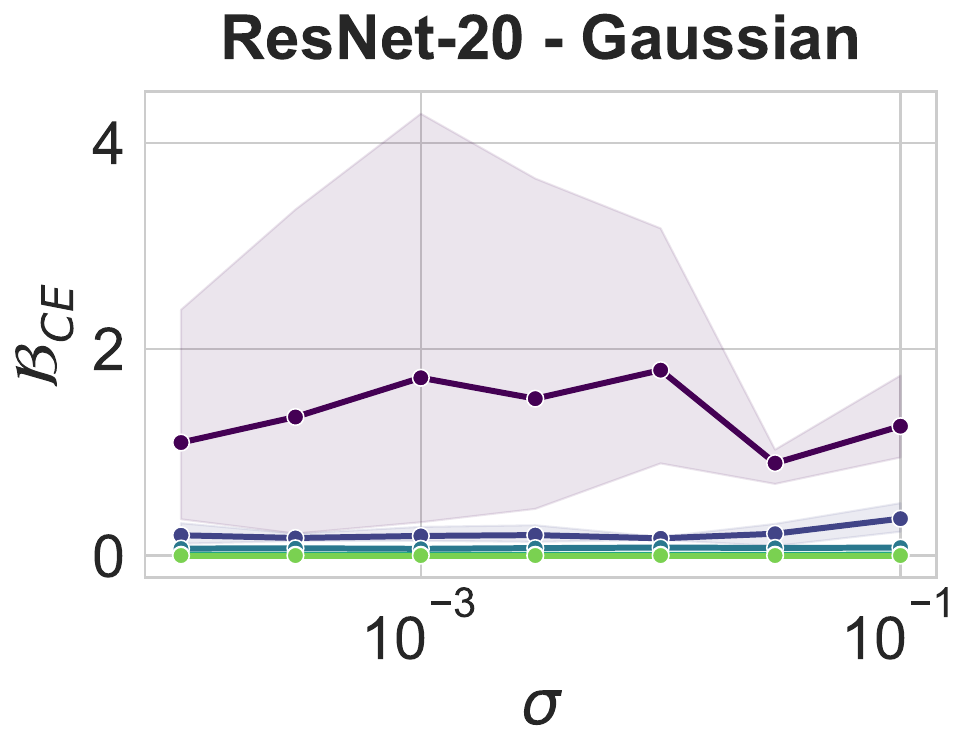}
}
\vskip -0.05in
\centerline{
    \includegraphics[width=0.6\linewidth]{figures/butterfly-hparams/no-decay-sanity-batch-lmc-0-1-loss-weighted-barrier-legend.pdf}
}
\vskip -0.2in
\caption{
Stability of ResNet-20 trained on CIFAR-10 with SGD (details in \cref{ap:sec:training}). Loss barriers on training data at the end of training (y-axis) are plotted against perturbation magnitude (x-axis) and perturbation step (color indicates fraction of total training time).  \textbf{Left:} barriers due to batch perturbation. \textbf{Middle:} batch perturbation barriers after accounting for permutations. \textbf{Right:} barriers due to Gaussian perturbation. For the same plots with log-scaled y-axes, see \cref{fig:butterfly-time-log}.
}
\label{fig:butterfly-time}
\end{center}
\vskip -0.2in
\end{figure*}
%%%%%%%

\subparagraph{Barriers modulo permutation.}

To consider whether training instability causes networks to converge to different permuted versions of the same loss basin, we apply the weight and activation matching algorithms from \citet{ainsworth2022git} to find a permutation of neurons $P$ that approximately minimizes the $L^2$ distance between two networks' weights or intermediate activations, respectively.
To measure the degree to which $\theta_T$ and $\theta'_T$ have been permuted with respect to each other, we record both the barrier between $\theta_T$ and $P[\theta'_T]$, and the fraction of identity elements (unpermuted neurons) in $P$.

As the two matching algorithms may not necessarily find the permutation that best minimizes barriers (i.e.~the \emph{barrier modulo permutation}), we follow \citet{sharma2024simultaneous} in treating the barrier between $\theta_T$ and $P[\theta'_T]$ as an upper bound.
As a result, if the barrier after permuting by $P$ is significantly reduced, we can say that training instability mainly causes permutations that do not substantially change a network's function.
However, if $P$ does not reduce the barrier between $\theta_T$ and $\theta'_T$, this could either mean that training instability causes networks to learn different functions, or that we have merely failed to find a permutation that does reduce barriers.

\subparagraph{Representational similarity.}
We also consider whether networks in our experiments differ in their penultimate representations using the angular version of Centered Kernel Alignment (CKA), a type of representational similarity metric.
CKA measures the cross-correlation between two arbitrary representations of the same data \citep{kornblith2019similarity}.
As defined in \cref{eq:cka}, we use Angular CKA with a linear kernel \citep{williams2021generalized} to measure the distance between the outputs of the last residual or attention block.
Since the resulting distance is an angle, $0$ indicates that two networks have perfectly similar representations, and $\pi / 2$ indicates that two networks have dissimilar representations. \cref{ap:sec:cka} includes full details.

%%%%%%%%%%%%%%%

\section{Experiments}

Full training details, including hyperparameters and train/test performance, are listed in \cref{ap:sec:training}.

\subsection{Early vs.~Late Training Instability}
\label{sec:butterfly_intro}

We first train residual networks \citep{he2015delving} on CIFAR-10 \citep{Krizhevsky2009} with SGD, using standard data augmentations and hyperparameter settings, including a 2\% warm-up period followed by linearly decaying learning rate.

\begin{figure*}[ht]
\begin{center}
\centerline{
    \includegraphics[height=0.22\linewidth]{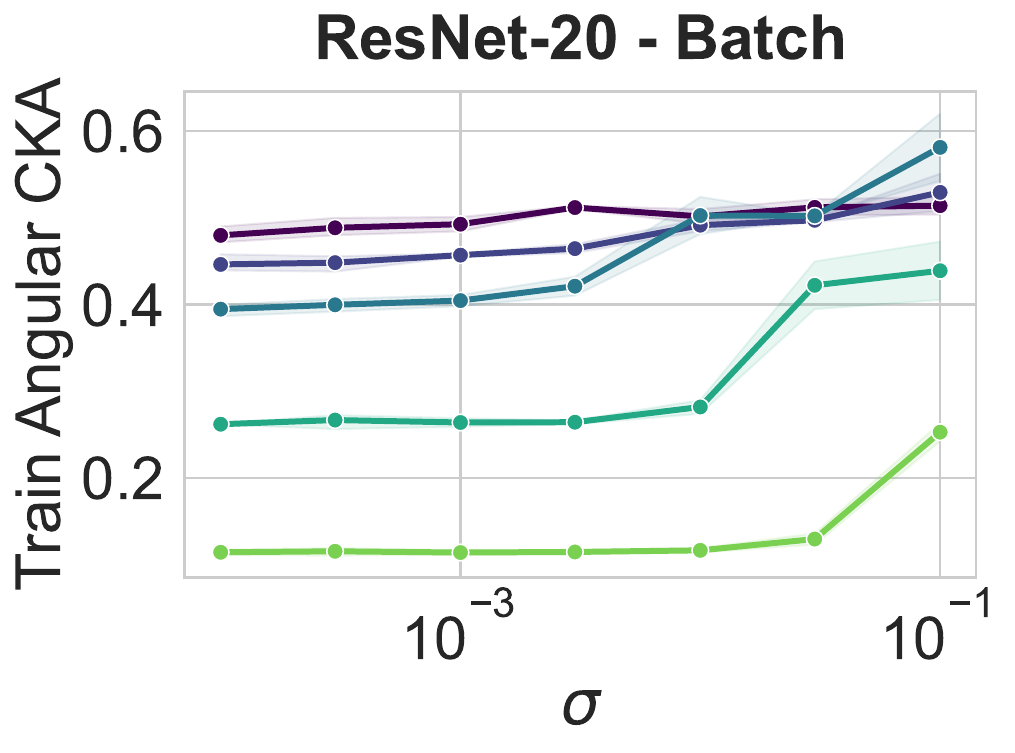}
    \includegraphics[height=0.22\linewidth]{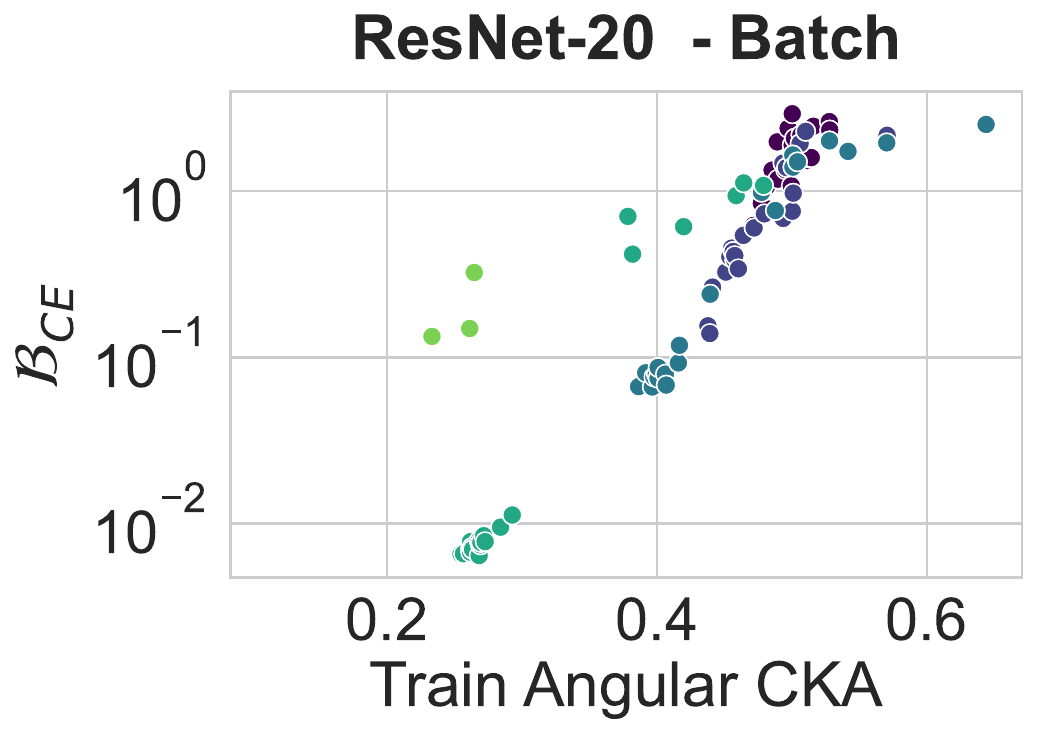}
    \includegraphics[height=0.22\linewidth]{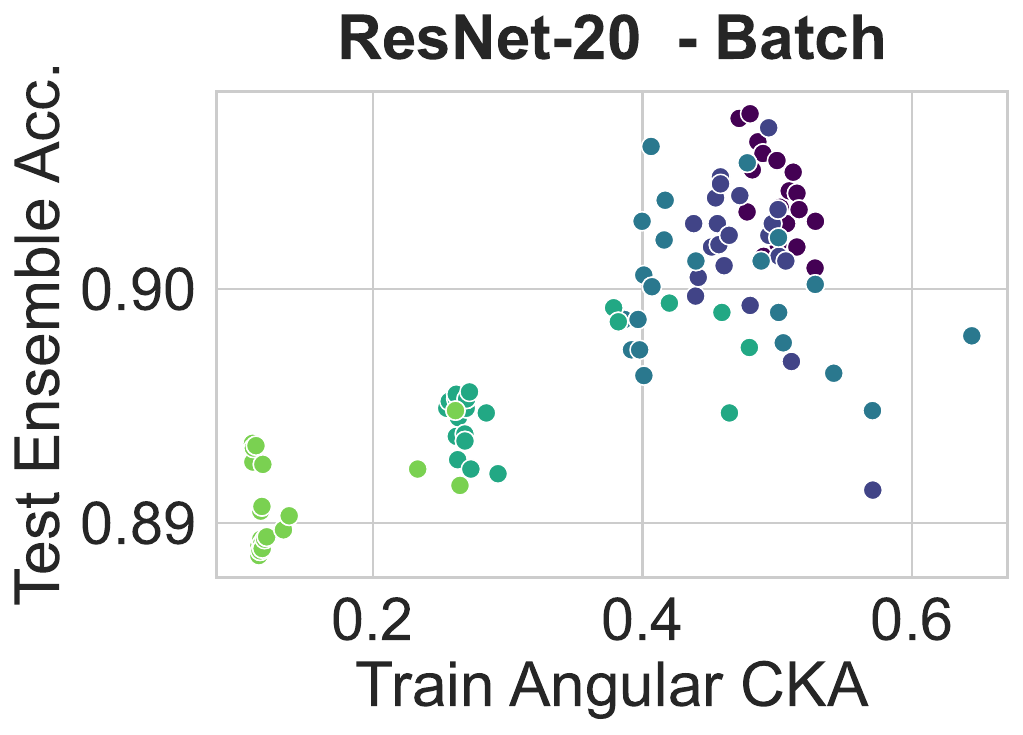}
}
\vskip -0.05in
\centerline{
    \includegraphics[height=0.04\linewidth]{figures/butterfly-hparams/no-decay-sanity-batch-lmc-0-1-loss-weighted-barrier-legend.pdf}
}
\vskip-0.2in
\caption{
\textbf{Left:} same as \cref{fig:barrier-perturb} but measuring representational similarity distance via Angular CKA (y-axis), defined in \cref{eq:cka}, between original and perturbed models after training for various perturbation times (colors) and magnitudes (x-axis).
\textbf{Middle:} barriers versus Angular CKA.
\textbf{Right:} test accuracy of an ensemble of the original and perturbed models after training (averaging logits), versus Angular CKA.
See \cref{fig:butterfly-cka-additional,fig:butterfly-cka-additional-barriers,fig:butterfly-cka-additional-ensemble,ap:fig:vit-cka-l2,ap:fig:bert-cka} for more hyperparameter settings and fine-tuning of ViT and BERT, respectively.
}
\label{fig:butterfly-cka}
\end{center}
\vskip -0.2in
\end{figure*}

\subparagraph{Early perturbations reliably cause large barriers.}
\cref{fig:butterfly-time} shows that training from randomly initialized networks is highly sensitive to initial conditions, as batch perturbations as small as 0.01\% of a network's weights produce large barriers.
We further reduce the perturbation magnitude by modifying only a fraction of the weights, finding that a single perturbed weight is sufficient to cause instability (\cref{fig:perturb-fraction}).

While prior work has shown that training noise near initialization causes barriers~\citep{fort2020deep, frankle2020linear}, we are the first to show that barriers can occur with extremely small perturbations concentrated in the first few steps, as applying the same perturbations as early as 0.5\% of the way through training results in significantly reduced barriers.
We name this initial instability after the ``butterfly effect'' in chaotic dynamical systems.

The stability increases over the first 0.5\% of training time is well within the 2\% warm-up period we use, and only very large perturbations (10\% of initialization) result in non-zero barrier after 50\% of training time.
While prior works find that models become stable to training noise after the first few epochs of training~\citep{fort2020deep, frankle2020linear}, we quantify the scale of perturbation needed to induce barriers beyond this critical point, showing that stability continues to increase throughout training.

\subparagraph{Early instability is direction-independent.}
Comparing batch versus Gaussian perturbations (\cref{fig:butterfly-time}), we find that although networks are more stable to the latter (which are evenly distributed in direction), networks perturbed at initialization have high barriers for both.
This shows early instability is mainly attributable to the network's state, and not the direction or magnitude of perturbation.
However, later instability does vary depending on the  direction of perturbation, which suggests that findings that use training noise \citep{fort2020deep, frankle2020linear} may not be transferrable to other kinds of perturbations.

\paragraph{Training divergence is unlikely to be caused by permutations.}
Comparing barriers with and without permutation alignment (\cref{fig:butterfly-time}), we find that applying permutations to minimize the $L^2$ distance between networks does not reduce barriers.\footnote{We omit $L^2$ distance between networks after alignment, as it is generally not reduced greatly by weight matching \citep{ito2024analysis}.}
While we cannot rule out the possibility that better (and more costly) alignment methods may reduce barriers, we argue that this is unlikely.
Prior work aligning \emph{differently} initialized networks finds that weight matching can reduce barriers to some degree even when it is outperformed by other methods \citep{pena2023re, navon2023equivariant, ainsworth2022git}, whereas in our case of \emph{identically} initialized networks, weight matching is unable to reduce barriers at all.
This suggests that training instability produces real functional differences between networks, as opposed to simply permuting weights that are otherwise equivalent.

%%%% Warmup, shallow, combo
\begin{figure*}[ht]
\vskip 0.1in
\begin{center}
\centerline{
    \includegraphics[height=0.22\linewidth]{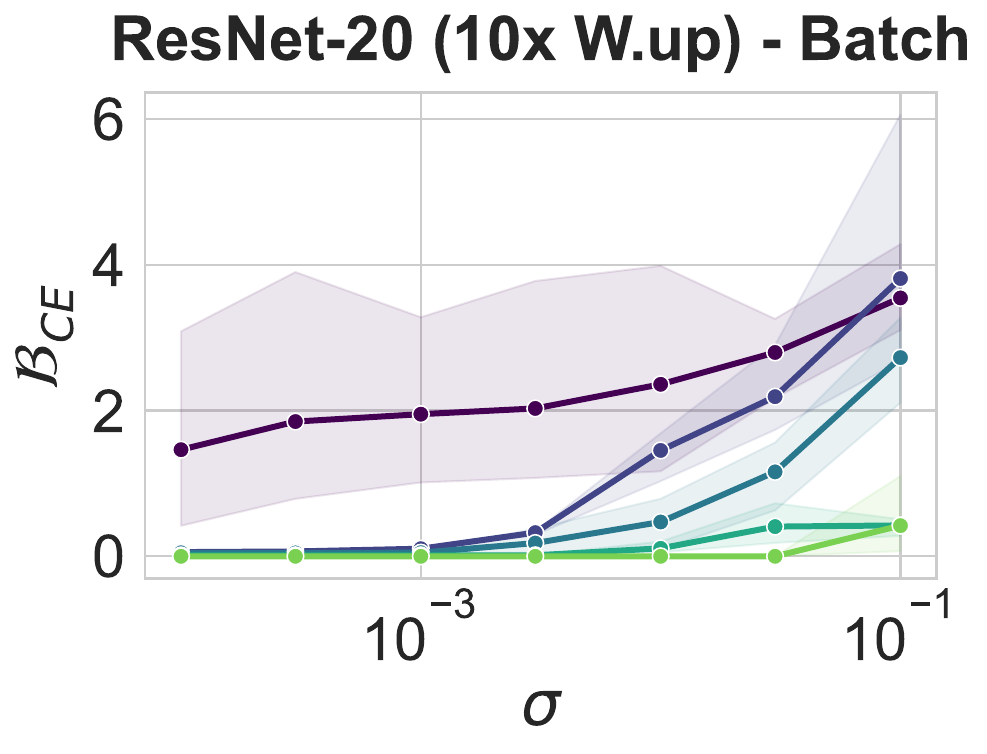}
    \includegraphics[height=0.22\linewidth]{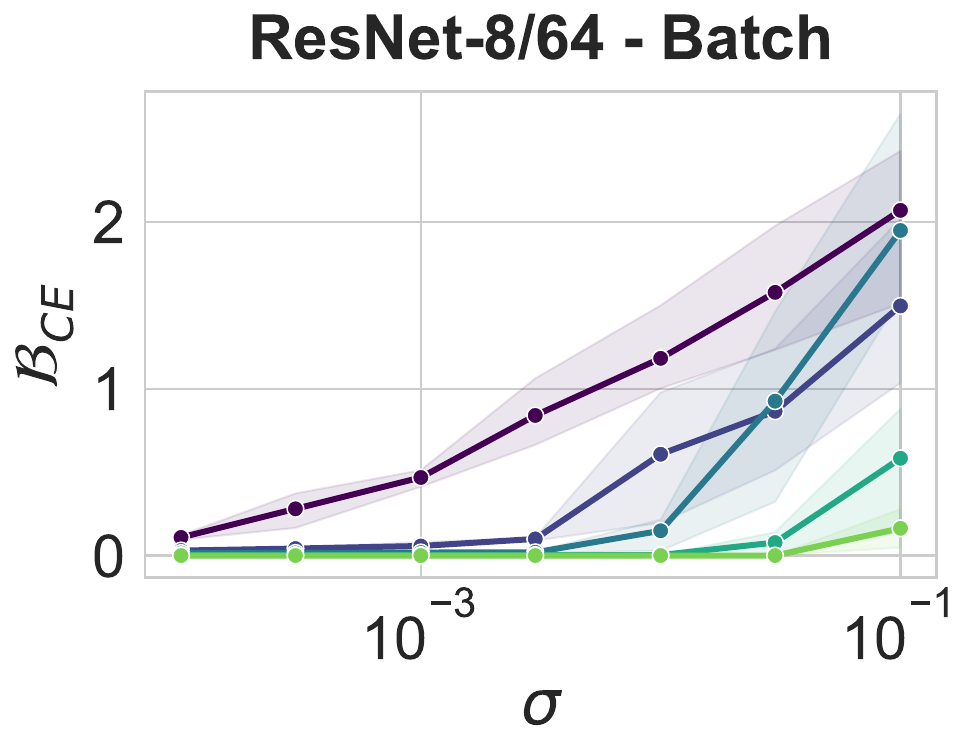}
    \includegraphics[height=0.22\linewidth]{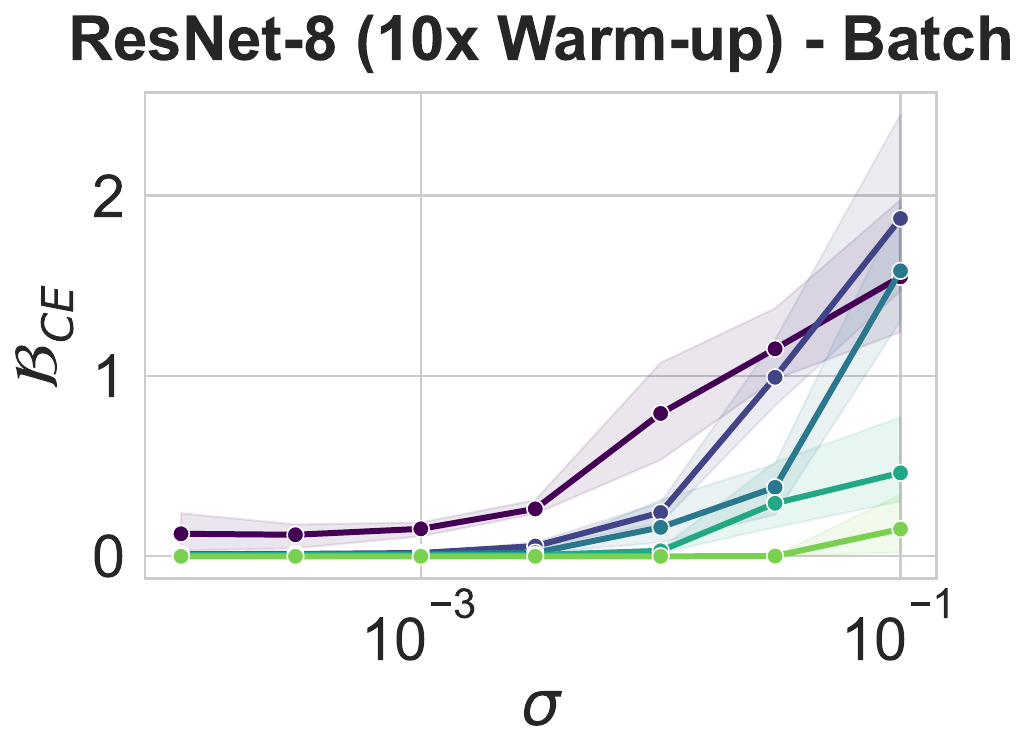}
}
\vskip -0.05in
\centerline{
    \includegraphics[width=0.5\linewidth]{figures/butterfly-hparams/no-decay-sanity-batch-lmc-0-1-loss-weighted-barrier-legend.pdf}
}
\vskip -0.2in
\caption{
Same as \cref{fig:butterfly-time} (left), but for models trained with $20\%$ warm-up time (\textbf{left}), a wider/shallower ResNet8 architecture (\textbf{middle}), and both settings (\textbf{right}). See \cref{fig:butterfly-warmup-arch-combo-log} for the same plots with log-scaled y-axes, and \cref{ap:sec:butterfly-hparams} for additional settings.}
\label{fig:butterfly-warmup-arch-combo}
\end{center}
\vskip -0.2in
\end{figure*}

\subsection{Functional Diversity}
\label{sec:functional-diversity}
Comparing the similarity of intermediate representations in \cref{fig:butterfly-cka,ap:fig:bert-cka}, we find that Angular CKA (Eq. \ref{eq:cka}) correlates with earlier and larger perturbations (left), as well as barriers after training (middle). This again indicates functional differences beyond weight symmetries.\footnote{Note that networks with zero barrier still have non-zero Angular CKA, likely because linearly connected networks can perform differently on individual examples \citep{yunis2022on}.}

Since model ensembling benefits from diversity, we also consider whether intentionally perturbing networks can improve ensembling performance.
This effect is most useful for fine-tuned networks, which necessarily have reduced diversity due to being trained from the same initial state far from random initialization. \cref{fig:butterfly-cka} (right) and \cref{fig:butterfly-cka-additional-ensemble} show that when ensembling the original and perturbed networks, ensemble performance indeed scales with Angular CKA dissimilarity.
However, fine-tuning ViT models on CIFAR-100 does not share this trend (\cref{ap:fig:vit-cka-l2}).
This contradiction may be explained by observations of similar performance between ensembling and averaging in \citet{utans1996weight}.

\subsection{Effect of Hyperparameter Settings}

We next compare the stability of different training schemes $\tmap$ for ResNets trained on CIFAR-10: no weight decay, 10x learning rate warm-up, 4x batch size, Adam, a shallow-wide architecture with similar numbers of parameters (exact details in \cref{ap:tab:CIFAR10hparams}).

\cref{fig:butterfly-warmup-arch-combo,fig:butterfly-bs-512} show that, in line with prior work \citep{altintas2023disentangling,vlaar2022can}, reducing learning rate (by increasing warm-up) and increasing batch size improve stability.
Adam and weight decay reduce stability, which we speculate may be due to their effect on the loss landscape's sharpness, which is known to affect SGD stability \citep{wu2018sgd}.
The shallow-wide architecture is most stable of these settings, which we speculate is due to its training dynamics being more closely aligned with the infinite-width, linearized kernel regime \citep{lee2019wide, fort2019deep}.

Next, we consider if a combination of stability-increasing hyperparameters could reduce barriers to 0 for networks perturbed at initialization.
We find that training the shallow-wide architecture combined with 10x learning rate warm-up improves stability over each individual setting (\cref{fig:butterfly-warmup-arch-combo} right), but does not eliminate barriers at initialization.

\subsection{Fine-tuning \label{sec:fine-tuning}}

\begin{figure*}[ht]
\begin{center}
\vskip 0.1in
\centerline{
    \includegraphics[height=0.23\linewidth]{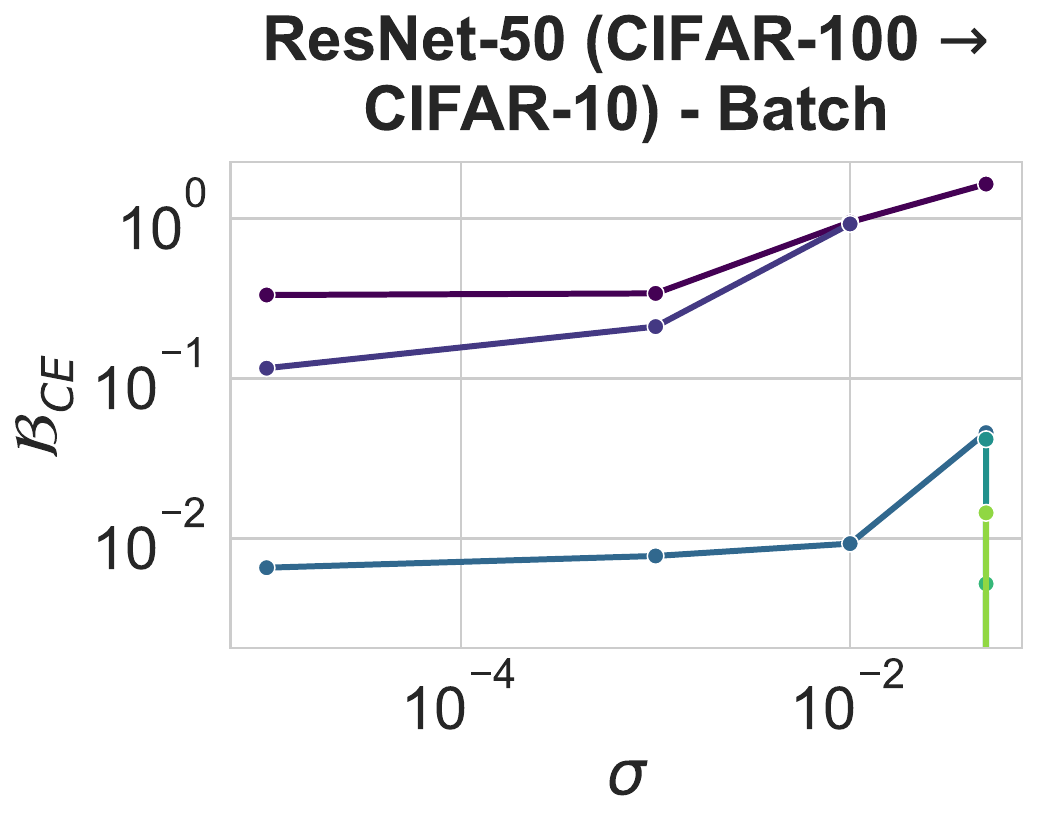}
    \includegraphics[height=0.23\linewidth]{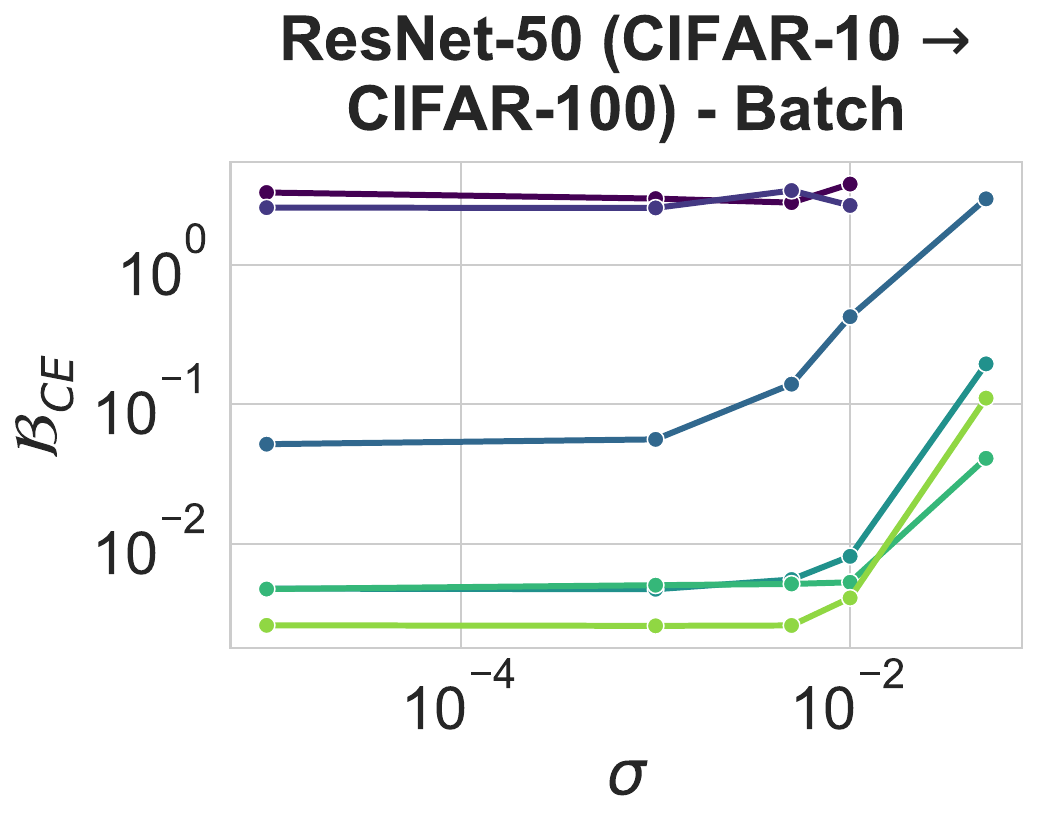}
    \includegraphics[height=0.23\linewidth]{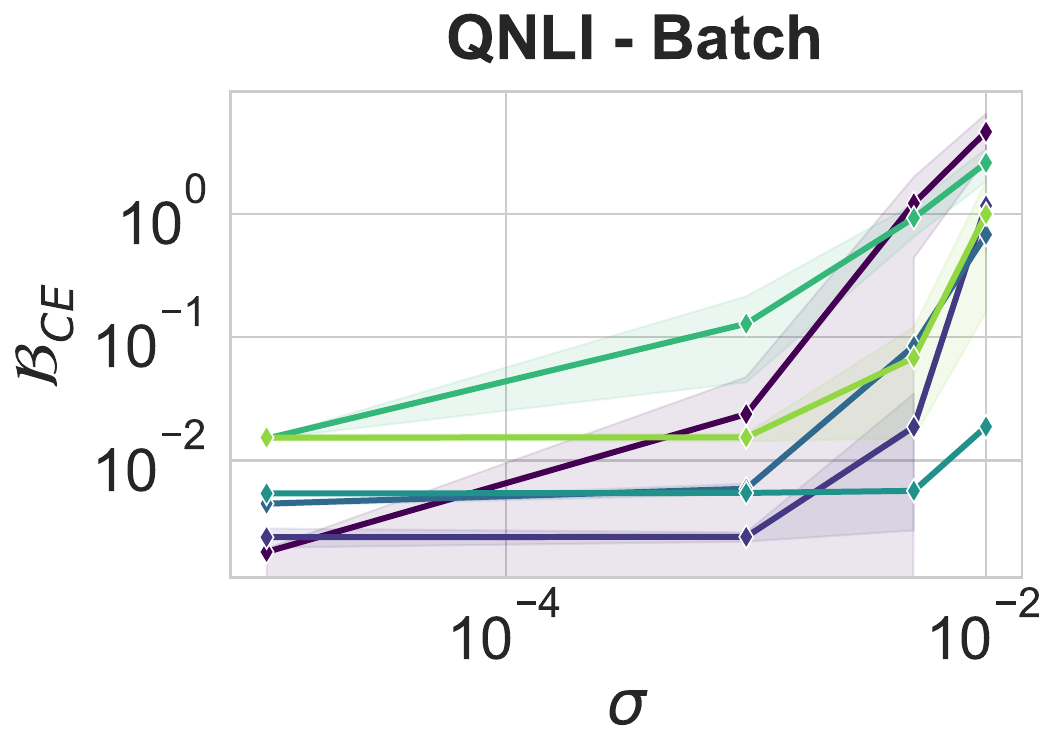}
}
\centerline{
    \includegraphics[height=0.04\linewidth]{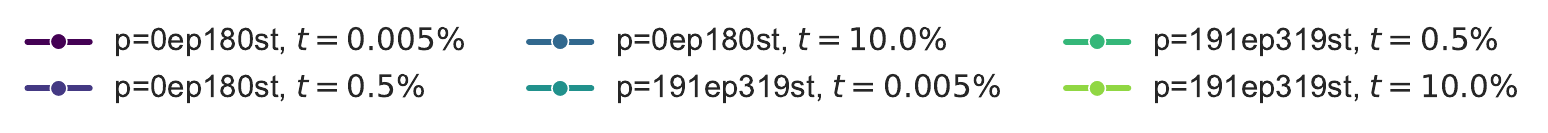}
    \hfill\includegraphics[height=0.04\linewidth]{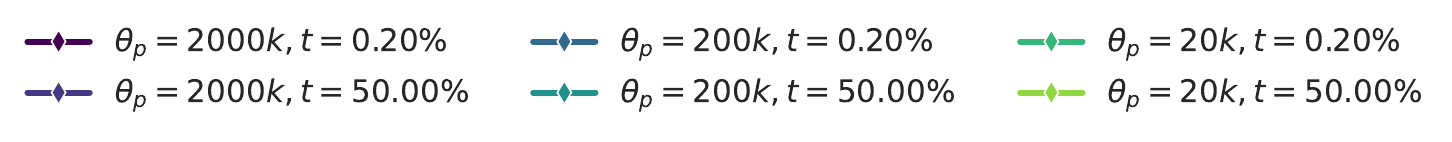}
}
\vskip -0.2in
\caption{
Stability of transfer learning on vision tasks: a ResNet-50 is pre-trained on CIFAR-100 and fine-tuned on CIFAR-10 (\textbf{left}) or vice versa (\textbf{middle}). Barriers (y-axis) are plotted against perturbation magnitudes (x-axis) for various pre-training durations and perturbation times (circle marker colors).  See \cref{ap:sec:finetuning-details} for details, and Tables \ref{tab:fine-tuning:across-task-late-cifar100-to-10}-\ref{tab:fine-tuning:across-task-late-cifar10-to-100} for barriers less than $10^{-2}$.
\textbf{Right}: fine-tuning stability of Multi-BERT on QNLI, starting from 20K, 200K, and 2000K checkpoints with early and late perturbation times (diamond marker colors).
For other tasks (MRPC, RTE, SST-2, and CoLA), see \cref{ap:fig:bert-fine-tuning:time}. For ViT and OLMo, see \cref{ap:fig:vit-fine-tuning:time,fig:olmo-fine-tuning:time} respectively.
}
\label{fig:fine-tuning}
\end{center}
\vskip -0.2in
\end{figure*}

Having explored the stability of the loss landscape along trajectories starting from random initialization, we next examine stability on transfer learning trajectories.
Fine-tuning is known to have greater stability since it starts from pre-trained networks that have non-random patterns of weights \citep{neyshabur2020being}, but the relative difficulty of the pre-training and transfer task can either increase or decrease stability to training noise \citep{vlaar2022can}.

We again move beyond the effects of training noise to quantify the exact perturbation times and scales at which transfer learning is unstable.
We consider task combinations from both vision and language domains, as fine-tuning the latter is known to be unstable to training noise \citep{juneja2023linear}.

\subparagraph{Pre-training stability depends on the tasks involved.} 
Starting with CIFAR-10 and CIFAR-100~\citep{Krizhevsky2009}, we pre-train two ResNet-50 networks with layer normalization on either task, and then fine-tune them on the opposite task starting from both early (0.24\% of pre-training) and late (100\% of pre-training) pre-trained checkpoints.

\cref{fig:fine-tuning} (left, center) and \cref{ap:fig:fine-tuning:across-task-early,ap:fig:fine-tuning:across-task-late} show that fine-tuning is generally more stable than compared to ResNet-20 (\cref{fig:butterfly-time}) or training the same models from random initialization (\cref{{ap:fig:resnet50-cifar10-random-init}}). This is especially true for later checkpoints ($p = 191ep319st$) and larger perturbations ($\sigma=0.1$, equivalent to 10\% of initialization), whereas fine-tuning from earlier checkpoints is more similar in barriers with regular training (\cref{fig:fine-tuning}).
This shows that, as in regular training, instability is mainly a function of pre-training time.

Stability is task-dependent, as transfer from CIFAR-100 to CIFAR-10 is more stable than in reverse.
This agrees with \citet{vlaar2022can}, who find that pre-training on related vs. random data improves or worsens (respectively) the barriers between two points along a training trajectory.\footnote{Our work differs in that we consider two diverging trajectories.}

\emph{Vision Transformers (ViTs).}
To study a different architecture in the vision domain, we perturb the fine-tuning trajectories of ViTs \citep{dosovitskiy2021vit} of varying sizes on CIFAR-100 (\cref{par:vit-finetune}). While we were only able to consider checkpoints at the end of pre-training, \cref{sec:ap:vit-stability} shows that, consistent with our previous findings, larger and earlier perturbations during fine-tuning lead to larger barriers.

\paragraph{Pre-training does not always increase fine-tuning stability.}
Having established that longer pre-training improves stability for small-scale vision models, we next examine heavily pre-trained language models.
This setting is particularly interesting because recent work by \citet{juneja2023linear} demonstrates that, unlike vision models \citep{neyshabur2020being}, training noise can cause language models to converge to distinct basins after fine-tuning.

To investigate this, we analyze the stability of Multi-BERT \cite{sellamMultiBERTsBERTReproductions2022}, which provides intermediate checkpoints for every 20,000 steps during pre-training. We take checkpoints at 20k, 200k, and 2000k (100\%) steps of pre-training time as starting points and fine-tune on various GLUE tasks \cite{wangGLUEMultiTaskBenchmark2018}: natural language inference (QNLI, RTE), paraphrase and similarity assessment (MRPC), sentiment classification (SST-2), and linguistic acceptability (CoLA).

\cref{fig:fine-tuning,ap:fig:bert-fine-tuning:time} show that BERT is more sensitive to the size of perturbations when compared with our vision experiments (\cref{fig:fine-tuning} left, middle).
For all pre-training checkpoints, earlier perturbations during fine-tuning consistently lead to larger barriers. However, unlike our vision settings, stability does not consistently improve with pre-training time. 
Notably, for QNLI and RTE (\cref{ap:fig:bert-fine-tuning:time}), the final pre-trained checkpoint (2000k) has the largest barriers.
When evaluating the pre-trained network on these tasks (\cref{tab:multibert-zero-shot}), we observe that the 2000k checkpoint has worse test accuracy, despite having lower cross-entropy when compared with the 200k and 20k checkpoints.
We speculate that this may be due to overfitting on the pre-training distribution,\footnote{MultiBERT was pre-trained for around 100 epochs.} which could cause the model to become brittle to perturbations during fine-tuning---a phenomenon termed ``catastrophic overfitting'' by \citet{springer2025overtrainedlanguagemodelsharder}.

\emph{Decoder-only models. } 
Billion-parameter decoder-only models are widely used in fine-tuning and model merging, but their training dynamics remain severely understudied. To address this gap, we fine-tune intermediate checkpoints of OLMo \citep{groeneveldOLMoAcceleratingScience2024} on the math problem dataset GSM8K \citep{cobbe2021trainingverifierssolvemath}. \cref{fig:olmo-fine-tuning:time} shows that pre-training longer can again reduce stability to fine-tuning, corroborating our MultiBERT findings.
Moreover, we observe the same trends---where earlier and larger perturbations result in higher barriers---as in previous settings.

\section{$L^2$ Divergence and Barriers}

\begin{figure}[t]
\vskip 0.1in
\begin{center}
\centerline{
    \includegraphics[width=0.48\columnwidth]{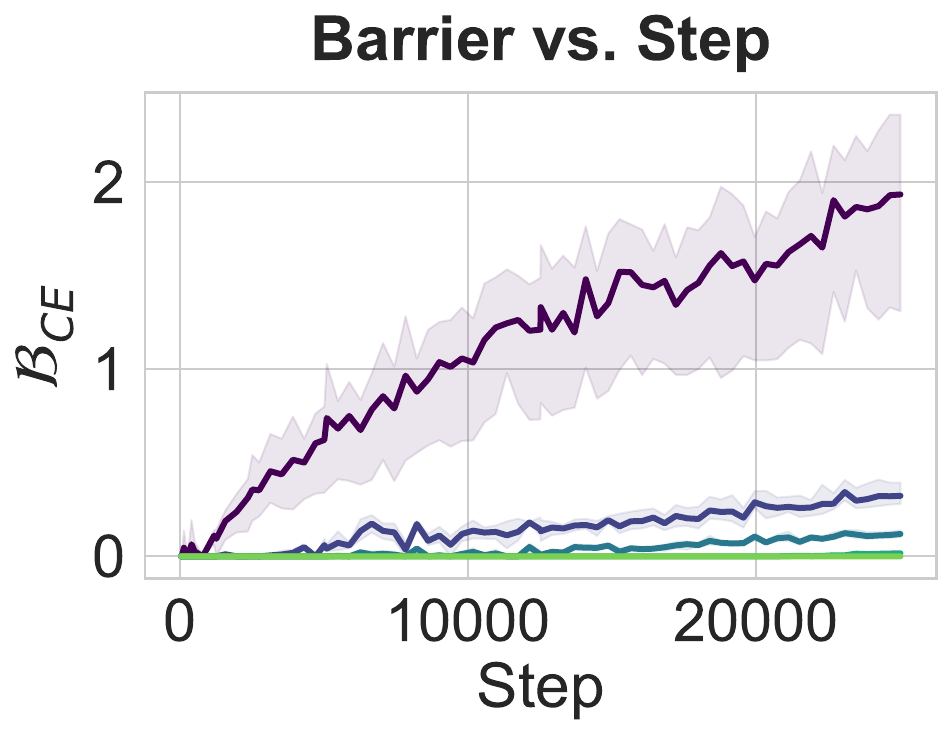}
    \includegraphics[width=0.48\columnwidth]{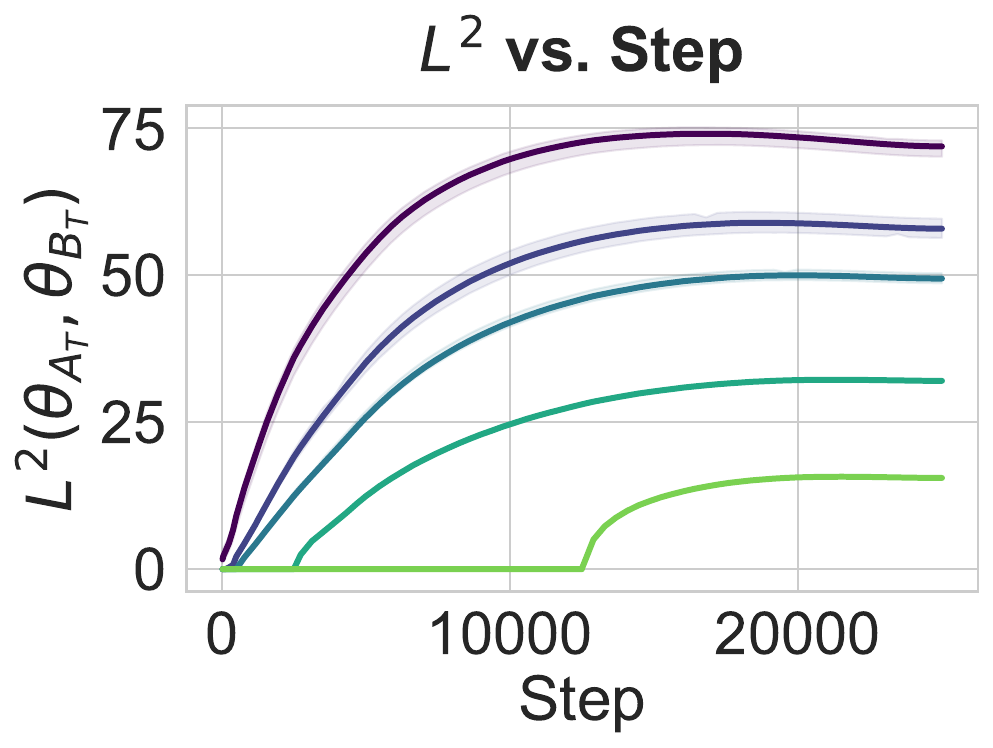}
}
\centerline{
    \includegraphics[width=0.65\columnwidth]{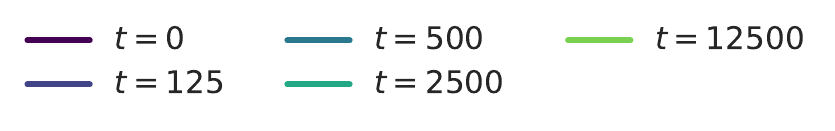}
}
\vskip -0.1in
\caption{Evolution of barriers (\textbf{left}) and $L^2$ (\textbf{right}) over training for standard ResNet20 trained on CIFAR-10. Each colored line averages over all perturbation magnitudes, as they are nearly indistinguishable.
}
\label{fig:butterfly-l2-lyapunov}
\end{center}
\vskip -0.2in
\end{figure}

\subparagraph{Barriers and $L^2$ divergence do not evolve according to a linearized dynamical system.}

\cref{fig:butterfly-l2-lyapunov} shows the rates at which barriers and $L^2$ divergence increase as training progresses.\footnote{
Since barriers and $L^2$ are negligible at perturbation time and grow throughout training, this indicates that our results are due to instability and not just the initial perturbation.}
Contrary to the linearized dynamics derived in \cref{ap:sec:lyapunov}, neither barrier nor $L^2$ increase exponentially over training.
More work is needed to explore other mechanisms that could drive these observed rates of divergence.

\subparagraph{Barriers scale with with exponential $L^2$ divergence in vision settings.}

\begin{figure}[t]
\vskip 0.1in
\centerline{
    \includegraphics[width=0.49\linewidth]{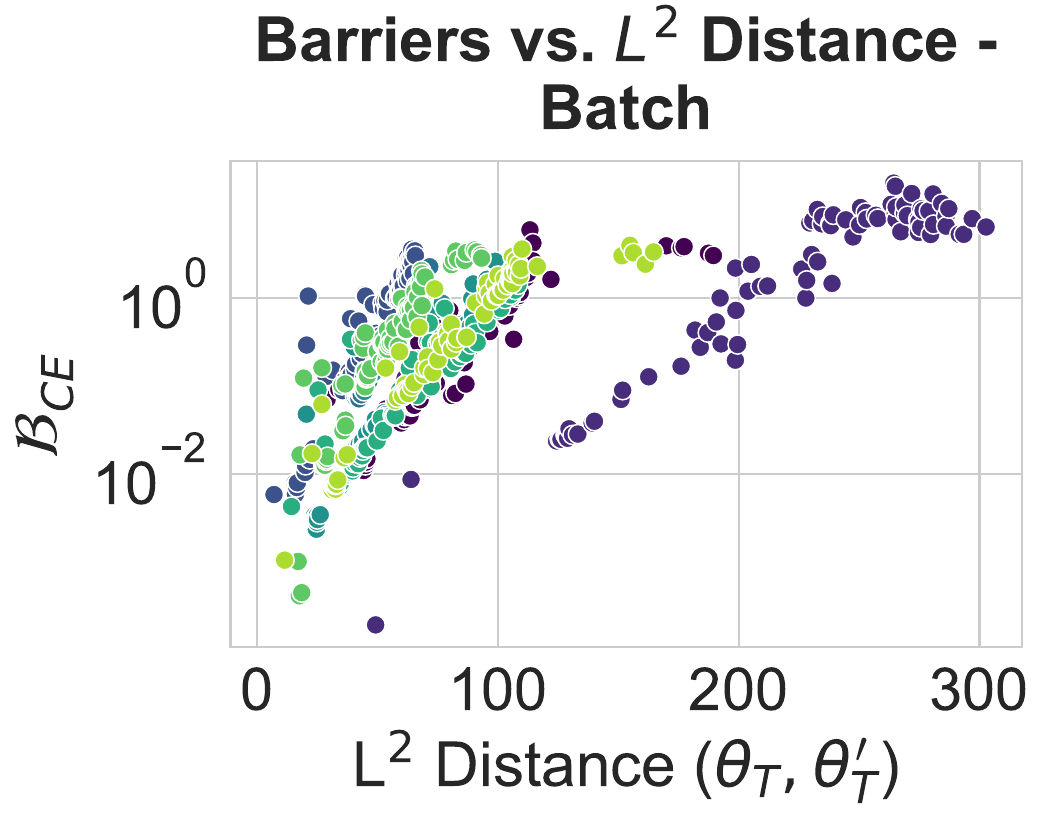}
    \includegraphics[width=0.49\linewidth]{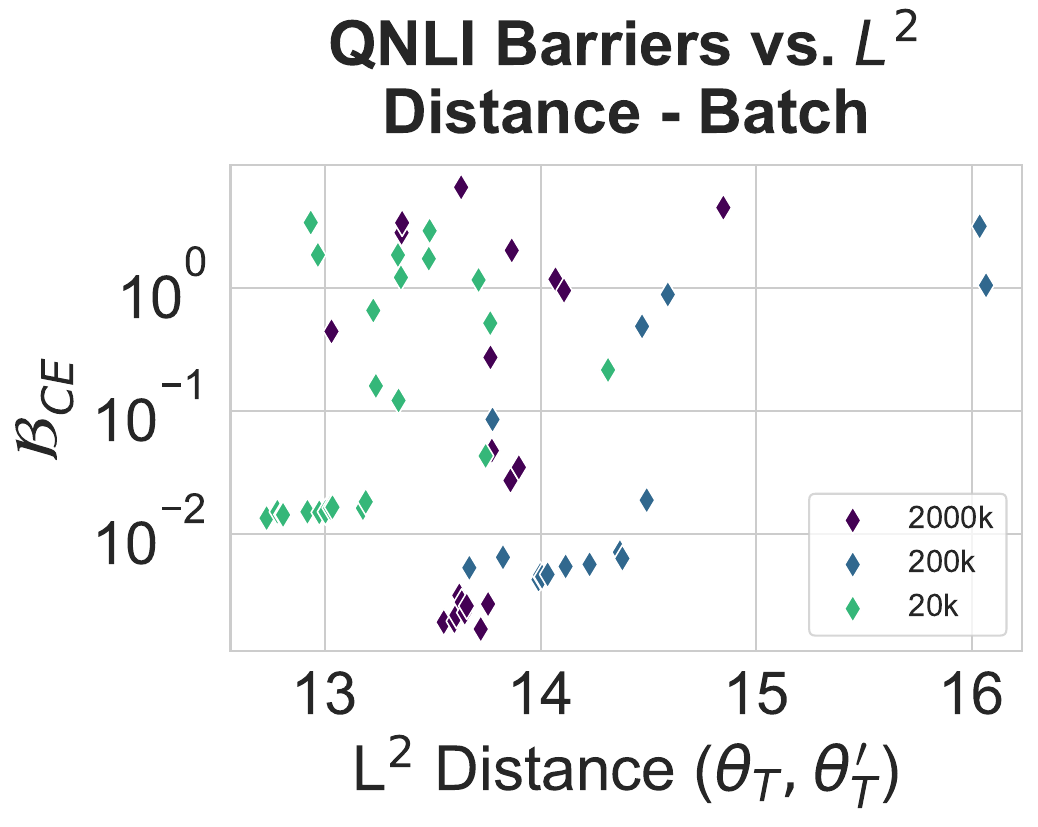}
}
\centerline{
    \includegraphics[width=0.99\linewidth]{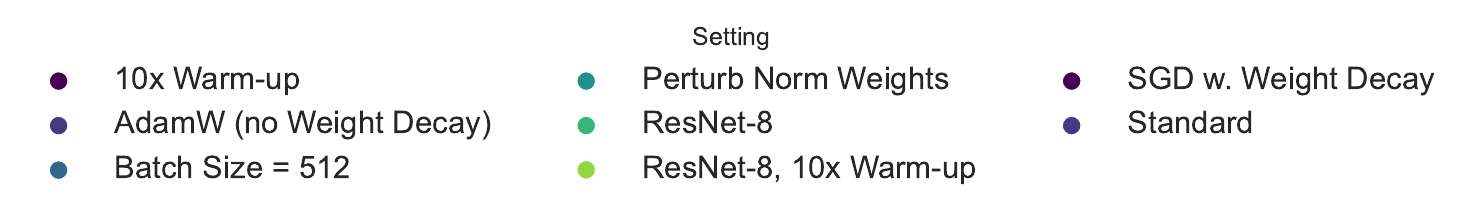}
}
\vskip -0.2in
\caption{
Train loss barriers vs. $L^2$ distance after training between the original and batch-perturbed models for ResNet-20 models trained on CIFAR-10 using various hyperparameter settings (\textbf{left}), and BERT models fine-tuned on QNLI (\textbf{right}).
For additional fine-tuning results, see \cref{ap:fig:finetune-cifar-l2-barr,ap:fig:bert-transfer-l2-barr,ap:fig:vit-cka-l2,fig:olmo-fine-tuning:time} for other BERT tasks, ResNet, ViT, and OLMo respectively.
}
\label{fig:l2-barriers}
\end{figure}

Although networks can diverge in weight space without increasing barriers \citep{frankle2020linear, vlaar2022can}, we find in our experiments that barriers and $L^2$ divergence after training exhibit a strong log-linear relationship (\cref{fig:l2-barriers} left).
This finding differs from \citet{vlaar2022can} in that they look at the distance traveled from initialization, whereas we look at the distance between training trajectories which started from the same point.
We find that the proportion of identity elements in the aligning permutations $P$ is also related to barriers, albeit to a weaker extent (\cref{ap:fig:butterfly-fixed-barriers}). Since $P$ minimizes $L^2$ distance between $\theta_T$ and $\theta'_T$, this is likely due to the correlation between barriers and $L^2$ divergence.

Interestingly however, fine-tuned language models show little or no correlation between $L^2$ divergence and barriers (\cref{fig:l2-barriers} right, \cref{ap:fig:bert-transfer-l2-barr}).
This suggests that the relationship between $L^2$ and barrier may only appear in smaller-scale models, which highlights the need for large-scale and multi-modal experiments \citep{juneja2023linear}.

\section{Discussion \& Conclusion}

We present a method for measuring whether neural network training is stable (reliably converging to the same basin), for a distribution of perturbations, applied at any time in training, on any initial network weights, and for any training procedure.
This method allows us to evaluate stability over more conditions, and at a finer precision, than was possible in prior works that only consider the effects of training noise \citep{vlaar2022can, fort2020deep, frankle2020linear}. 
Our experiments show that although randomly initialized networks are extremely unstable, stability rapidly increases with training to be robust to perturbations much larger than training noise.

Our work is consistent with the finding in \citet{sarnthein2023random} that a student network initialized very close to a random teacher nevertheless diverges quite far after training.
Further work is needed to understand why, unlike in our setting, the student in \citet{sarnthein2023random} remains in the same linearly connected basin as the random teacher.

While instability near initialization is universal, many trends are inconsistent and depend on the task or model.
We find that (1) the rate at which stability increases along training trajectories varies greatly, (2) more pre-training does not always improve stability during fine-tuning, (3) $L^2$ divergence correlates strongly with barriers in some cases but not others, and (4) the rates at which $L^2$ and barriers diverge do not match that of a straightforward dynamical system.
While the specific counter-examples we have surfaced are sufficient evidence for these results, a detailed exploration of their underlying causes and the circumstances in which they hold (such as in isolating the effects of task versus architecture) is left for future work.
Further investigation is also needed to determine (1) if certain hyperparameter settings entirely eliminate instability at initialization, and (2) what perturbations, if any, can be used to reliably improve ensemble performance.

\newpage

\section*{Acknowledgements}

Special thanks to Gaurav Iyer, David Mickish, Ekansh Sharma, Sidak Pal Singh, and Julien Boussard for discussions and comments.
This research was supported by an NSERC Discovery grant, the Canada CIFAR AI Chairs program, and Fonds de recherche du Québec--Nature et Technologies (FRQNT doctoral research award \#352816).
Computing resources were provided by Mila--Quebec Artificial Intelligence Institute, and the NVIDIA Corporation.

\section*{Impact Statement}
This paper presents work whose goal is to advance the field of 
Machine Learning. There are many potential societal consequences 
of our work, none which we feel must be specifically highlighted here.

\bibliography{citations}
\bibliographystyle{icml2025}

%%%%%%%%%%%%%%%%%%%%%%%%%%%%%%%%%%%%%%%%%%%%%%%%%%%%%%%%%%%%%%%%%%%%%%%%%%%%%%%
%%%%%%%%%%%%%%%%%%%%%%%%%%%%%%%%%%%%%%%%%%%%%%%%%%%%%%%%%%%%%%%%%%%%%%%%%%%%%%%
% APPENDIX
%%%%%%%%%%%%%%%%%%%%%%%%%%%%%%%%%%%%%%%%%%%%%%%%%%%%%%%%%%%%%%%%%%%%%%%%%%%%%%%
%%%%%%%%%%%%%%%%%%%%%%%%%%%%%%%%%%%%%%%%%%%%%%%%%%%%%%%%%%%%%%%%%%%%%%%%%%%%%%%
\newpage
\appendix
\onecolumn
\section{Training Details}\label{ap:sec:training}

In this section, we provide details about our training methodology.   
Unless otherwise specified, we conducted all ResNet experiments on individual NVIDIA RTX 8000 GPU with 4 CPU cores. ViT and language model experiments were conducted on NVIDIA L40S GPUs.

\subsection{CIFAR-10 Hyperparameter Experiments}

\begin{figure}[ht]
\vskip 0.1in
\begin{center}
\centerline{
    \includegraphics[height=0.18\linewidth]{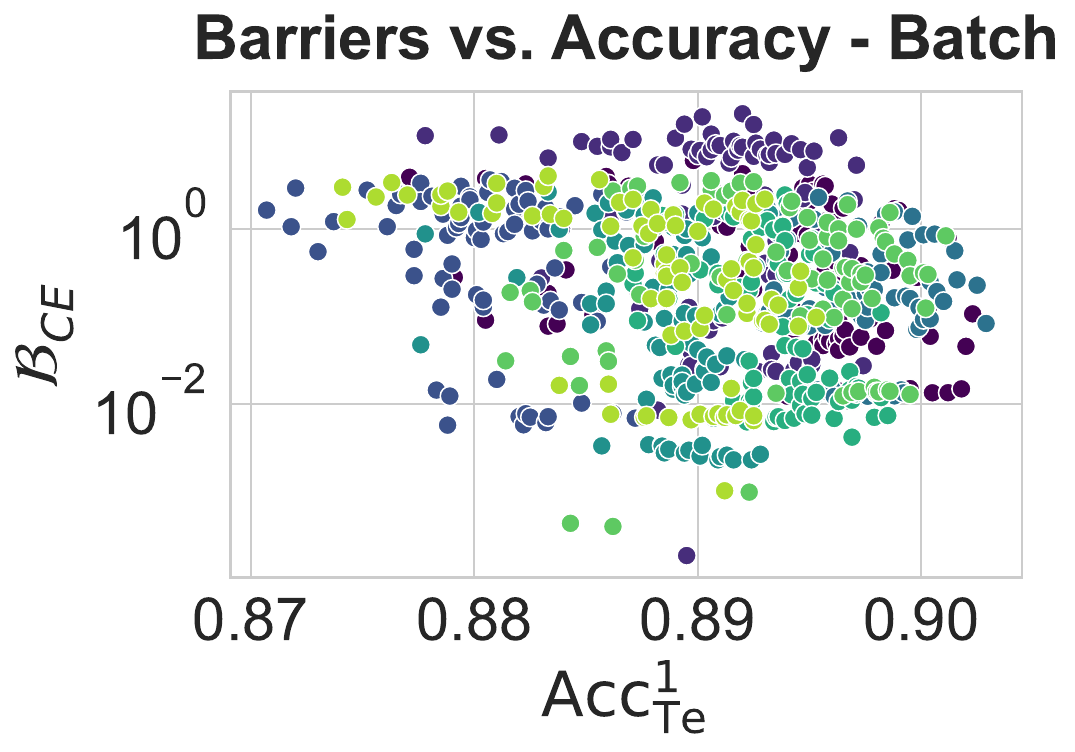}
    \includegraphics[height=0.18\linewidth]{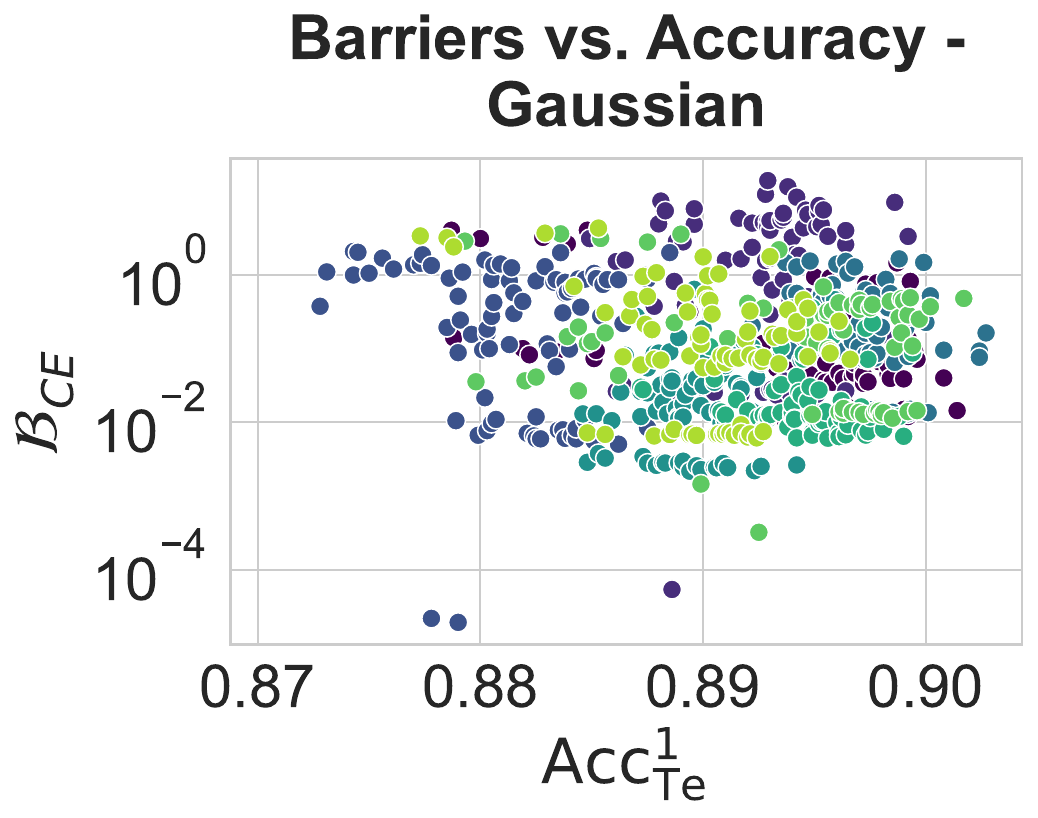}
    
    \includegraphics[height=0.18\linewidth]{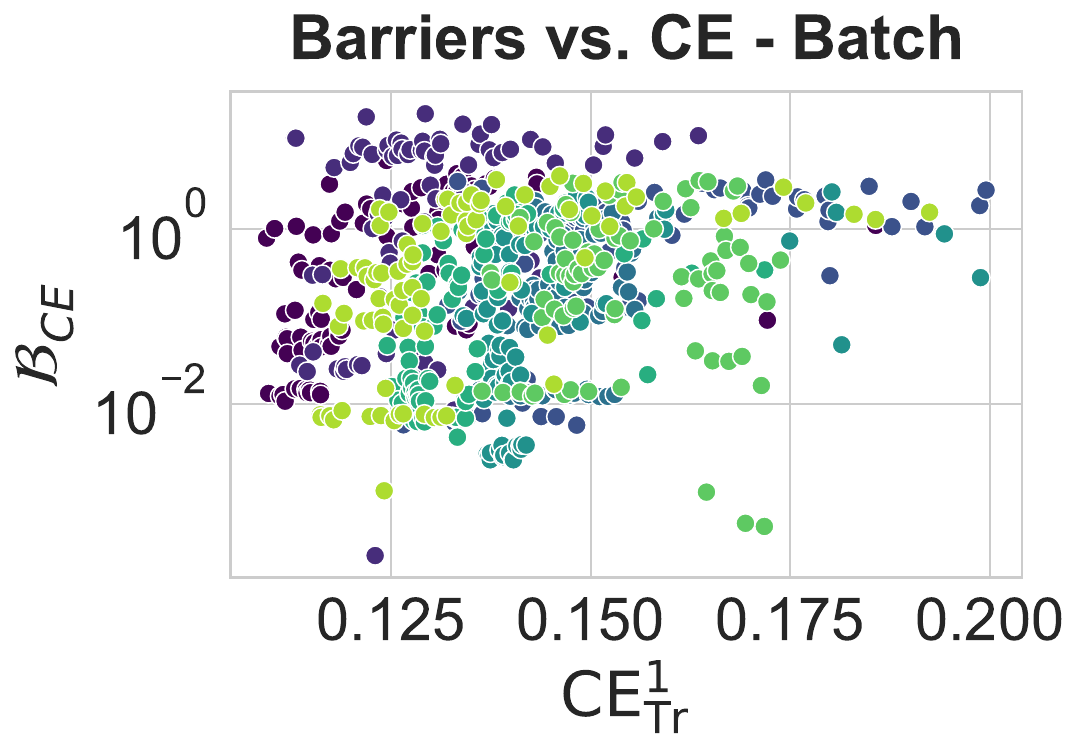}
    \includegraphics[height=0.18\linewidth]{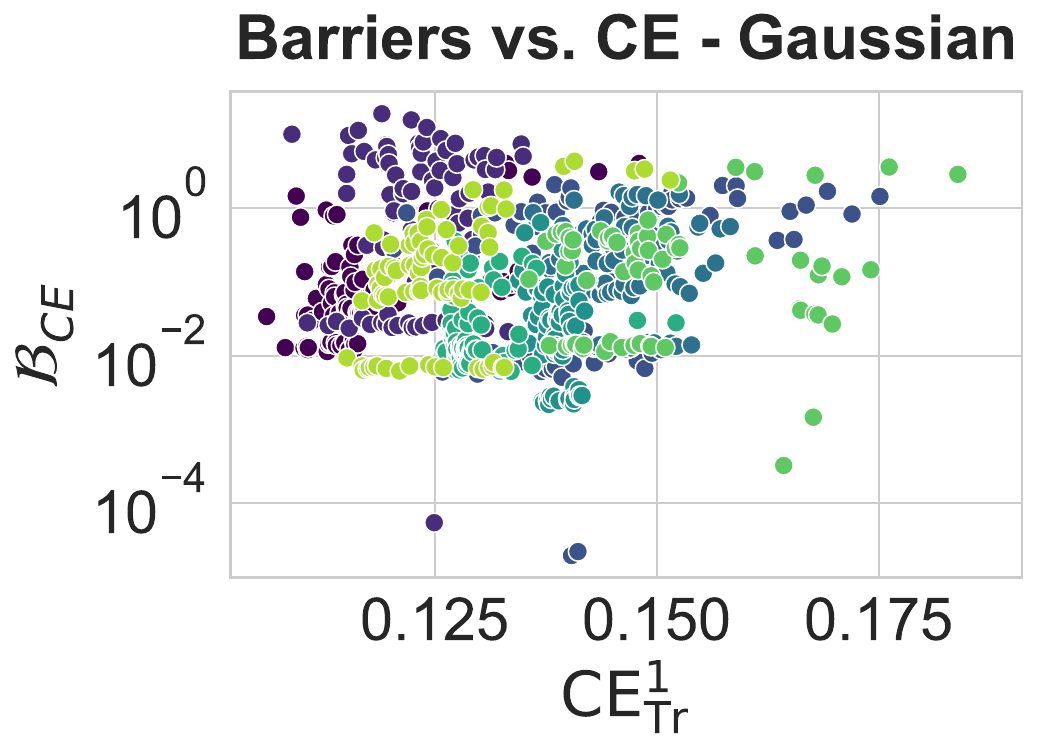}
}
\centerline{
    \includegraphics[width=0.6\columnwidth]{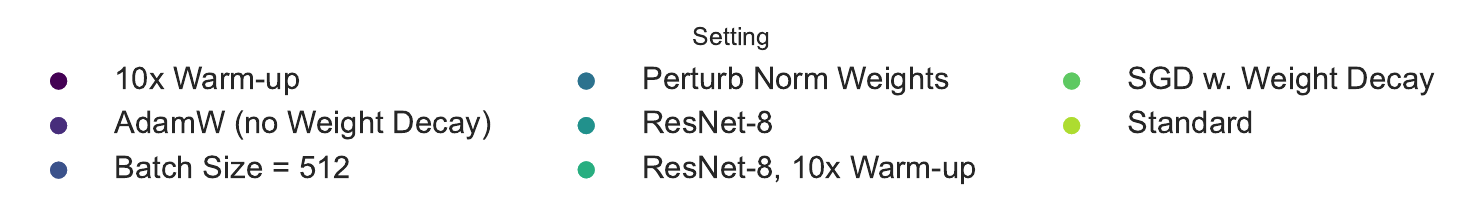}
}
\vskip-0.2in
\caption{
Train loss barriers against test accuracy (left) and training cross entropy loss (right) of the perturbed model at the end of training.
}
\label{fig:ce-barr}
\end{center}
\end{figure}

We train residual convolutional models \citep{he2015delving} on the CIFAR-10 dataset \citep{Krizhevsky2009} using the hyperparameter settings in \cref{ap:tab:CIFAR10hparams}.
Training times are chosen so that cross-entropy loss on the training data for the perturbed model is below $0.15$ after training, on average.
All models have test accuracies within 2 percentage points (88-90\%).
For ease of interpretation, training times are rounded up to the nearest 5000 steps.
Although the models in our experiments are not fully converged, and some variations remain between different hyperparameter settings, we did not find our results correlate with different training times, or the network's final train or test performance (\cref{fig:ce-barr}).

\begin{table*}[t]
\caption{
Hyperparameter settings for ResNet-20 trained on CIFAR-10, along with the test accuracy and training cross-entropy loss of the perturbed model at the end of training. Each setting is averaged over batch and Gaussian perturbations applied at various time steps and scales, with each configuration repeated with three seeds.
}
\label{ap:tab:CIFAR10hparams}
\vskip 0.15in
\begin{center}
\begin{small}
\begin{tabular}{lrlrlrlrlr}
\toprule
Setting & Model & Optimizer & LR & W-up & WD & BS & Steps & $\mathrm{Acc}^1_{\mathrm{te}}$ & $\mathrm{CE}^1_{\mathrm{tr}}$ \\
\midrule
Standard & ResNet20-32 & SGD & 0.100 & 0.020 & -- & 128 & 25000 & 0.89 ± 0.00 & 0.13 ± 0.01 \\
10x Warm-up & ResNet20-32 & SGD & 0.100 & 0.200 & -- & 128 & 25000 & 0.89 ± 0.01 & 0.12 ± 0.01 \\
AdamW (no Weight Decay) & ResNet20-32 & AdamW & 0.003 & 0.020 & -- & 128 & 20000 & 0.89 ± 0.00 & 0.13 ± 0.01 \\
Batch Size = 512 & ResNet20-32 & SGD & 0.100 & 0.020 & -- & 512 & 10000 & 0.88 ± 0.00 & 0.15 ± 0.02 \\
Perturb Norm Weights & ResNet20-32 & SGD & 0.100 & 0.020 & \checkmark & 128 & 20000 & 0.90 ± 0.00 & 0.15 ± 0.00 \\
ResNet-8 & ResNet8-64 & SGD & 0.100 & 0.020 & -- & 128 & 25000 & 0.89 ± 0.00 & 0.14 ± 0.01 \\
ResNet-8, 10x Warm-up & ResNet8-64 & SGD & 0.100 & 0.200 & -- & 128 & 25000 & 0.89 ± 0.00 & 0.13 ± 0.01 \\
SGD w. Weight Decay & ResNet20-32 & SGD & 0.100 & 0.020 & \checkmark & 128 & 20000 & 0.89 ± 0.01 & 0.15 ± 0.01 \\
\bottomrule
\end{tabular}
\end{small}
\end{center}
\vskip -0.1in
\end{table*}

To simplify weight and activation matching, we use layer normalization \citep{ba2016layer} instead of batch normalization, resulting in a slight reduction in performance.
When evaluating barriers after permutation alignment, this avoids having to do additional inference passes to correct the batch normalization statistics at each interpolation step \citep{jordan2022repair}.

\subsection{Finetuning Experiments\label{ap:sec:finetuning-details}}
\paragraph{CIFAR pre-training \label{ap:CIFAR100-pre-train-details}}
We pre-trained two ResNet-50 models with different initializations on on both CIFAR-10 and CIFAR-100 datasets, with layer normalization. Each model was trained for 75,000 steps (approximately 200 epochs) using SGD with momentum 0.9 and cosine annealing schedule. We used a peak learning rate of 0.1 with a 2.5\% warm-up, with a weight decay of $10^{-4}$ and a batch size of 128. The model was trained with horizontal flips, random rotations up to 10 degrees, random translations up to 4 pixels, and cutout augmentation with 2 pixels. 

\paragraph{CIFAR Fine-tuning}
Starting from our pre-trained ResNet-50 checkpoints, we fine-tuned each model on CIFAR-10 (or CIFAR-100) using stochastic gradient descent with momentum 0.9 for 20,000 steps.

\paragraph{BERT Fine-tuning Experiments}

We conducted experiments on the GLUE benchmark \cite{wangGLUEMultiTaskBenchmark2018} using the MultiBERT model \cite{sellamMultiBERTsBERTReproductions2022}, specifically using checkpoints from \texttt{google/multiberts-seed\_0} and \texttt{google/multiberts-seed\_1} available on HuggingFace.\footnote{\url{https://huggingface.co/google/multiberts-seed_0-step_0k} and \url{https://huggingface.co/google/multiberts-seed_1-step_0k}.}
All tasks share the same base hyperparameters of AdamW \citep{adamw} with a learning rate of $2 \times 10^{-5}$ and a weight decay of 0.01, while the batch size and training duration were scaled according to dataset size, as detailed in \cref{ap:tab:multibert-fine-tuning-params}.

Due to computational constraints, we selected QNLI from among the larger datasets with more than 100k examples (QNLI, QQP, MNLI), and followed \citet{devlinBERTPretrainingDeep2019} by fine-tuning for three epochs. For the medium-sized SST-2 dataset, we trained for 2,500 steps. Small datasets (RTE, MRPC) with 2.5k–3.7k examples were trained for 500 steps using a batch size of 32, while the medium-small dataset CoLA (5.7k–8.5k examples) was trained for 1,500 steps with the same batch size.
These settings ensure that all models achieved a training cross-entropy loss below 0.2, although our networks appear to have overfit the fine-tuning task in some cases (\cref{tab:multibert-zero-shot}).

\begin{table}[h]
\vskip 0.15in
\begin{center}
\caption{Task-specific hyperparameters for fine-tuning MultiBERT on GLUE tasks. Training schedule transitions from step-based to epoch-based for larger datasets to ensure sufficient coverage of training data.}
\label{ap:tab:multibert-fine-tuning-params}

\begin{small}
\begin{tabular}{lrrr}
\toprule
Dataset & Examples & Batch Size & Training Schedule \\
\midrule
QNLI    & 105k     & 128        & 3 epochs         \\
SST-2   & 67k      & 128        & 2500 steps       \\
CoLA    & 8.5k     & 32         & 1500 steps       \\
MRPC    & 3.7k     & 32         & 500 steps        \\
RTE     & 2.5k     & 32         & 500 steps        \\
\bottomrule
\end{tabular}
\vskip -0.1in
\end{small}
\end{center}
\end{table}

\begin{table}[ht]
\caption{Multi-BERT (seed 0) test performance of pre-trained checkpoints before fine-tuning (zero-shot evaluation) and after fine-tuning.}
\label{tab:multibert-zero-shot}
    \centering
    \begin{tabular}{rlrlrl}
\toprule
 & & \multicolumn{2}{c}{Before Training} & \multicolumn{2}{c}{End of Fine-tuning}\\
Dataset & Starting Checkpoint & Acc & CE & Acc & CE\\
\midrule
COLA & 2000k & 0.61&0.68 &0.84&0.58 \\
 & 200k & 0.38&0.75 &0.77 & 0.79\\
 & 20k & 0.57&0.68 & 0.72 & 0.87 \\
MRPC & 2000k& 0.69&0.68  &0.85 & 0.43\\
 & 200k &0.32 &0.87 &0.83 & 0.45\\
 & 20k &  0.32&0.77 & 0.77 & 0.82 \\
QNLI & 2000k & 0.44 & 0.70 & 0.91 & 0.26 \\
 & 200k & 0.52&0.75 &0.89&0.32\\
 & 20k & 0.54&0.70 &0.84 & 0.38\\
RTE & 2000k & 0.47 & 0.70 & 0.66 & 0.86 \\
 & 200k & 0.53&0.74 &0.64 & 1.06\\
 & 20k & 0.53 & 0.70&0.64&1.13\\
SST2 & 2000k & 0.49&0.70 & 0.92 & 0.30\\
 & 200k &0.48 &0.72 &0.91&0.34\\
 & 20k & 0.51&0.70&0.88&0.48 \\
\bottomrule
\end{tabular}
\end{table}
%%%%%%%%%%%%%%%%%%%%%%%%%%%%%%%%%%%%%%%%%%%%%%%%%%%%%%%%%%%%%%%%%%%%%%%%%%%%%%%
%%%%%%%%%%%%%%%%%%%%%%%%%%%%%%%%%%%%%%%%%%%%%%%%%%%%%%%%%%%%%%%%%%%%%%%%%%%%%%%

\paragraph{ViT Fine-tuning \label{par:vit-finetune}}
We fine-tune on CIFAR-100 starting from four Vision Transformers (ViTs) \citep{dosovitskiy2021vit} on HuggingFace: \texttt{google/vit-base-patch16-224} (86M parameters),\footnote{\url{https://huggingface.co/google/vit-base-patch16-224}} \texttt{google/vit-base-patch16-224-in21k} (86M parameters),\footnote{\url{https://huggingface.co/google/vit-base-patch16-224-in21k}} \texttt{google/vit-large-patch16-224-in21k} (304M parameters),\footnote{\url{https://huggingface.co/google/vit-large-patch16-224-in21k}} and \texttt{google/vit-huge-patch14-224-in21k} (632M parameters).\footnote{\url{https://huggingface.co/google/vit-huge-patch14-224-in21k}}
All models were pre-trained on ImageNet-21k, with \texttt{vit-base-patch16-224} additionally fine-tuned on ImageNet-1k \citep{russakovsky_imagenet_2015}. 

We use the same hyperparameters across all model sizes: AdamW \citep{adamw} optimizer with learning rate $2 \times 10^{-4}$, weight decay $1 \times 10^{-4}$, batch size 32, and cosine annealing schedule with 10\% warm-up over 5 epochs. Data augmentation consisted of horizontal flips, random rotation ($\pm 10^\circ$), random translation ($\pm 16$ pixels), and cutout patches ($16\times 16$). Images are resized to $224\times 224$ to match the input resolution expected by the models. 

\paragraph{OLMo Fine-tuning}
We fine-tune OLMo-1B\footnote{\url{https://huggingface.co/allenai/OLMo-1B-hf}} on GSM8K \citep{gsm8k} starting from various checkpoints provided throughout its $\approx 740K$ pre-training steps (3 trillion tokens). For our setting, we select three checkpoints from different training phases: (1) first available checkpoint (4B tokens), (2) mid-way through pretraining (1.5T tokens), and (3) final checkpoint (3T tokens).  We fine-tuned each checkpoint for 5,000 steps using AdamW with learning rate of $2 \times 10^{-5}$ and cosine annealing with $10\%$ warm-up.

\section{Methodological Details}

In all of our experiments, we train two networks simultaneously with deterministic computations enabled, using identical random seeds for random initialization (if applicable), batch order and data augmentation.
We confirm that training with no perturbations results in exactly identical networks as expected.

All evaluations of models trained from initialization are averaged over three runs, while all evaluations for fine-tuned models listed in \cref{ap:sec:finetuning-details} are averaged over two runs.

\subsection{Computing Barriers}
\label{ap:sec:barrier}

To compute barriers, we evaluate 11 equidistant values of $\alpha \in [0, 1]$ along the linear path between $\theta_T$ and $\theta'_T$.

In our definition of barriers (\cref{eq:barrier}), we interpolate between the loss of the endpoints following \citep{sharma2024simultaneous}, rather than taking their average loss as in \citet{frankle2020linear}.
This is because the former follows from the definition of convexity and is more appropriate for describing a convex loss basin.
In practice, since $\theta_T$ and $\theta'_T$ have near-identical loss in our experiments, the definitions are interchangeable.

We also measure test error barriers by replacing $\ell$ with the 0-1 loss over test data. In practice test barriers are slightly less than train barriers as the network reach near-zero loss on the training data but not the test data, allowing for larger barriers in the former.
However, since test error barriers follow the same trends as training cross-entropy barriers, we omit them from the text.

\subsection{Computing Angular CKA}
\label{ap:sec:cka}

There are many different representational similarity methods, and a comparison of them is beyond the scope of this work.
We use CKA for a number of reasons: it is invariant to linear transformations (other than affine) which aligns well with the capacity of neural network layers \citep{kornblith2019similarity, lange2023deep}, it has been used to compare neural networks in other contexts \citep{nguyen2021wide}, is less dependent on the weighting of principal components than SVCCA \citep{raghu2017svcca}, and it can be applied to networks with more intermediate outputs $n_k$ than the number of test inputs $m$ \citep{kornblith2019similarity}. We report Angular CKA for the simple reason that it gives a distance which increases with dissimilarity, which is in concordance with the other measurements we make. Note that Angular CKA and CKA differ only in the application of arccosine.

We compute the Angular CKA between the final hidden representation as follows:
\begin{align}
\label{eq:cka}
d_{\operatorname{CKA}}(\theta_T, \theta'_T))
&= \operatorname{CKA} \left[ f_{L-1}(\theta_T), f_{L-1}(\theta'_T) \right]
\\
\operatorname{CKA}(\bf{X}, \bf{Y}) &= \arccos \left( \frac{\operatorname{HSIC}(\bf{X}, \bf{Y})}{\operatorname{HSIC}(\bf{X}, \bf{X}) \operatorname{HSIC}(\bf{Y}, \bf{Y})} \right) \notag
\end{align}
where $L$ is the number of residual or attention blocks, $f_{L-1}: \mathbb{R}^{m \times n_0} \to \mathbb{R}^{m \times n_k}$ is the last block's output on a fixed set inputs $\mathcal{X} \in \mathbb{R}^{m \times n_0}$, and $\operatorname{HSIC}$ is the Hilbert-Schmidt Independence Criterion, which measures cross correlation between centered similarity matrices.

We use the implementation by \citet{lange2023deep}, which includes certain modifications to speed computation.
Namely, we sample $m = 1000$ examples as \citet{lange2023deep} shows CKA can be reliably estimated using reasonably few examples, and we use their reduced-bias estimator for $\operatorname{HSIC}$:
\begin{align*}
\operatorname{HSIC}({\bf X, Y}) &= \frac{2}{m(m-3)}\langle \operatorname{tril}({\bf H X X^\top H}), \operatorname{tril}({\bf H Y Y^\top H}) \rangle_F
\end{align*}
where $m$ is the number of test inputs, $\operatorname{tril}$ extracts the lower triangular portion of a matrix, ${\bf H = \mathbb{I} - 11}^\top / m$ is a centering matrix that subtracts the mean, and $\langle \cdot, \cdot \rangle_F$ is the Frobenius norm.
Effectively, this estimator ignores the diagonal of the similarity matrix $\bf H X X^\top H$.

\subsection{Perturbation Scale}
\label{ap:sec:perturb-all}

To ensure fair comparisons between different perturbation methods and network architectures, we normalize all perturbations to have a consistent $L^2$ magnitude, which for ease of interpretation is given relative to the network's size at initialization (Eq. \ref{eq:expected-l2}).
Formally, we ensure the squared norm of $\noise$ matches the total variance at initialization of the perturbed weights, so that
\begin{align}
    \noise &= \frac{\hat\noise \cdot M}{\|\hat\noise \cdot M \|_2} \sqrt{Var[\theta_0 \cdot M]}
\label{eq:expected-l2}
\end{align}
where $\hat\noise$ is a batch or Gaussian perturbation sample, $M$ is a 0-1 mask of the weights to perturb, $\cdot$ is the element-wise product, $Var$ is the expected variance (\emph{not} the sample variance), and $\theta_0$ are the network's initial weights.
Thus, for example, a perturbation of magnitude $\sigma = 0.01$ is approximately 1\% of the size of $\theta_0$.

Finally, in order to preserve the distribution of activations after each normalization layer, we do not perturb biases or normalization weights in our experiments.
While prior work has found that linear mode connectivity can vary between different layers \citep{vlaar2022can, zhouGoingLinearMode2023, adilovaLayerwiseLinearMode2023}, we did not find that our results changed significantly depending on which layers were perturbed. \cref{fig:butterfly-norm} shows results when only perturbing biases and normalization weights.
In this case, we use the same scale of perturbations as we would normally assign to the weights in the layer following the biases or normalization weights.

\subsection{Linearized Approximation For $L^2$ Divergence}
\label{ap:sec:lyapunov}

A classical result of dynamical systems states that a linearized system subject to a small perturbation can diverge exponentially with respect to time at a rate which depends on the largest eigenvalue (i.e.~top Lyapunov exponent) of the gradient of the training map \citep{strogatzNonlinearDynamicsChaos2019}:
\begin{align}
    \tmap(\theta_i + \noise) &\approx \tmap(\theta_i) + \noise^\top \nabla_{\tmap}  \theta_i ,  \notag \\
    \tmap^T(\theta_0 + \noise) &\approx \tmap^T(\theta_0) + \noise^\top 
    \prod_{t=1}^T \nabla_{\tmap} \theta_0 \notag \\
     \| \theta_T - \theta'_T \|_2 &\leq \|\noise\|_2 e^{\lambda t}
\end{align}
where $\nabla_\tmap \theta_0$ is the gradient of the training map with respect to the weights and $\lambda$ is the top eigenvalue over all the gradients at each step $t$.

Substituting in the definition of SGD, we find that the $L^2$ divergence between the original and perturbed models after training depends on the curvature of the loss landscape:
\begin{align}
    \tmap(\theta_i) &= \theta_i - \eta \nabla_\ell \theta_i, \qquad
    \nabla_\tmap \theta_i = I - \eta_i H_i , \notag \\
     &\| \theta_T - \theta'_T \|_2 \leq \|\noise\|_2 e^{\lambda_H t}  
\end{align}
where $\eta_i$ is the learning rate, $H_i$ is the Hessian of the weights, and $\lambda_H$ is the largest eigenvalue over all $I - \eta_i H_i$.
Divergence results either from high positive curvature as in \citet{wu2018sgd}, but additionally if there is any negative curvature in the perturbation direction, and $|\lambda_H| > 1$ implies the possibility of exponential growth in divergence over training.

\section{Further Experiments}

\subsection{Baselines}

\begin{figure}[ht]
\vskip 0.1in
\begin{center}
\centerline{
    \includegraphics[width=0.24\columnwidth]{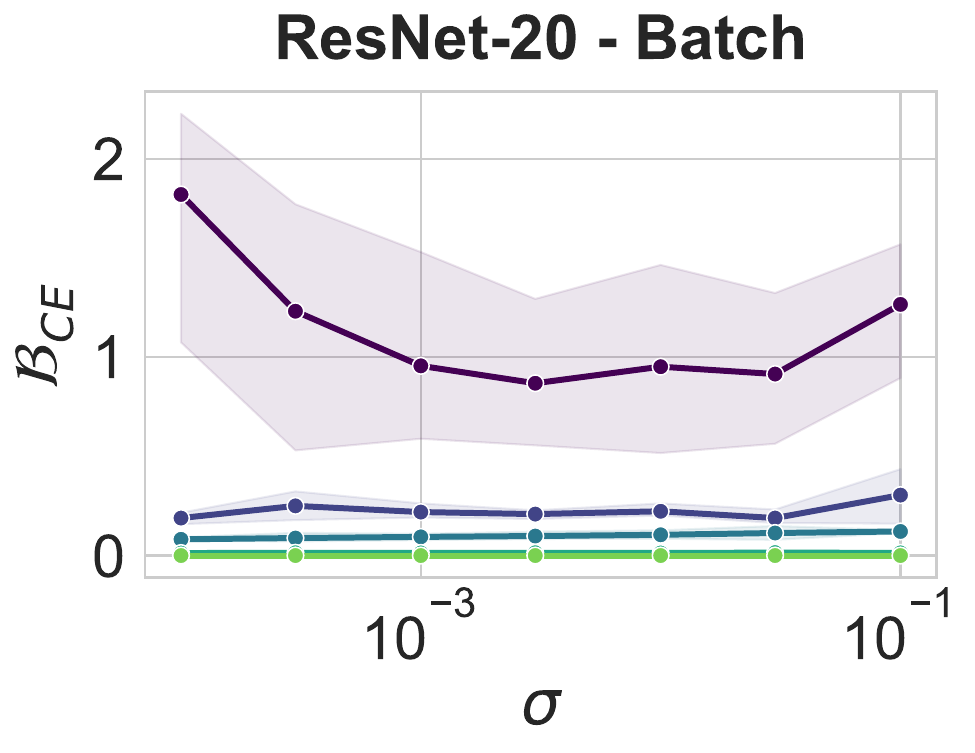}
    \includegraphics[width=0.24\columnwidth]{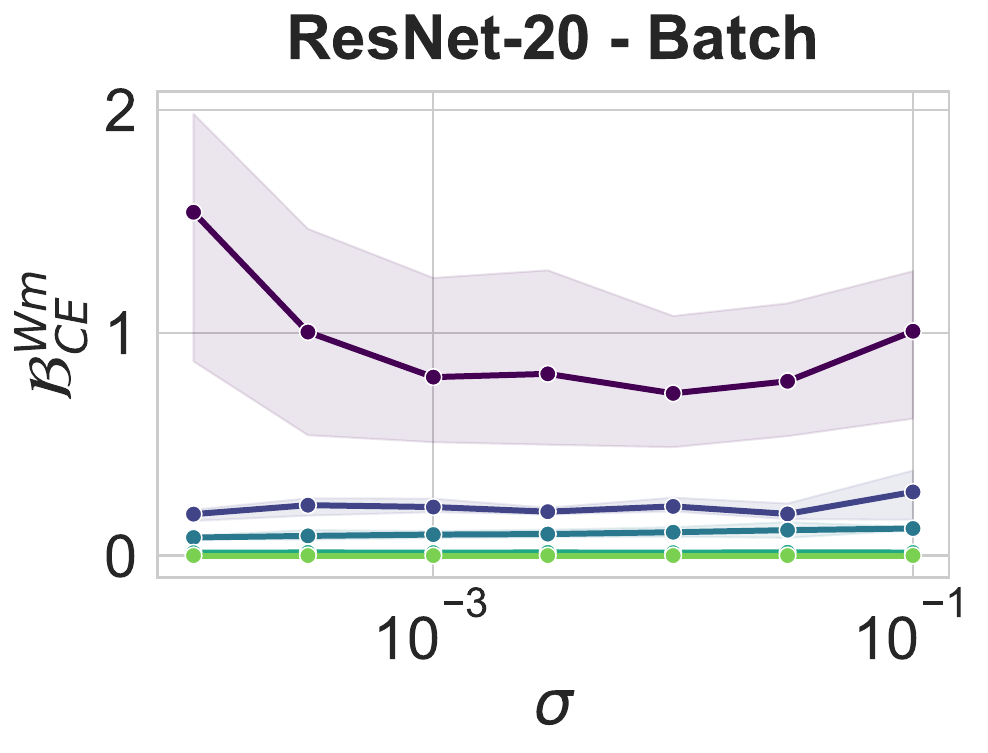}
    \includegraphics[width=0.24\columnwidth]{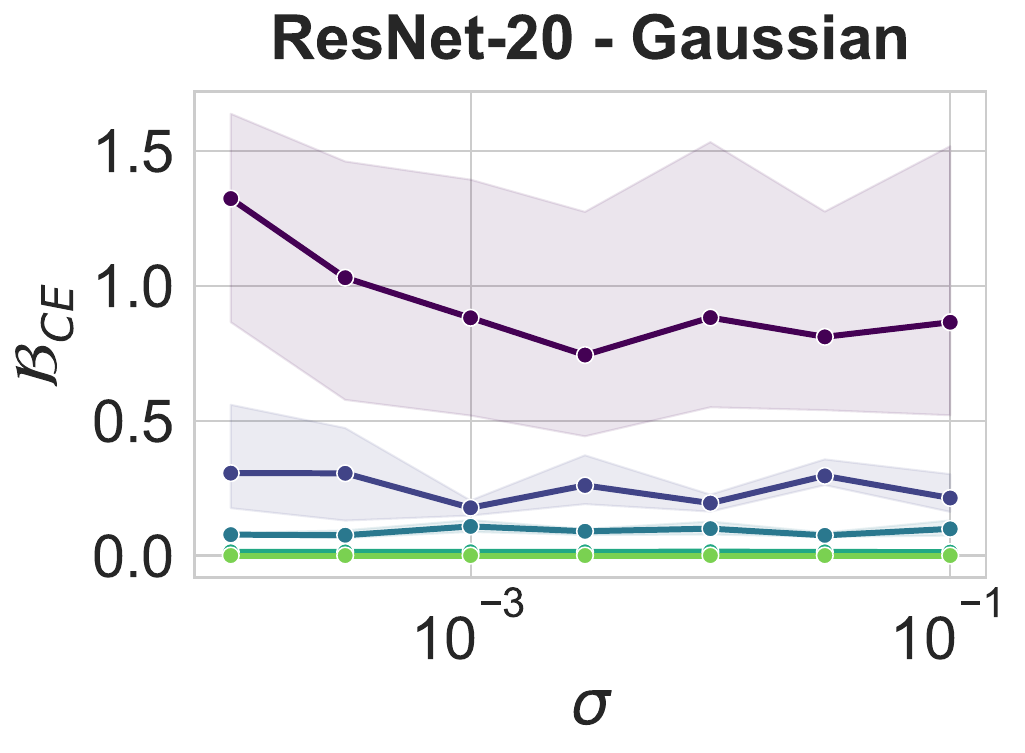}
    \includegraphics[width=0.24\columnwidth]{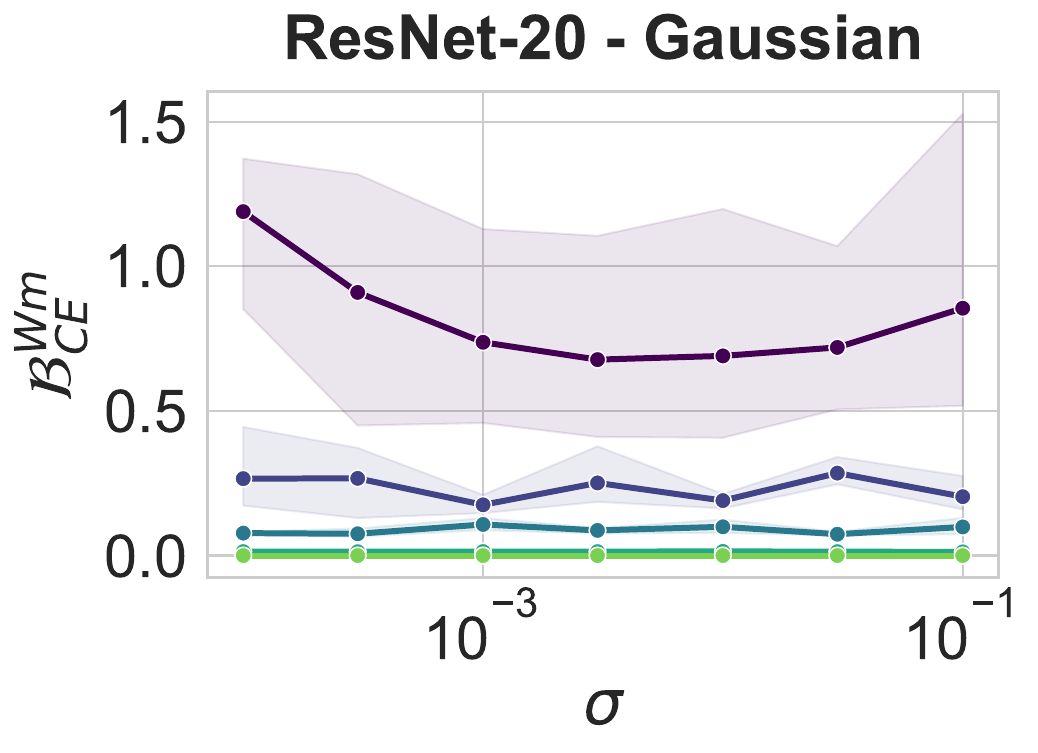}
}
\centerline{
    \includegraphics[width=0.6\columnwidth]{figures/butterfly-hparams/no-decay-sanity-batch-lmc-0-1-loss-weighted-barrier-legend.pdf}
}
\caption
{
Train loss barriers before and after permutations when perturbing only normalization layers. Results are shown for batch (left) and Gaussian (right) perturbations on ResNet-20 trained with SGD (momentum, no weight decay), using a learning rate of 0.1, $2\%$ warm-up, and a batch size of 128 for 20,000 steps.
}
\label{fig:butterfly-norm}
\end{center}
\vskip -0.2in
\end{figure}

\emph{Stability is not specific to layer type.} In the main text (\cref{fig:butterfly-time}), we excluded norm weights and biases from perturbations. However, in \cref{fig:butterfly-norm}, we show that perturbing only norm layers leads to similar trends. This suggests that fine-tuning stability is influenced more by overall network dynamics rather than specific layer types. While individual parameters or layers may have varying importance as noted in previous work \citep{adilovaLayerwiseLinearMode2023}, batch perturbations already capture this effect to some extent.
The increased barriers for the smallest perturbations at initialization are an artifact of numerical instability---in our regular experiments, we avoid this problem by reducing the fraction of perturbed weights instead of reducing the perturbation scale beyond $\sigma = 10^{-4}$.

\begin{figure}[ht]
\vskip 0.1in
\begin{center}
\centerline{
    \includegraphics[width=0.33\linewidth]{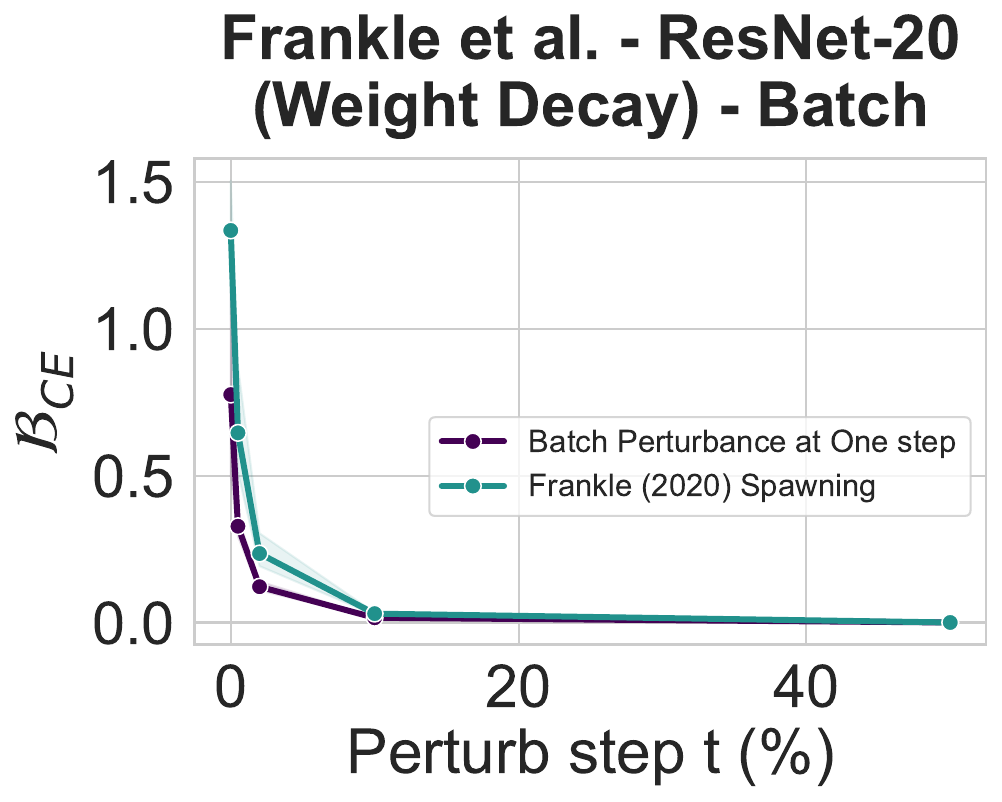}
    \includegraphics[width=0.33\linewidth]{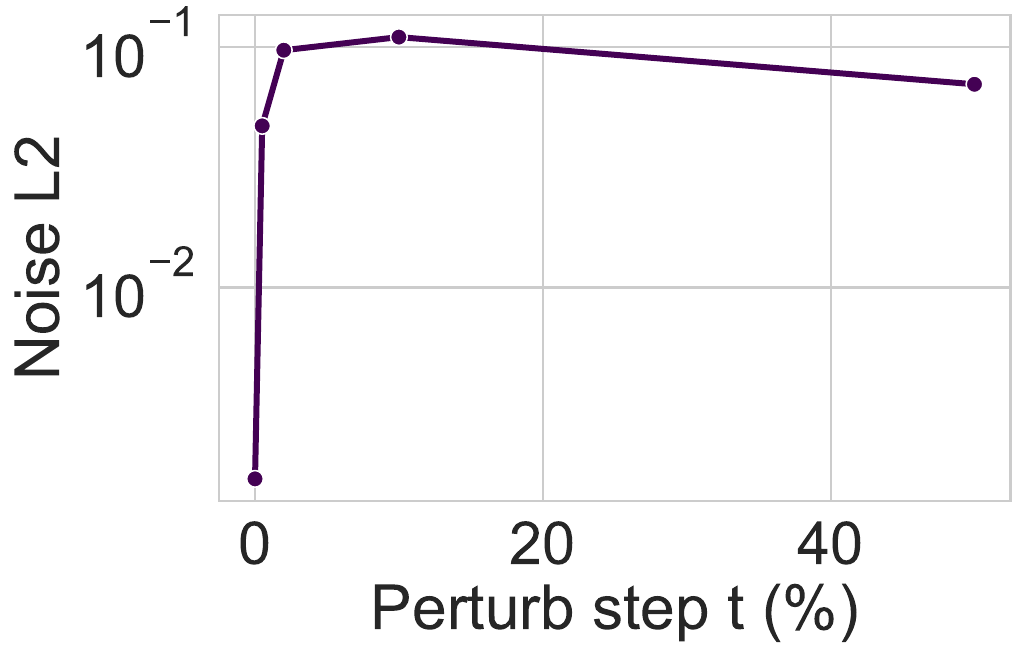}
    }
    \caption{\textbf{Left:} Comparison of training loss barriers between our butterfly setting and \citeauthor{frankle2020linear}'s spawning setting. In the spawning setting, each network is trained with different non-determinism after perturbation step $t$, while our method applies a single perturbation. As expected, our single-perturbation approach provides a lower bound on the spawning barriers. \textbf{Right:} $L^2$ magnitude of the expected deviation between two copies of the same model, when each model takes a single independent training step at time $t$.}
    \label{fig:perturb-frankle}
\end{center}
\vskip -0.2in
\end{figure}

\emph{How do our perturbations compare to SGD noise? } To establish the relative magnitude of our perturbations vs. SGD noise, we replicate the parent-child spawning experiment of \citeauthor{frankle2020linear}. \cref{fig:perturb-frankle} shows that our batch perturbations are a lower bound on the Frankle baseline's barriers, meaning that the instability resulting from training independently for multiple steps must be at least the instability resulting from a single independent training step.

Note that for this comparison, we scale batch perturbations to the expected magnitude of SGD noise at the perturbation time $t$.
This makes batch perturbation equivalent to taking only one step at time $t$ with different SGD noise, as opposed to using different SGD noise from $t$ onwards in the Frankle baseline.

\subsection{Perturbing Only A Fraction of Weights.} 
\begin{figure}[ht]
\vskip 0.1in
\begin{center}
\centerline{
    \includegraphics[width=0.24\linewidth]{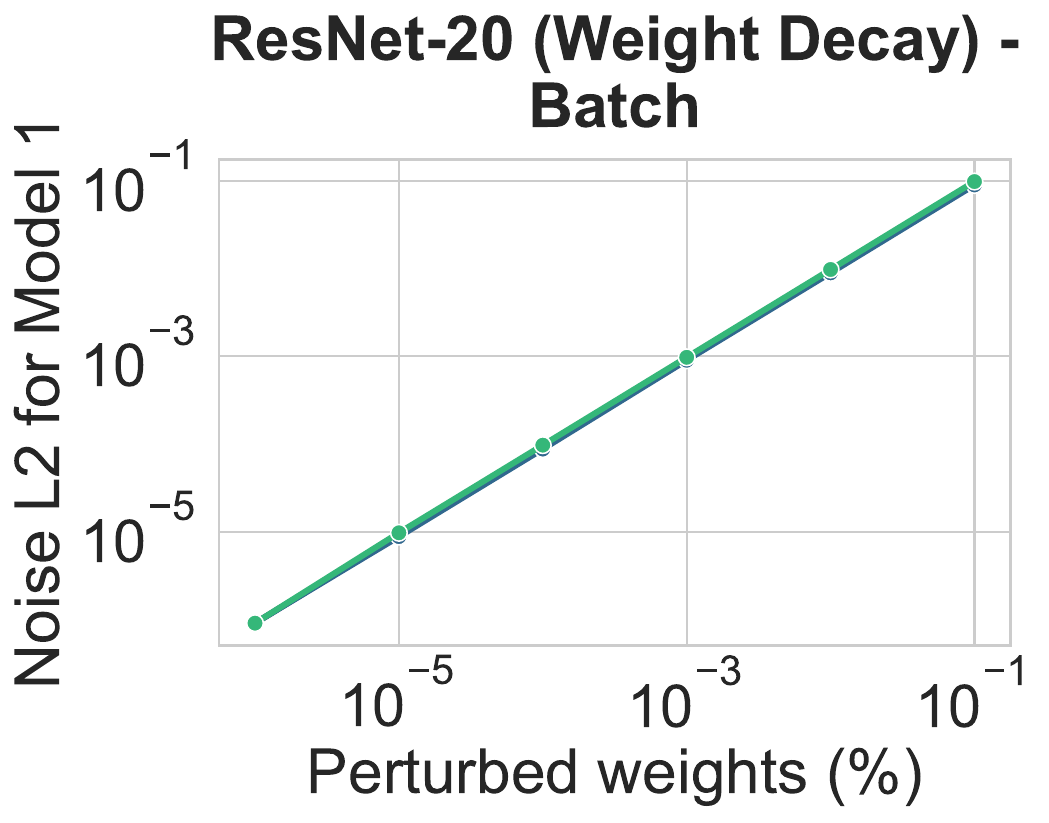}
    \includegraphics[width=0.24\linewidth]{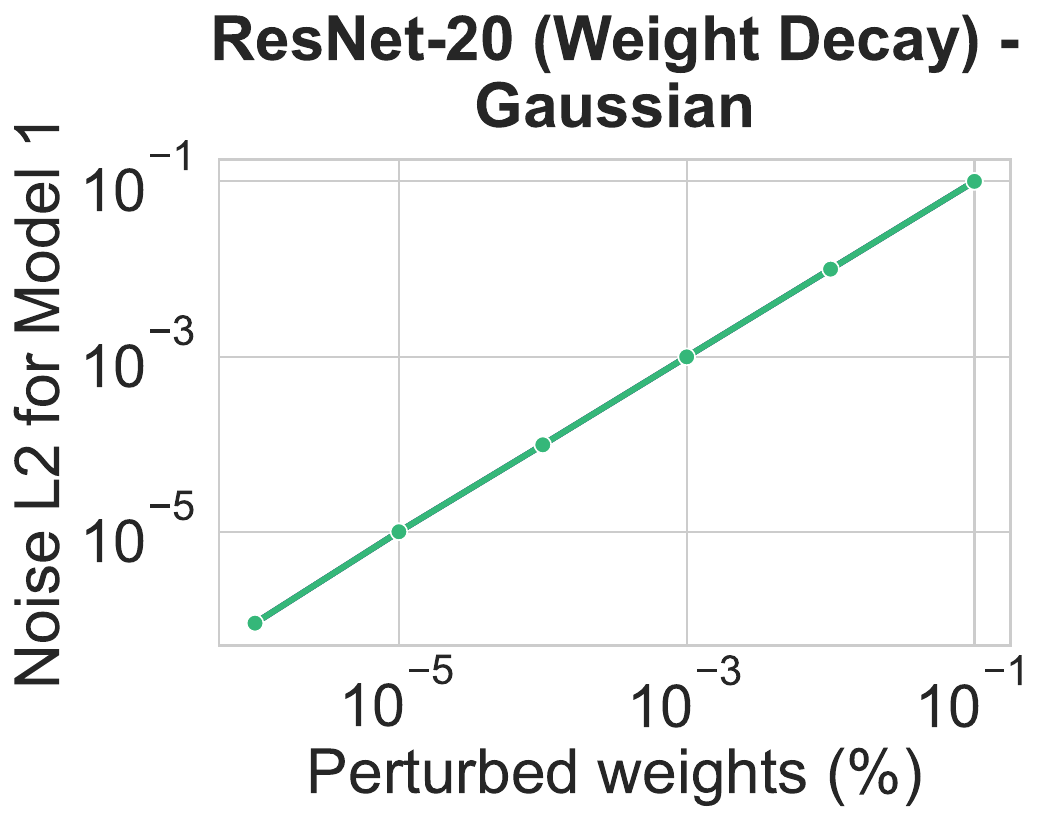}
    \includegraphics[width=0.24\linewidth]{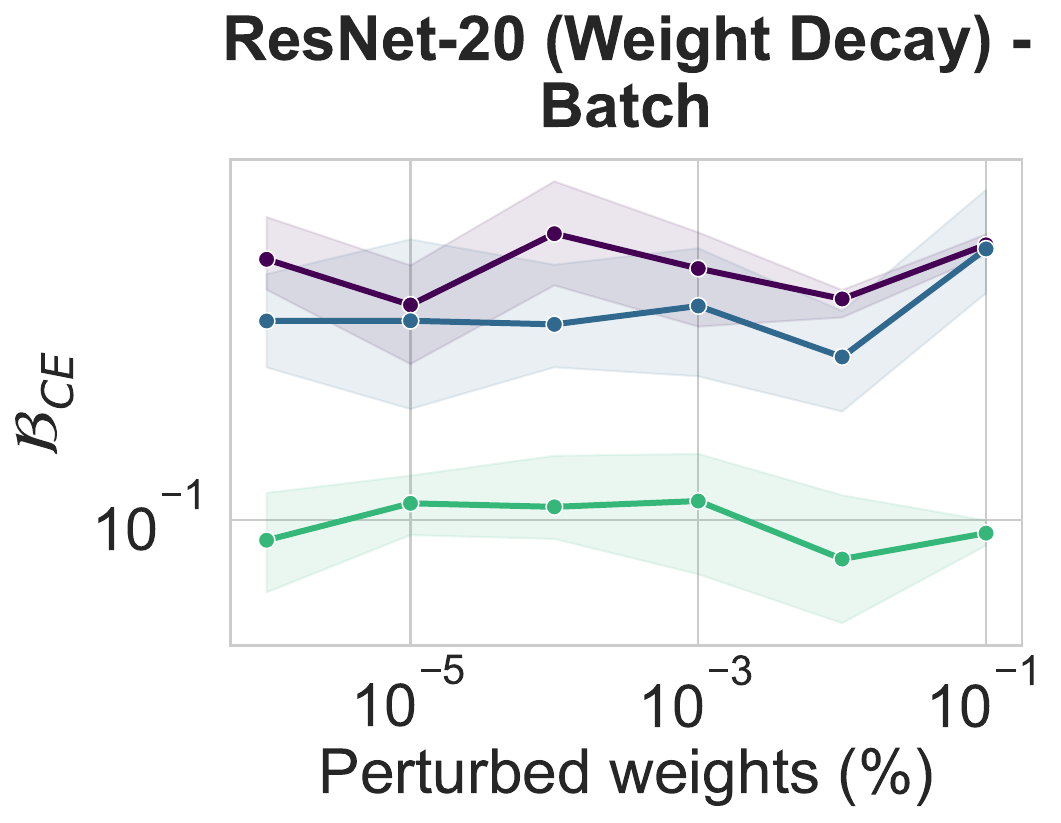}
    \includegraphics[width=0.24\linewidth]{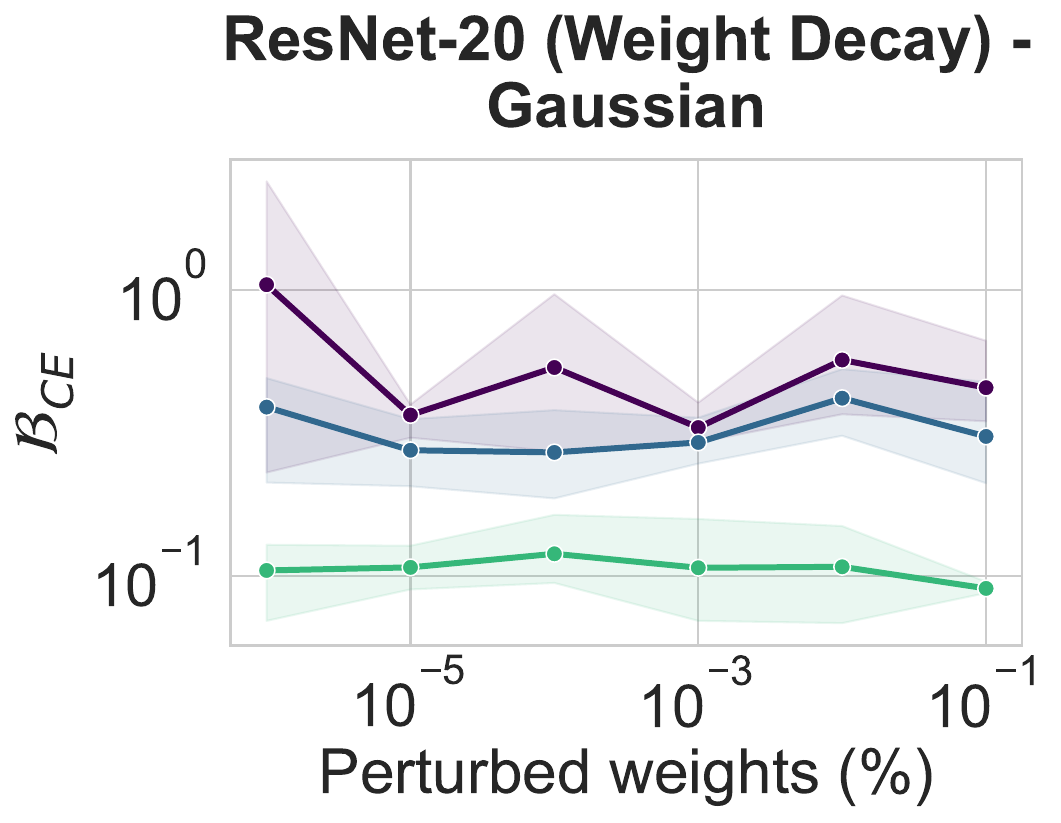}
    }
\centerline{
    \includegraphics[width=0.45\linewidth]{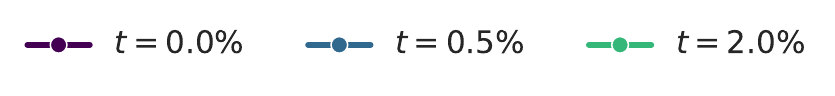}
    }
\vskip -0.2in
    \caption{\textbf{Left:} absolute $L^2$ norm of the noise as a function of the fraction of perturbed weights. \textbf{Right:} train loss barriers as a function of the fraction of perturbed weights.}
    \label{fig:perturb-fraction}
\end{center}
\end{figure}

We further decrease the scale of our perturbations from \cref{fig:butterfly-time} by perturbing only a fraction of the weights with our smallest perturbation scale of $10^{-4}$. Strikingly, we find that perturbing as little as a \textbf{single weight}, which occurs when the fraction of perturbed weights is $10^{-6}$, is sufficient to create barriers at initialization (\cref{fig:perturb-fraction}, right). The scale of this perturbation (\cref{fig:perturb-fraction}, left) is well below that of noise caused by hardware indeterminacy.

\subsection{Additional Hyperparameter Settings} 

\begin{figure}[ht]
\vskip 0.1in
\begin{center}
\centerline{
    \includegraphics[height=0.23\columnwidth]{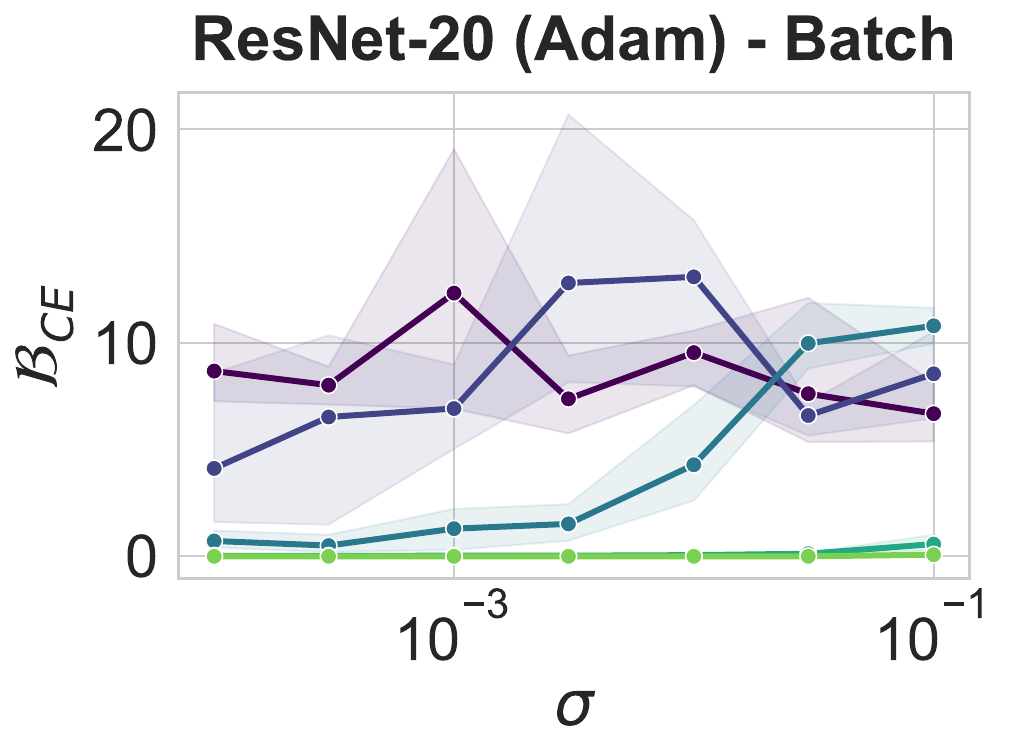}
    \includegraphics[height=0.23\columnwidth]{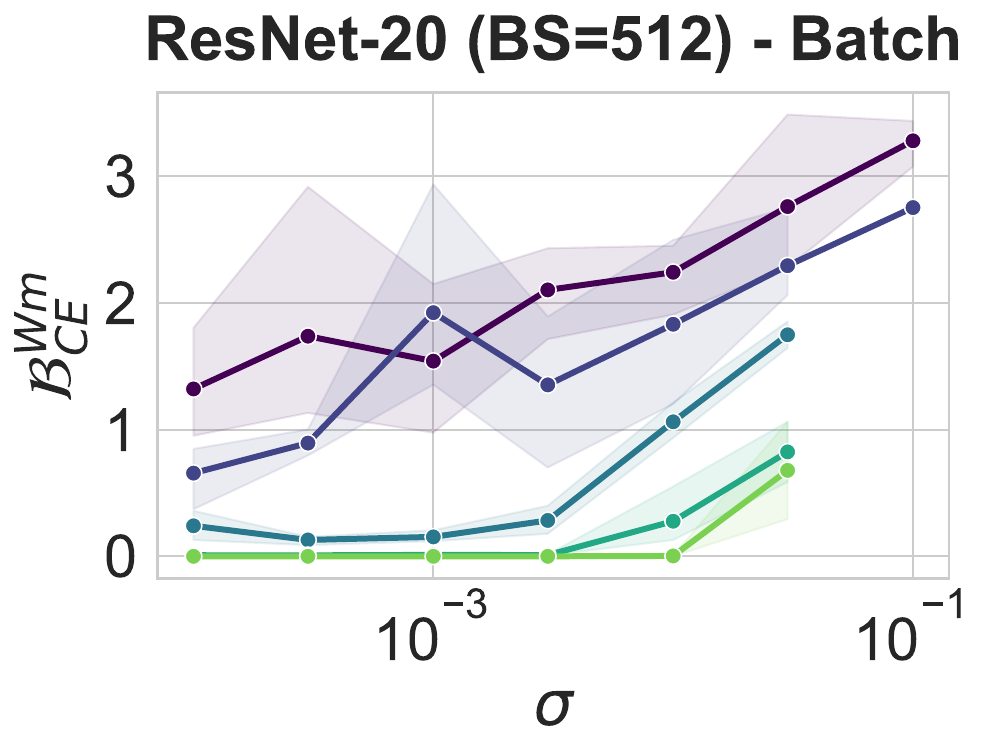}

    \includegraphics[height=0.23\columnwidth]{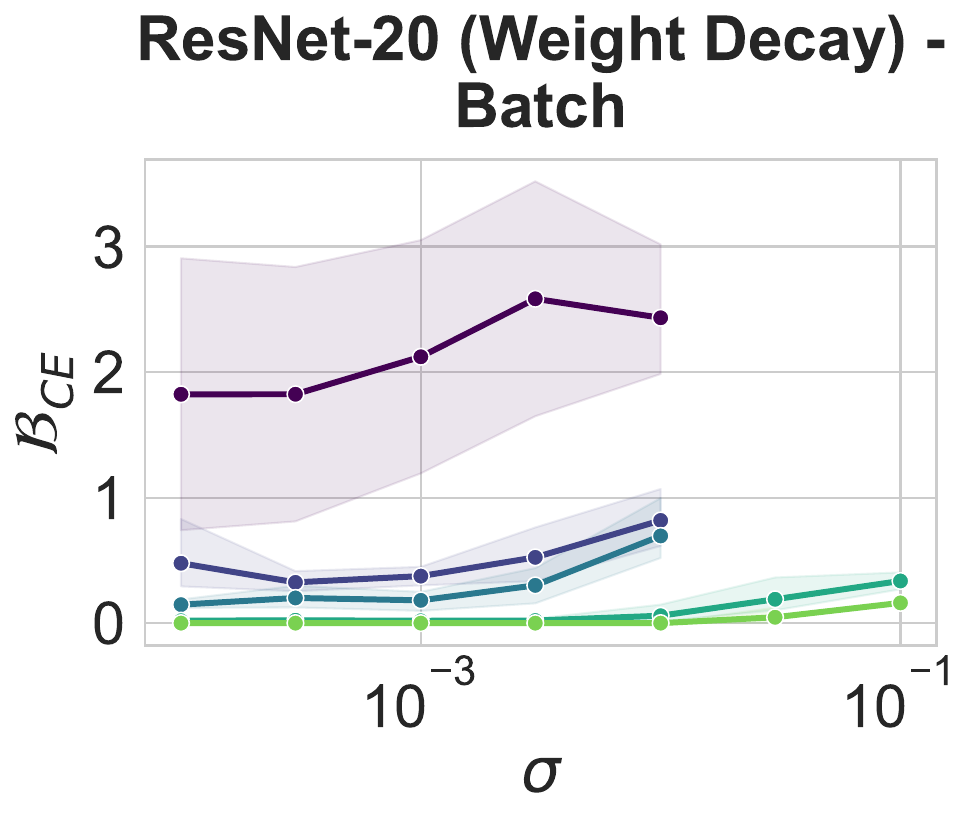}
}
\centerline{
    \includegraphics[width=0.66\columnwidth]{figures/butterfly-hparams/no-decay-sanity-batch-lmc-0-1-loss-weighted-barrier-legend.pdf}
}
\vskip -0.2in
\caption
{ 
Same as \cref{fig:butterfly-time} with AdamW without weight decay and learning rate of 0.003 (\textbf{left}), batch size of 512 (\textbf{middle}), and weight decay (\textbf{right}).
}
\label{fig:butterfly-bs-512}
\end{center}
\end{figure}

\label{ap:sec:butterfly-hparams}
In this section we present the results of training ResNet-20 on CIFAR-10 for all hyperparameter combinations listed in \cref{ap:tab:CIFAR10hparams}.

It is well-known that neural network training is highly sensitive to optimization hyperparameters \citep{smithDisciplinedApproachNeural2018}.
Our experiments corraborate the complex interdependencies between key optimization hyperparameters, suggesting that further theoretical exploration is needed. We specifically focus on the impact of optimizer choice, batch size, and weight decay.

\emph{SGD enhances training stability.} The choice of optimizer significantly impacts the stability of the training map. Additionally, networks trained with Adam also exhibit higher $L^2$ distances (\cref{fig:l2-barriers}). 
We stipulate that this phenomenon is linked to the implicit bias of SGD. Specifically, \citet{bradleyShiftCurvatureSGDGeneralization2022} highlight that SGD's inherent noise helps it avoid high-curvature regions, which are often linked to poor generalization and, in our setting, could likely contain linearly connected minima.

In our experiments, we observed that larger batch sizes lead to higher loss barriers, suggesting a trade-off between batch size and stability. This phenomenon aligns with findings from \citet{keskarLargeBatchTrainingDeep2017}, who show that larger batch sizes can lead to sharper minima, which might increase the loss barriers and potentially hurt generalization. 

Weight decay provides slightly more stability, particularly right after initialization.
\citet{dangeloWhyWeNeed2024} highlight that weight decay modifies the optimization dynamics, enhancing the implicit regularization of SGD through loss stabilization mechanisms. They also emphasize the role of gradient clipping, which stabilizes training, particularly at the edge of stability.

\subsection{Functional Diversity For Additional Hyperparameter Settings}

\begin{figure*}[ht]
\vskip 0.1in
\begin{center}
\centerline{
    \includegraphics[height=0.23\linewidth]{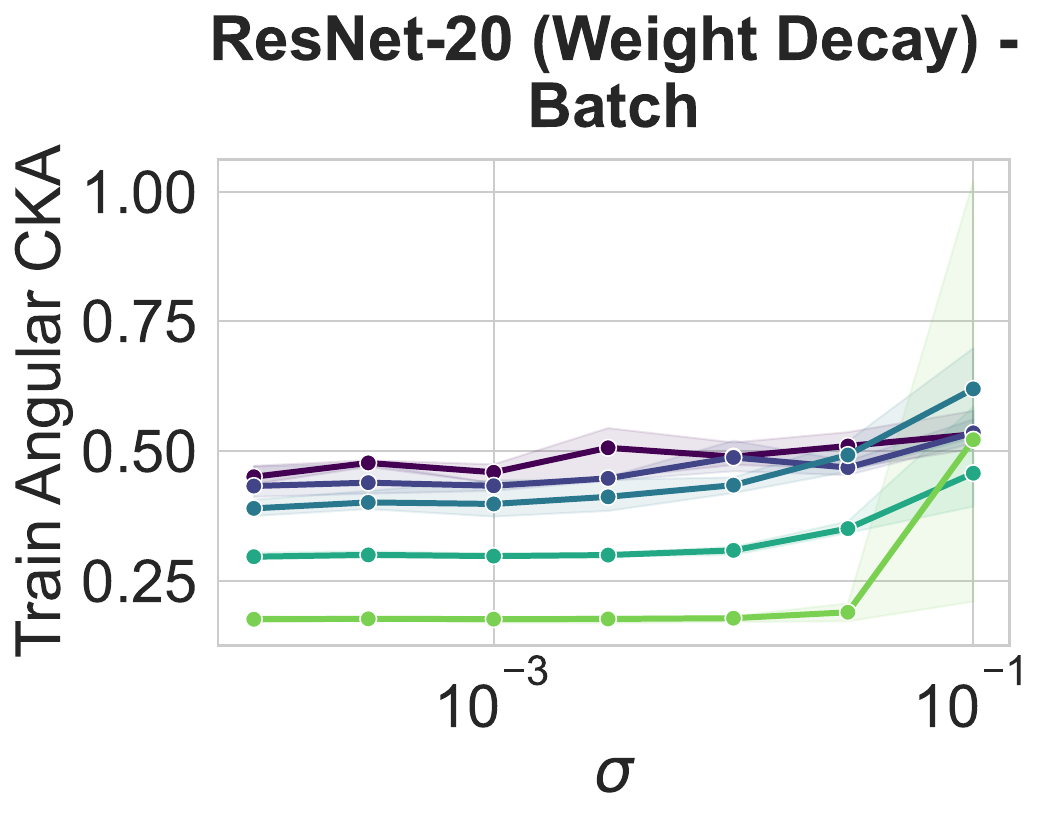}
    \includegraphics[height=0.23\linewidth]{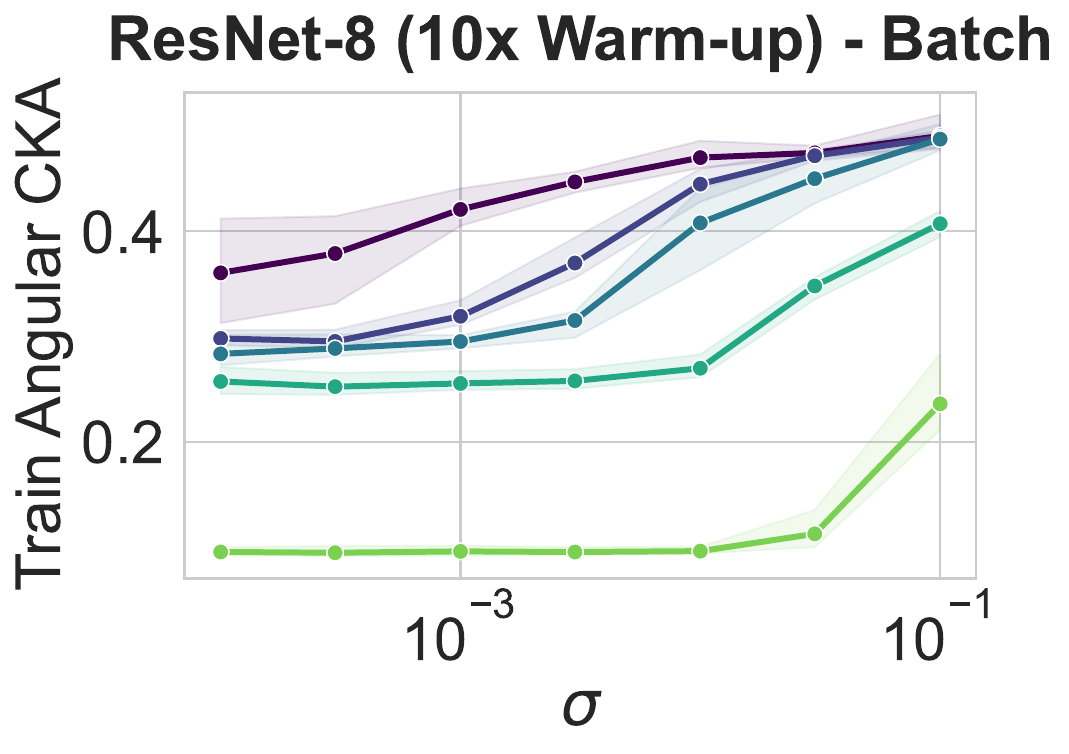}
}
\centerline{
    \includegraphics[width=0.4\linewidth]{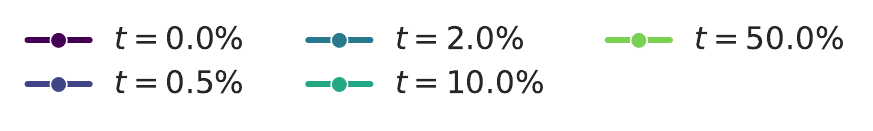}
}
\vskip -0.2in
\caption{Representational similarity distance measured via Angular CKA for ResNet-20 with weight decay (\textbf{left}) or 10x warm-up (\textbf{right}).
}
\label{fig:butterfly-cka-additional}
\end{center}
\vskip -0.2in
\end{figure*}

\begin{figure*}[ht]
\vskip 0.1in
\begin{center}
\centerline{
    \includegraphics[height=0.23\linewidth]{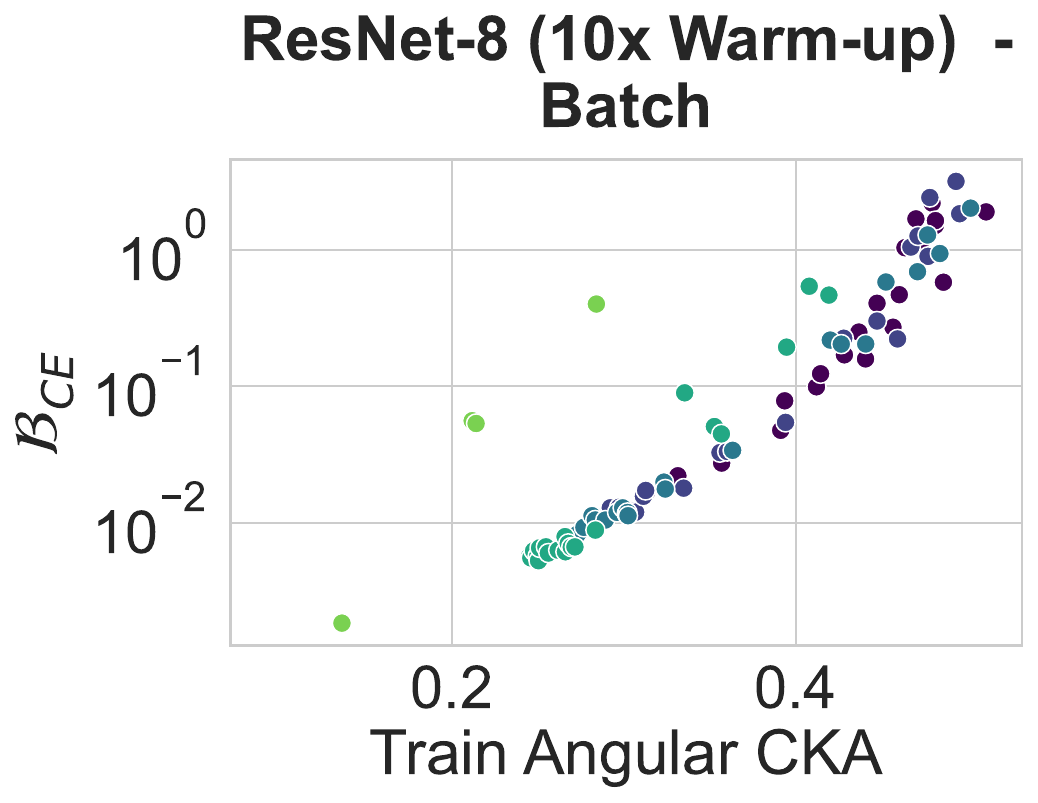}
    \includegraphics[height=0.23\linewidth]{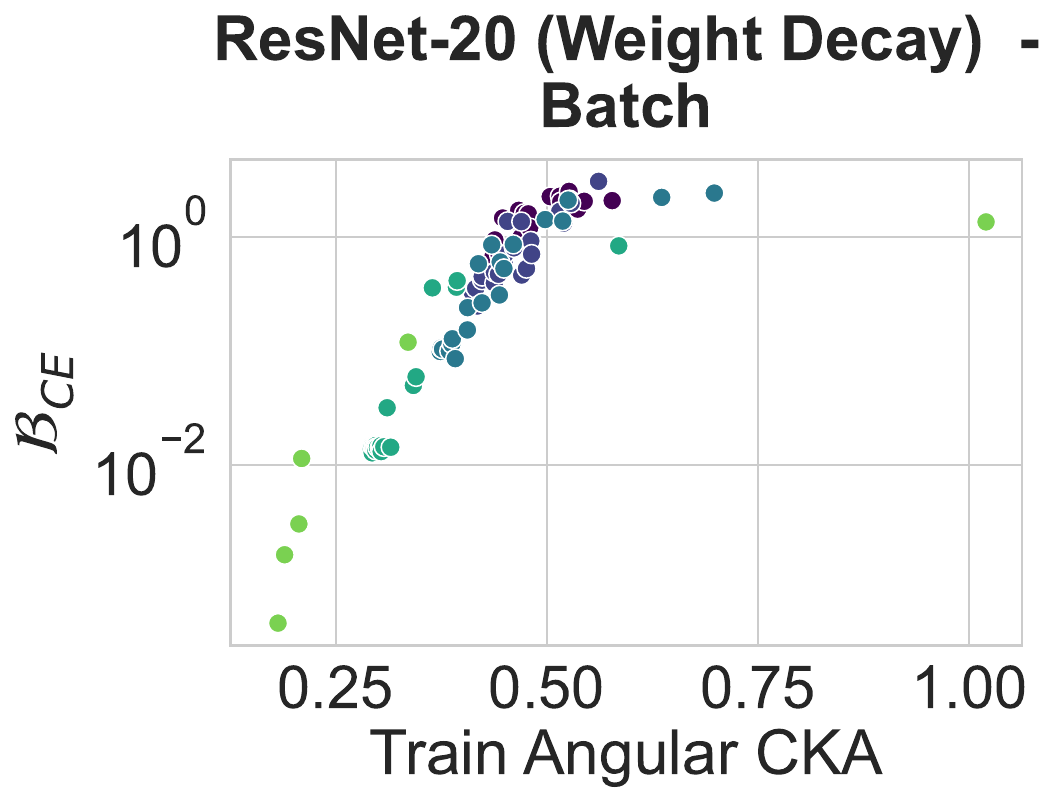}
    \includegraphics[height=0.23\linewidth]{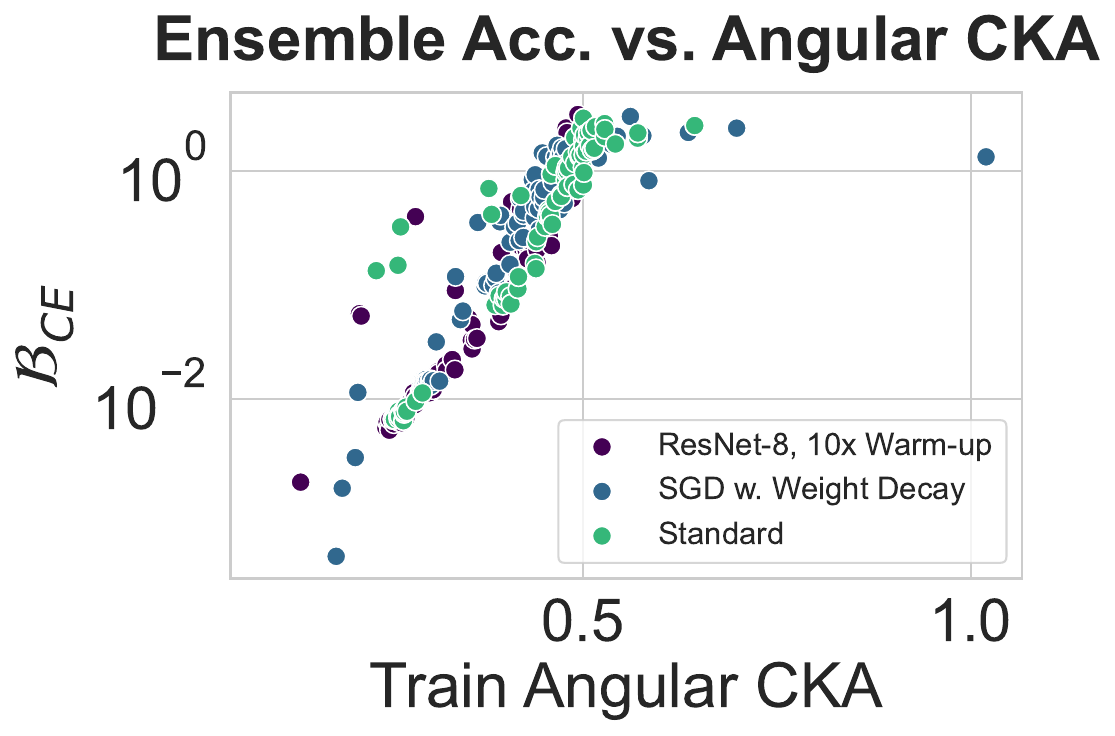}
}
\vskip -0.05in
\centerline{
    \includegraphics[width=0.6\linewidth]{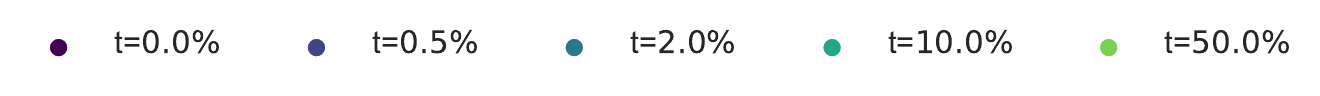}
}
\vskip -0.2in
\caption{Train loss barrier versus representational similarity distance on ResNet-20 with weight decay (\textbf{left}), ResNet-8 with 10x warm-up (\textbf{middle}) with perturbation time indicated by color, and including the standard setting from \cref{fig:butterfly-cka}, with settings indicated by color (\textbf{right)}.
}
\label{fig:butterfly-cka-additional-barriers}
\end{center}
\end{figure*}

\begin{figure*}[ht]
\vskip 0.1in
\begin{center}
\centerline{
    \includegraphics[height=0.23\linewidth]{figures/rebuttal/cka/cka-new/ensemble-acc-no-decay-cka-batch-ensemble_acc_test.pdf}
    \includegraphics[height=0.23\linewidth]{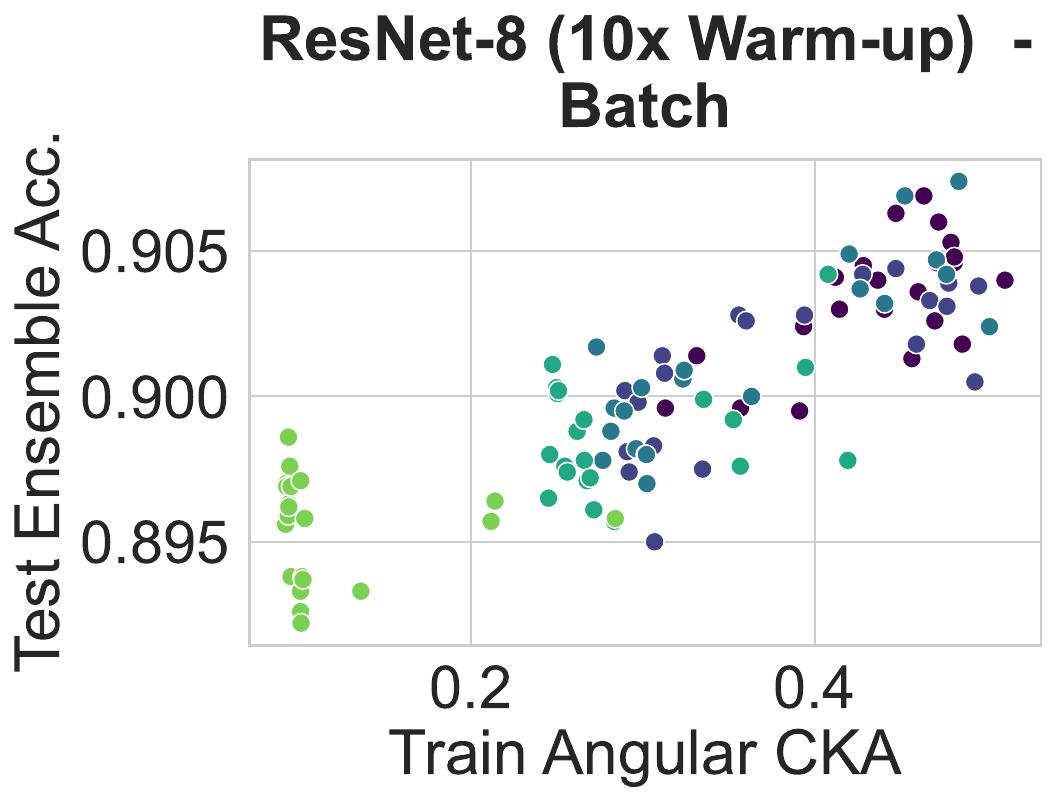}
    \includegraphics[height=0.23\linewidth]{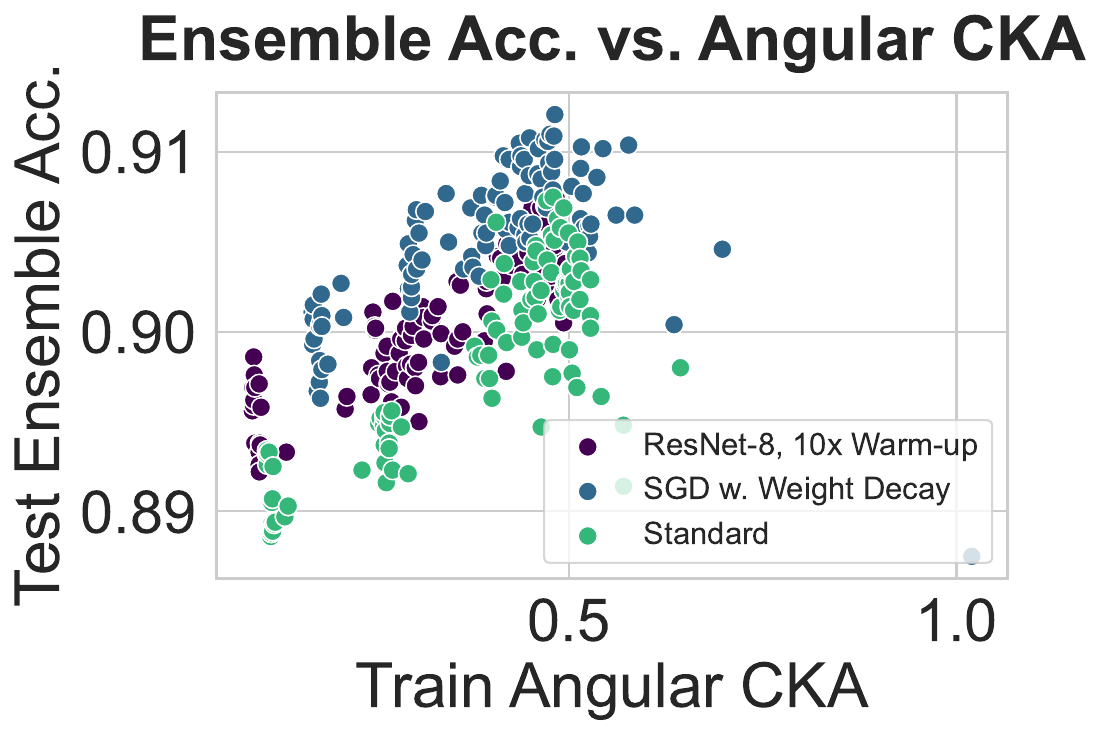}
}
\centerline{
    \includegraphics[width=0.6\linewidth]{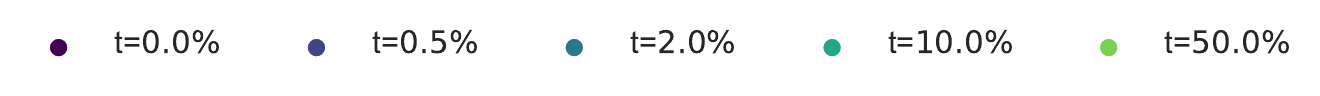}
}
\vskip -0.2in
\caption{Test accuracy of an ensemble of the original and perturbed models after training, versus representational similarity distance, on ResNet-20 with weight decay (\textbf{left}), ResNet-8 with 10x warm-up (\textbf{middle}) with perturbation time indicated by color, nd including the standard setting from \cref{fig:butterfly-cka}, with settings indicated by color (\textbf{right)}.
}
\label{fig:butterfly-cka-additional-ensemble}
\end{center}
\vskip -0.2in
\end{figure*}

We replicate \cref{sec:functional-diversity} on two of our additional hyperparameter settings: with weight decay, and with 10x warm-up.
The results in \cref{fig:butterfly-cka-additional,fig:butterfly-cka-additional-barriers,fig:butterfly-cka-additional-ensemble} closely agree with our findings in \cref{fig:butterfly-cka}.

\newpage
\subsection{Fixed Points of Aligning Permutations}
While weight matching is unable to reduce barriers in our case of \emph{identically} initialized networks, we investigate whether the underlying mechanism proposed by \citep{entezari2021role}---that barriers arise from network permutations---is still relevant to our observations.
We consider if the observed barriers and $L^2$ distances between the original and perturbed networks correlate with the number of fixed points in the permutations found by weight alignment \citep{ainsworth2022git}.
Here, fixed points refer to the un-permuted elements in the aligning permutations.
Since we expect two networks with identical weights to be aligned by permutations consisting only of fixed points, the fraction of fixed points can be used to indicate the degree to which two diverging networks have been permuted with respect to one another.

\begin{figure}[ht]
\vskip 0.1in
\begin{center}
\centerline{
    \includegraphics[width=0.32\columnwidth]{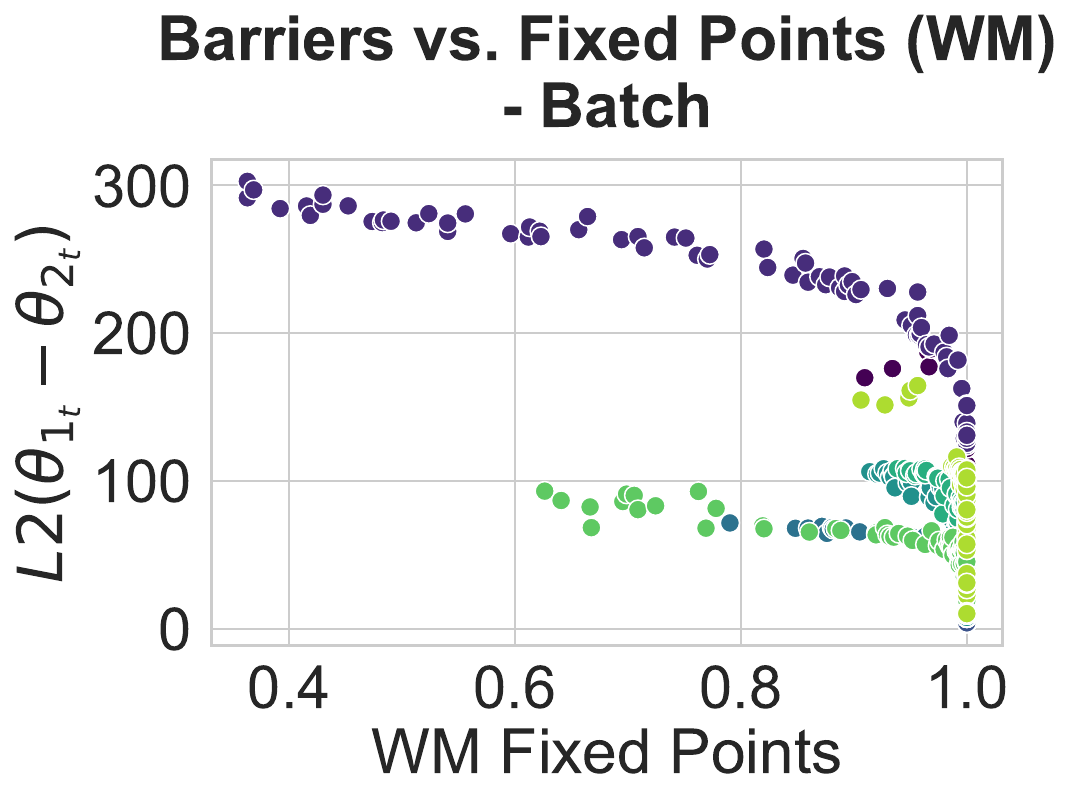}
    \includegraphics[width=0.32\columnwidth]{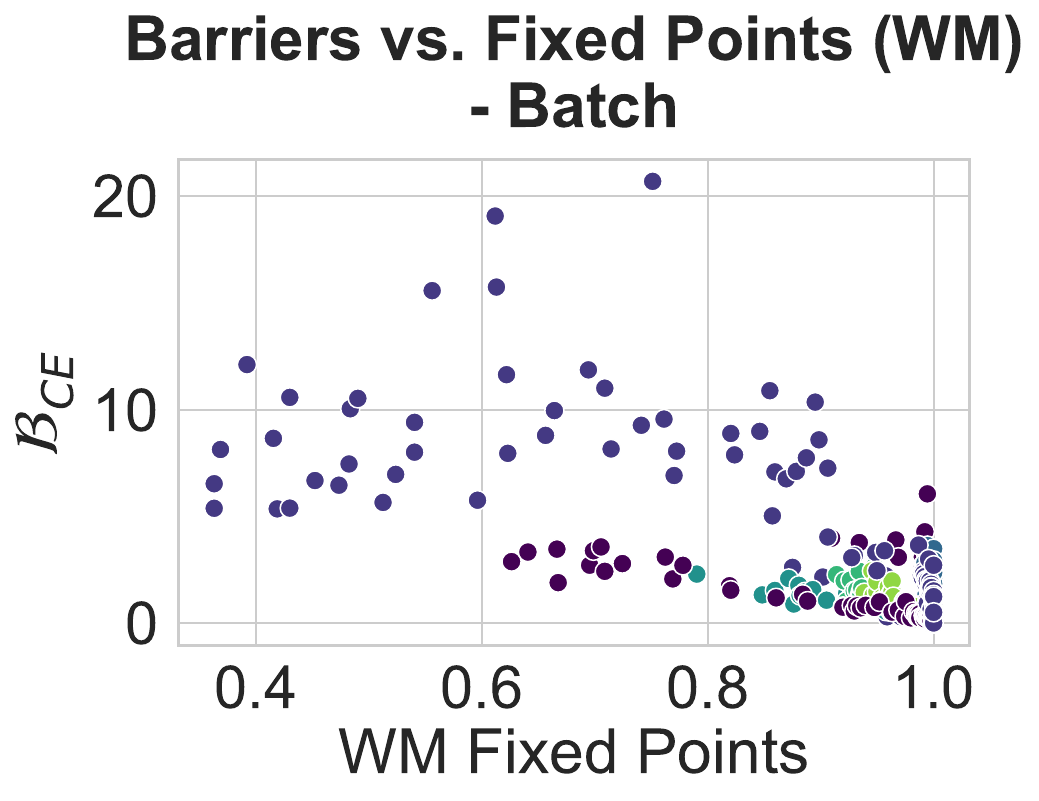}
}
\centerline{
    \includegraphics[width=0.49\columnwidth]{figures/scatter/fixed-pts-barr-group-all-batch-lmc-0-1-loss-weighted-barrier-legend.pdf}
}
\vskip -0.1in
\caption{Relationship between the fraction of fixed points (un-permuted elements) in the weight matching permutation aligning the ResNet-20 models trained on CIFAR-10 from \cref{fig:butterfly-time}, and $L^2$ divergence (\textbf{left}) or train barriers (\textbf{right}).
}
\label{ap:fig:butterfly-fixed-barriers}
\end{center}
\vskip -0.1in
\end{figure}

\cref{ap:fig:butterfly-fixed-barriers} suggests a weak correlation, where the number of fixed points is inversely proportional to both the
barrier heights and the $L^2$ distance between the networks before alignment in the ResNet settings.
Unfortunately, this observation does not extend to our ViT or BERT settings, as weight matching fails to identify non-trivial permutations (i.e., permutations other than the identity) for these transformer architectures.

\section{Further Fine-tuning Results}
\subsection{ResNet Fine-Tuning}

\begin{figure}[ht]
\begin{center}
\centerline{
    \includegraphics[width=0.3\linewidth]{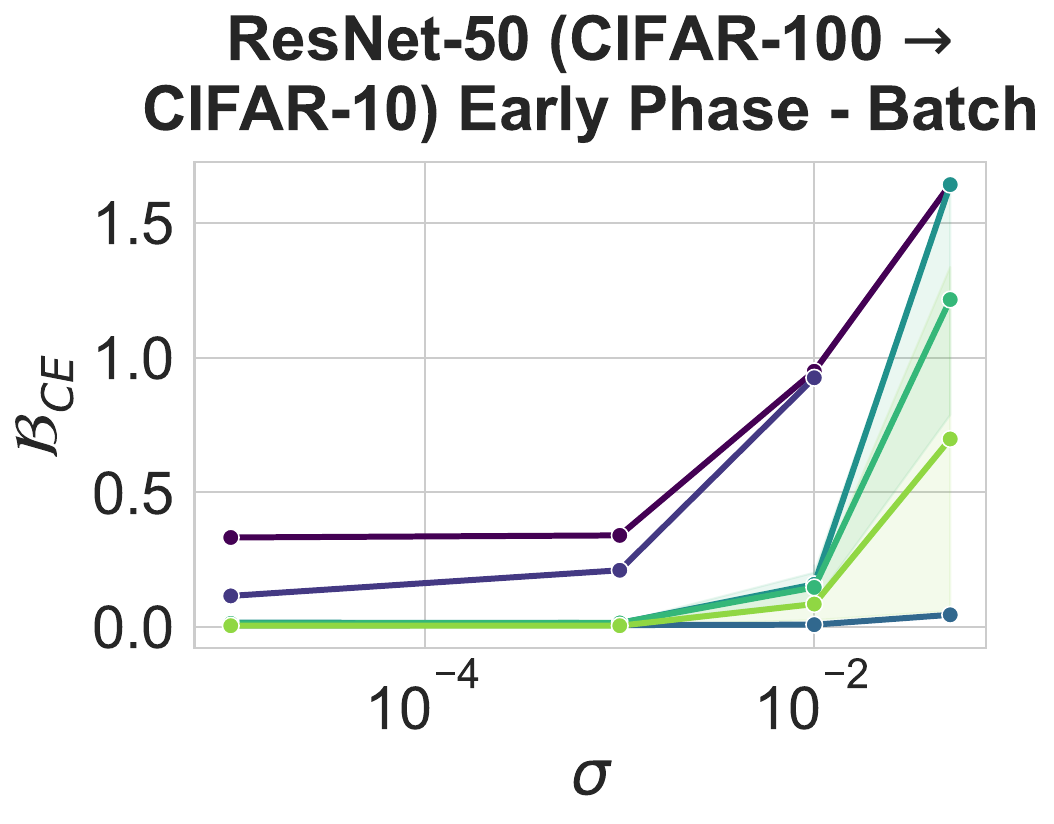}
    \includegraphics[width=0.3\linewidth]{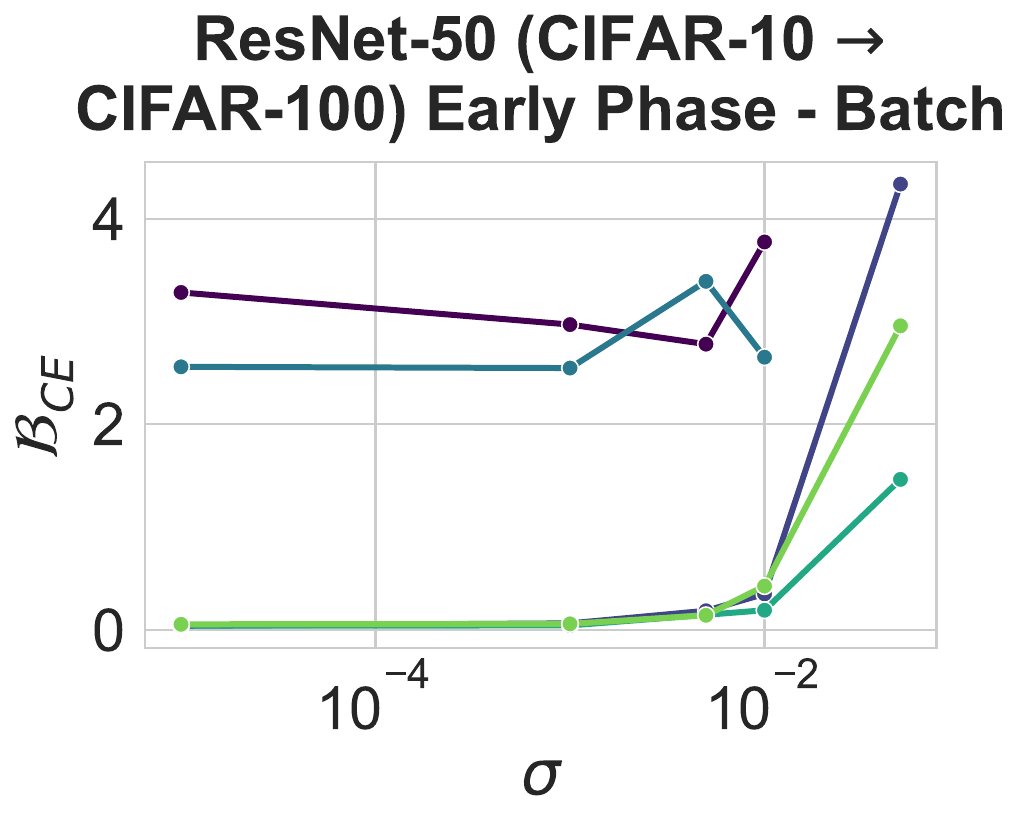}
}
\centerline{
    \includegraphics[width=0.5\linewidth]{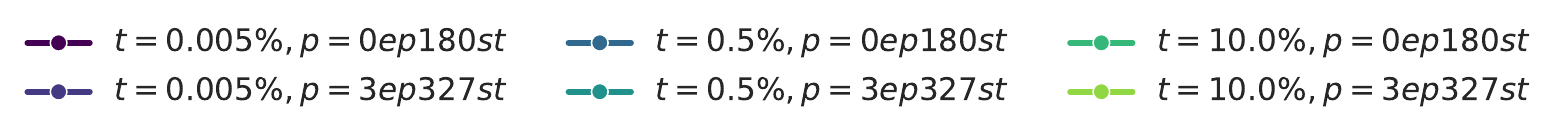}
}
\vskip -0.1in
\caption{
Same as \cref{fig:fine-tuning} (left), but with additional pre-training times and perturbation times from early in training.
Stability of transfer learning on vision tasks: a ResNet-50 is pre-trained and fine-tuned (see \cref{ap:sec:finetuning-details} for details) from CIFAR-100 to CIFAR-10 (\textbf{left}) or vice versa (\textbf{right}). Barriers (y-axis) are plotted against perturbation magnitudes (x-axis) for various combinations of initial pre-trained weights and perturbation times (colors). 
}
\label{ap:fig:fine-tuning:across-task-early}
\end{center}
\end{figure}

\begin{figure}[ht]
\begin{center}
\centerline{
    \includegraphics[width=0.3\linewidth]{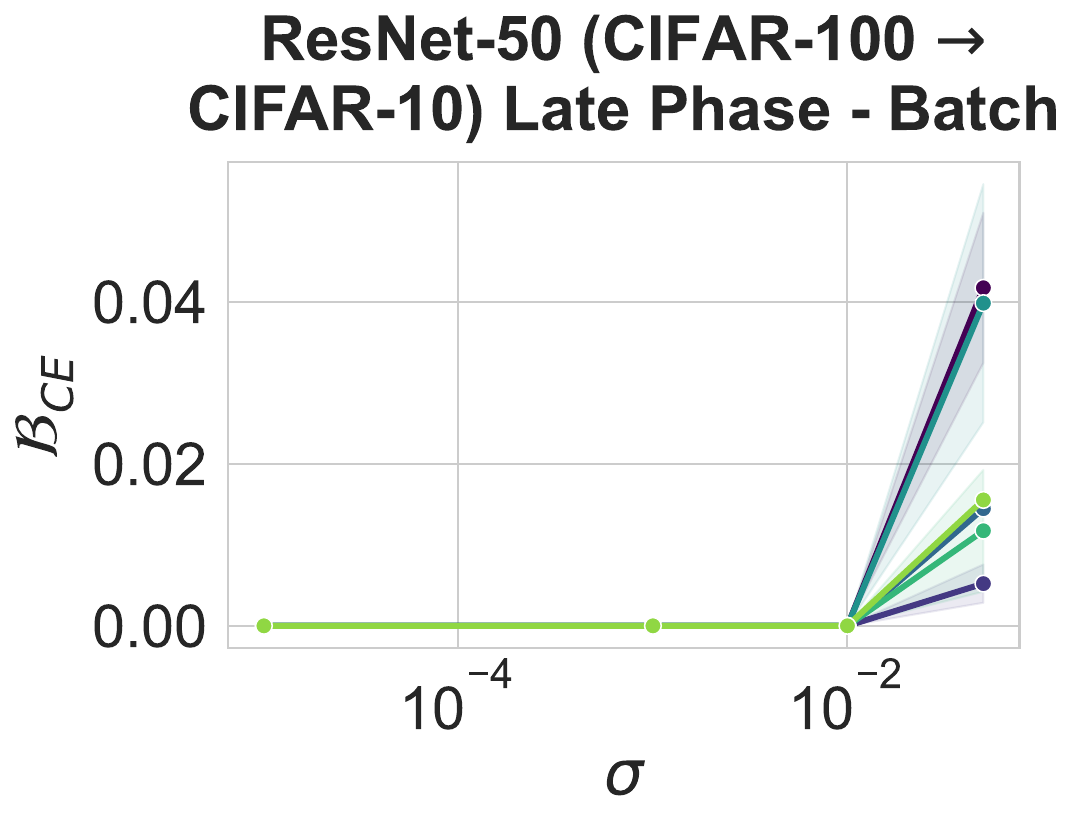}
    \includegraphics[width=0.3\linewidth]{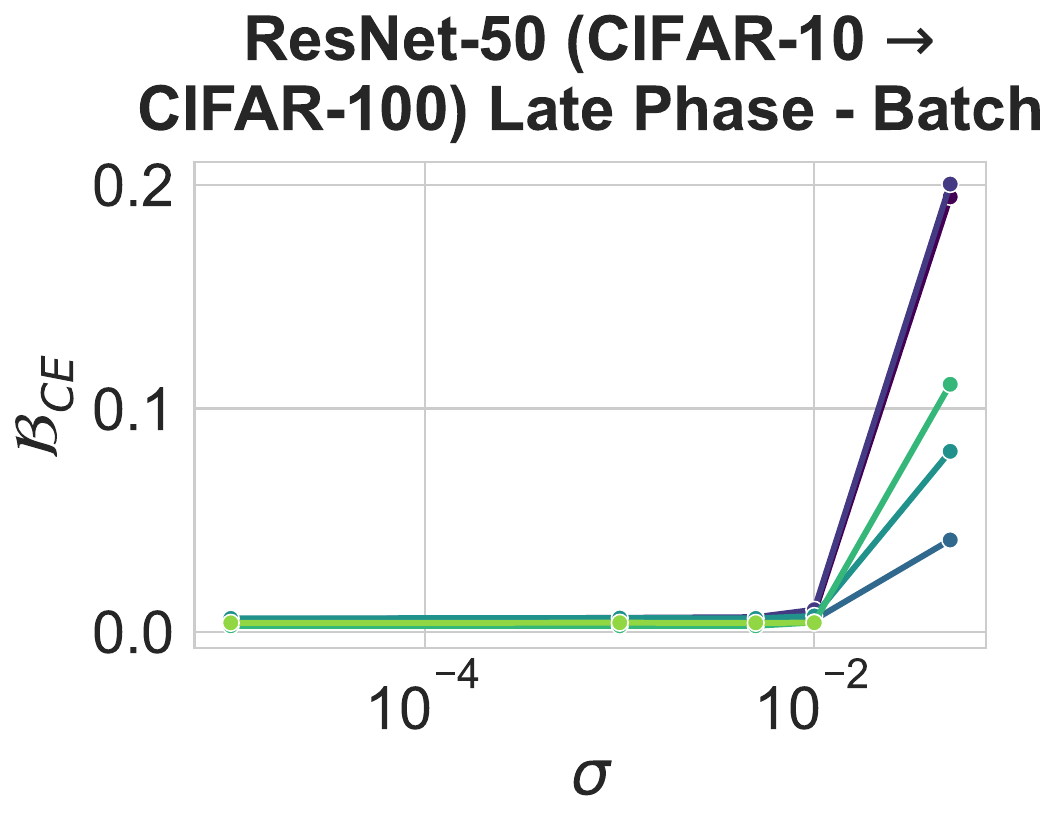}
}
\centerline{
    \includegraphics[width=0.5\linewidth]{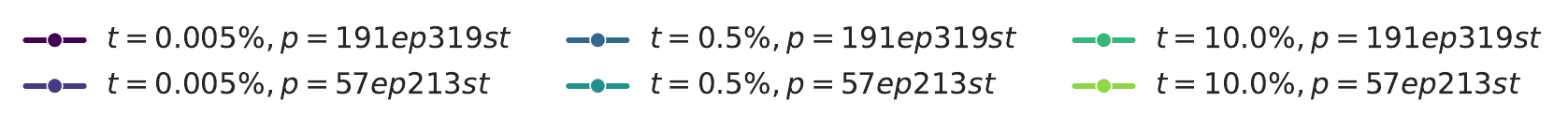}
}
\vskip -0.1in
\caption{
Same as \cref{fig:fine-tuning} (left) and \cref{ap:fig:fine-tuning:across-task-early}, but with additional pre-training times and perturbation times from late in training.
Barriers at scales $<10^{-2}$ are near-zero (see \cref{tab:fine-tuning:across-task-late-cifar100-to-10,tab:fine-tuning:across-task-late-cifar10-to-100} for exact values).
}
\label{ap:fig:fine-tuning:across-task-late}
\end{center}
\vskip -0.2in
\end{figure}

\emph{Extended pre-training yields near-zero fine-tuning barriers in ResNet-50/CIFAR settings.} \cref{fig:fine-tuning} shows that fine-tuning from later checkpoints greatly improves stability in ResNet-50 experiments. For improved readability, we zoom into the perturbations with $\sigma \leq 10^{-2}$ in \cref{ap:fig:fine-tuning:across-task-late,tab:fine-tuning:across-task-late-cifar10-to-100,tab:fine-tuning:across-task-late-cifar100-to-10}, revealing that transferring from CIFAR-100 to CIFAR-10 and vice-versa results in near-zero barriers. The CIFAR-100 to CIFAR-10 direction exhibits greater stability, which could indicate that CIFAR-100 pre-training is better suited for optimization on CIFAR-10 than the other way around.

\begin{figure}[ht]
\vskip 0.2in
\begin{center}
\centerline{
    \includegraphics[width=0.28\columnwidth]{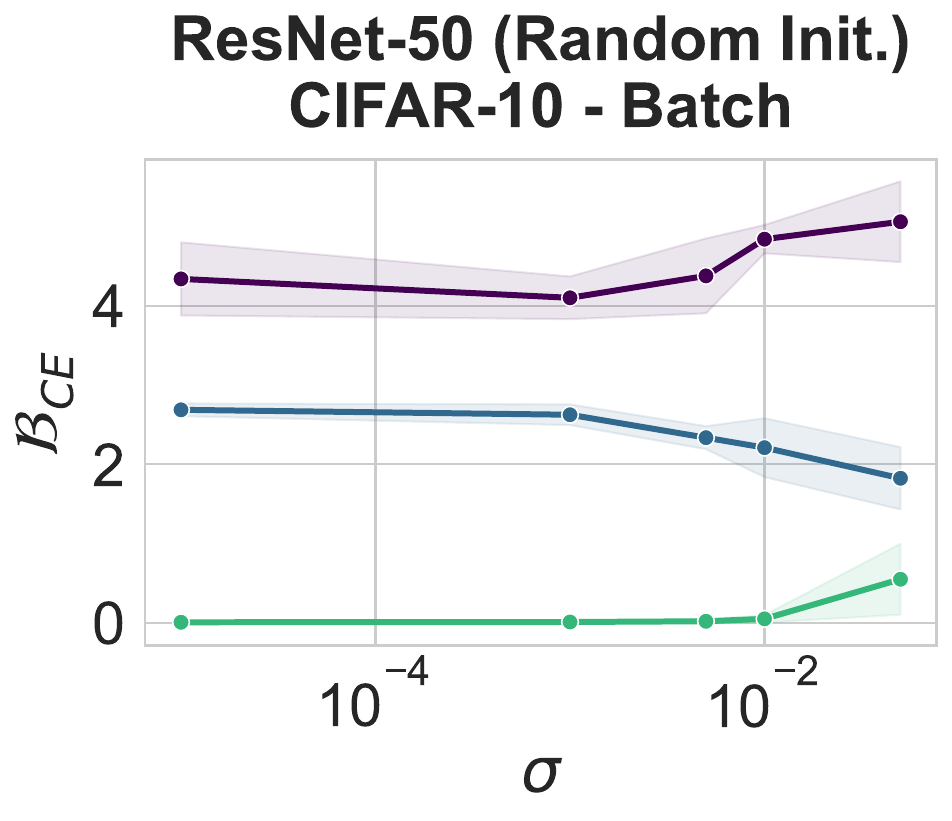}
}
\centerline{
    \includegraphics[width=0.33\columnwidth]{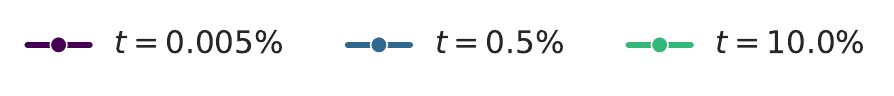}
}
\vskip -0.2in
\caption
{ 
Same as \cref{fig:fine-tuning,ap:fig:fine-tuning:across-task-early,ap:fig:fine-tuning:across-task-late} (left panel of each), but with randomly initialized ResNet-50 instead of networks pre-trained on CIFAR-100.
Train loss barriers are reported for batch perturbations during fine-tuning on CIFAR-10 using the recipe from \cref{ap:sec:finetuning-details}.
}
\label{ap:fig:resnet50-cifar10-random-init}
\end{center}
\vskip -2in
\end{figure}

\emph{Random initializations are less stable than pre-trained initializations.} In \cref{ap:fig:fine-tuning:across-task-late}, we demonstrated that pre-training improves fine-tuning stability, while \cref{ap:fig:fine-tuning:across-task-early} suggested that earlier checkpoints are more brittle to perturbations. As a baseline, we train a ResNet-50 from random initialization on CIFAR-10 and find that it exhibits even larger barriers (\cref{ap:fig:resnet50-cifar10-random-init}), with a similar magnitude to training from scratch.

\clearpage

\begin{table}[ht]
    \centering
        \caption{Train loss barriers and $L^2$ distance in Cifar-100 $\to$ Cifar-10 setting. Same as \cref{ap:fig:fine-tuning:across-task-late} (left).}
    \label{tab:fine-tuning:across-task-late-cifar100-to-10}
    \begin{tabular}{lrlrl}
\toprule
Starting Checkpoint & Relative Perturb Step (\%) & $\sigma$ & $L^2$ Distance & Train $\mathcal{B}_\mathrm{CE}$\\
\midrule
191ep319st & 0.005 & 0.00001 & 39.41 ± 0.23 & 0.00 ± 0.00 \\
191ep319st & 0.005 & 0.00100 & 39.46 ± 0.18 & 0.00 ± 0.00 \\
191ep319st & 0.005 & 0.01000 & 40.41 ± 0.82 & 0.00 ± 0.00 \\
191ep319st & 0.5 & 0.00001 & 39.49 ± 0.23 & 0.00 ± 0.00 \\
191ep319st & 0.5 & 0.00100 & 39.42 ± 0.19 & 0.00 ± 0.00 \\
191ep319st & 0.5 & 0.01000 & 40.03 ± 0.62 & 0.00 ± 0.00 \\
191ep319st & 10 & 0.00001 & 38.02 ± 0.18 & 0.00 ± 0.00 \\
191ep319st & 10 & 0.00100 & 38.04 ± 0.21 & 0.00 ± 0.00 \\
191ep319st & 10 & 0.01000 & 38.19 ± 0.13 & 0.00 ± 0.00 \\
57ep213st & 0.005 & 0.00001 & 40.81 ± 0.14 & 0.00 ± 0.00 \\
57ep213st & 0.005 & 0.00100 & 40.93 ± 0.15 & 0.00 ± 0.00 \\
57ep213st & 0.005 & 0.01000 & 41.49 ± 0.44 & 0.00 ± 0.00 \\
57ep213st & 0.5 & 0.00001 & 40.90 ± 0.06 & 0.00 ± 0.00 \\
57ep213st & 0.5 & 0.00100 & 40.82 ± 0.17 & 0.00 ± 0.00 \\
57ep213st & 0.5 & 0.01000 & 41.26 ± 0.29 & 0.00 ± 0.00 \\
57ep213st & 10 & 0.00001 & 39.50 ± 0.14 & 0.00 ± 0.00 \\
57ep213st & 10 & 0.00100 & 39.50 ± 0.13 & 0.00 ± 0.00 \\
57ep213st & 10 & 0.01000 & 39.63 ± 0.10 & 0.00 ± 0.00 \\
\bottomrule
\end{tabular}
\end{table}

\begin{table}[ht]
    \centering
        \caption{Train loss barriers and $L^2$ distance in Cifar-10 $\to$ Cifar-100 setting. Same as \cref{ap:fig:fine-tuning:across-task-late} (right).}
    \label{tab:fine-tuning:across-task-late-cifar10-to-100}
    \begin{tabular}{lrlrl}
\toprule
Starting Checkpoint & Relative Perturb Step (\%) & $\sigma$ & $L^2$ Distance & Train $\mathcal{B}_\mathrm{CE}$\\
\midrule
191ep319st & 0.005 & 0.00001 & 67.98 ± 0.00 & 0.00 ± 0.00 \\
191ep319st & 0.005 & 0.00100 & 68.05 ± 0.00 & 0.00 ± 0.00 \\
191ep319st & 0.005 & 0.00500 & 68.92 ± 0.00 & 0.01 ± 0.00 \\
191ep319st & 0.005 & 0.01000 & 70.73 ± 2.25 & 0.01 ± 0.00 \\
191ep319st & 0.5 & 0.00001 & 67.89 ± 0.00 & 0.00 ± 0.00 \\
191ep319st & 0.5 & 0.00100 & 67.96 ± 0.00 & 0.01 ± 0.00 \\
191ep319st & 0.5 & 0.00500 & 68.18 ± 0.00 & 0.01 ± 0.00 \\
191ep319st & 0.5 & 0.01000 & 68.34 ± 0.25 & 0.01 ± 0.00 \\
191ep319st & 10 & 0.00001 & 64.05 ± 0.00 & 0.00 ± 0.00 \\
191ep319st & 10 & 0.00100 & 64.11 ± 0.00 & 0.00 ± 0.00 \\
191ep319st & 10 & 0.00500 & 64.47 ± 0.00 & 0.00 ± 0.00 \\
191ep319st & 10 & 0.01000 & 65.54 ± 0.94 & 0.00 ± 0.00 \\
57ep213st & 0.005 & 0.00001 & 62.34 ± 0.00 & 0.01 ± 0.00 \\
57ep213st & 0.005 & 0.00100 & 62.68 ± 0.00 & 0.01 ± 0.00 \\
57ep213st & 0.005 & 0.00500 & 63.31 ± 0.00 & 0.01 ± 0.00 \\
57ep213st & 0.005 & 0.01000 & 65.23 ± 2.39 & 0.01 ± 0.00 \\
57ep213st & 0.5 & 0.00001 & 62.32 ± 0.00 & 0.01 ± 0.00 \\
57ep213st & 0.5 & 0.00100 & 62.34 ± 0.00 & 0.01 ± 0.00 \\
57ep213st & 0.5 & 0.00500 & 62.70 ± 0.00 & 0.01 ± 0.00 \\
57ep213st & 0.5 & 0.01000 & 63.21 ± 0.72 & 0.01 ± 0.00 \\
57ep213st & 10 & 0.00001 & 59.47 ± 0.00 & 0.00 ± 0.00 \\
57ep213st & 10 & 0.00100 & 59.43 ± 0.00 & 0.00 ± 0.00 \\
57ep213st & 10 & 0.00500 & 59.50 ± 0.00 & 0.00 ± 0.00 \\
57ep213st & 10 & 0.01000 & 59.70 ± 0.29 & 0.00 ± 0.00 \\
\bottomrule
\end{tabular}
\vskip -2in
\end{table}

\clearpage

\subsection{ViT Fine-Tuning \label{sec:ap:vit-stability}}
Initial fine-tuning stability experiments on ResNets and MultiBERTs suggested opposed trends between vision tasks with convolutional architectures, and language tasks with tarnsformer architectures.
To disentangle whether this difference is due to the nature of the task or architecture, we applied our experimental procedure to Vision Transformers (ViTs), which are representative of the transformer architecture and larger-scale models than ResNet-50.
The sources for the pre-trained ViTs and the fine-tuning procedure we use are specified in \cref{par:vit-finetune}.

While we did not have access to intermediate training checkpoints for ViT models, we instead compare four ViTs of different sizes which were pre-trained on ImageNet variants: google/vit-base-patch16-224 (86M parameters), google/vit-base-patch16-224-in21k (86M parameters), google/vit-large-patch16-224-in21k (304M parameters), and google/vit-huge-patch14-224-in21k (632M parameters).
The size of each pre-training dataset serves as a proxy for pre-training duration.

\begin{figure*}[ht]
\vskip 0.1in
\begin{center}
\centerline{
    \includegraphics[height=0.22\linewidth]{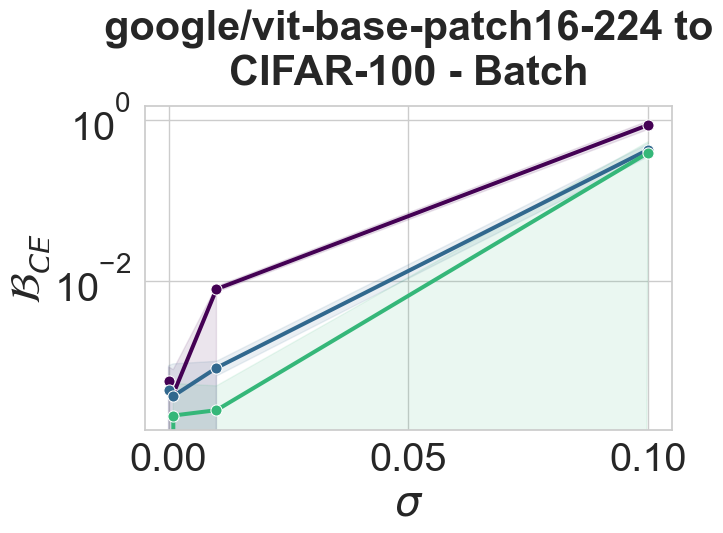}
    \includegraphics[height=0.22\linewidth]{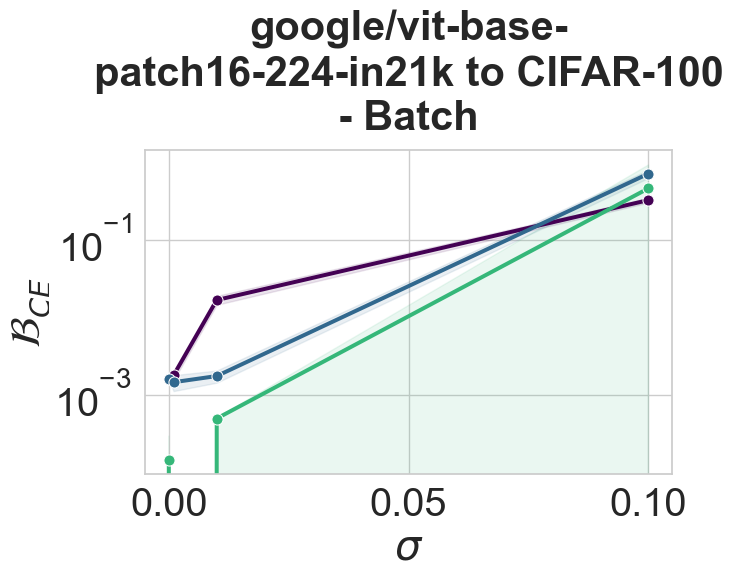}
}
\centerline{
    \includegraphics[height=0.22\linewidth]{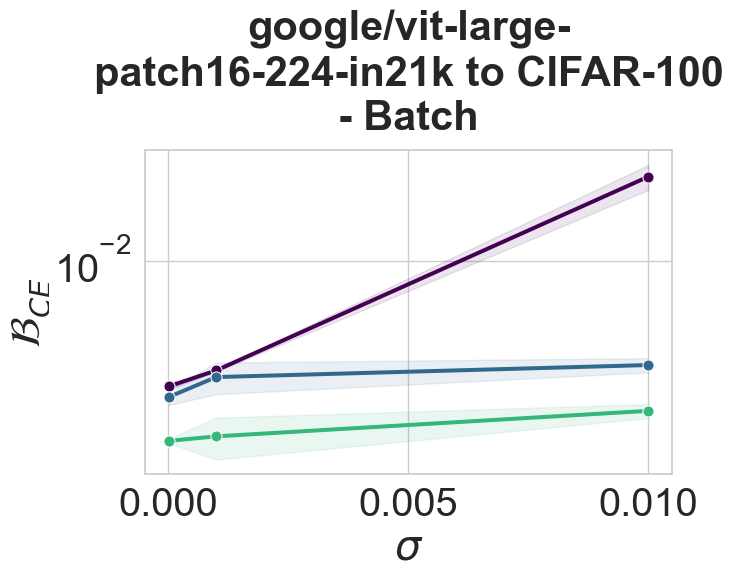}
    \includegraphics[height=0.22\linewidth]{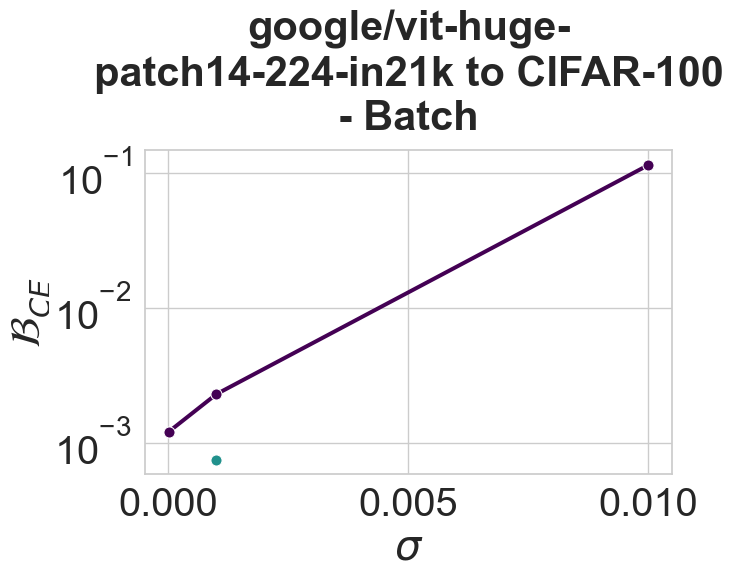}
}
\centerline{
    \includegraphics[width=0.5\linewidth]{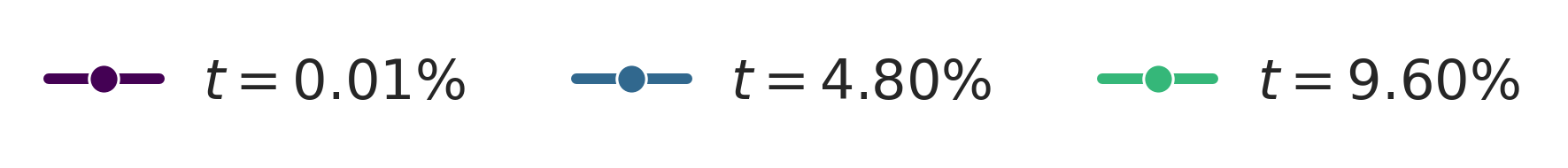}
}
\caption{Stability of various ViT architectures on CIFAR-100. Training loss barriers after training (y-axis) are plotted against perturbation magnitude (x-axis) and perturbation step (color). We fine-tune four pre-trained ViTs of different sizes: ViT-Base/16-224 (86M parameters, pre-trained on ImageNet-21K and then fine-tuned on ImageNet), ViT-Base/16-224-in21k (86M parameters pre-trained on ImageNet-21k), ViT-Large/16-224-in21k (304M parameters), and ViT-Huge/14-224-in21k (632M parameters).}
\label{ap:fig:vit-fine-tuning:time}
\end{center}
\vskip -0.2in
\end{figure*}

\emph{Extending our findings to ViTs fine-tuned on CIFAR-100.}
Consistent with our findings on smaller convolutional networks, in \cref{ap:fig:vit-fine-tuning:time} we observe that earlier and larger perturbations cause more pronounced barriers. 
Interestingly, the google/vit-base-patch16-224 variant, which underwent additional fine-tuning on ImageNet-1K after its initial ImageNet-21K pre-training, effectively represents a longer training process.
Although the exact learning rate schedule is unknown, this setting resembles the extended pre-training observed in later BERT checkpoints.
Models fine-tuned from this variant (\cref{ap:fig:vit-fine-tuning:time}, top left) exhibit the largest barriers among all settings, with barriers one order of magnitude larger than those from google/vit-base-patch16-224-in21k (\cref{ap:fig:vit-fine-tuning:time}, top right).
This provides additional evidence that extended pre-training reduces fine-tuning stability.
Future work could investigate the stability of intermediate ViT checkpoints.

\begin{figure*}[ht]
\vskip 0.1in
\begin{center}
\centerline{
    \includegraphics[height=0.22\linewidth]{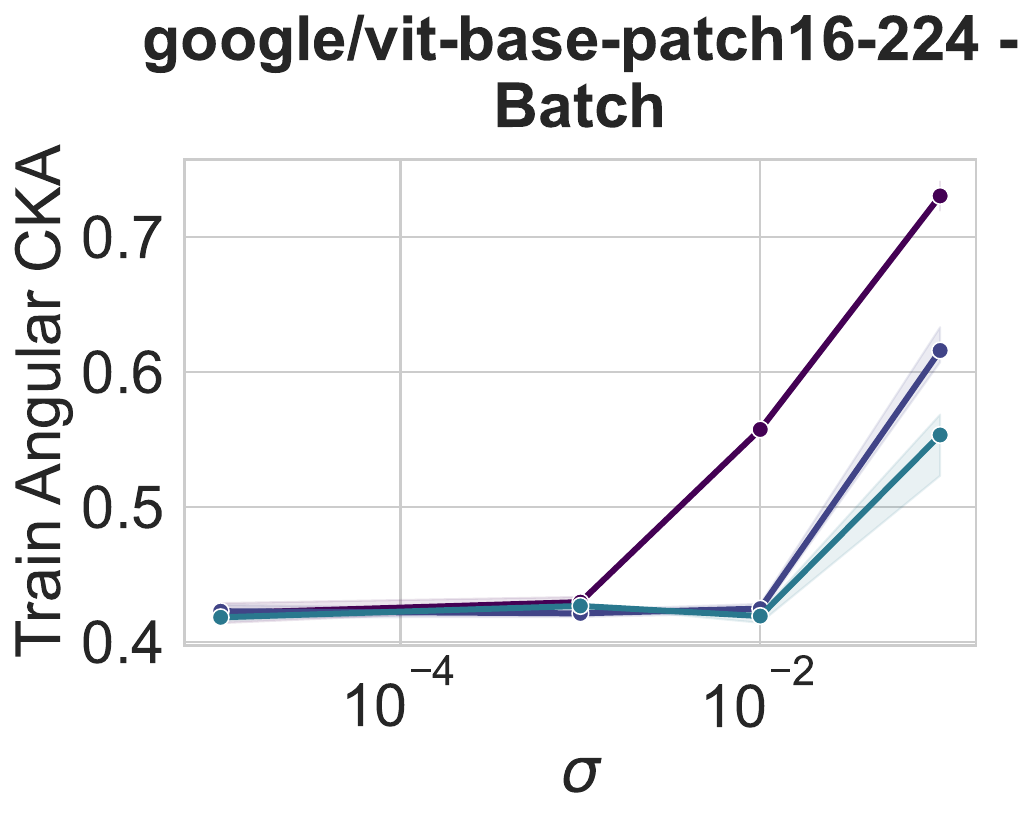}
    \includegraphics[height=0.22\linewidth]{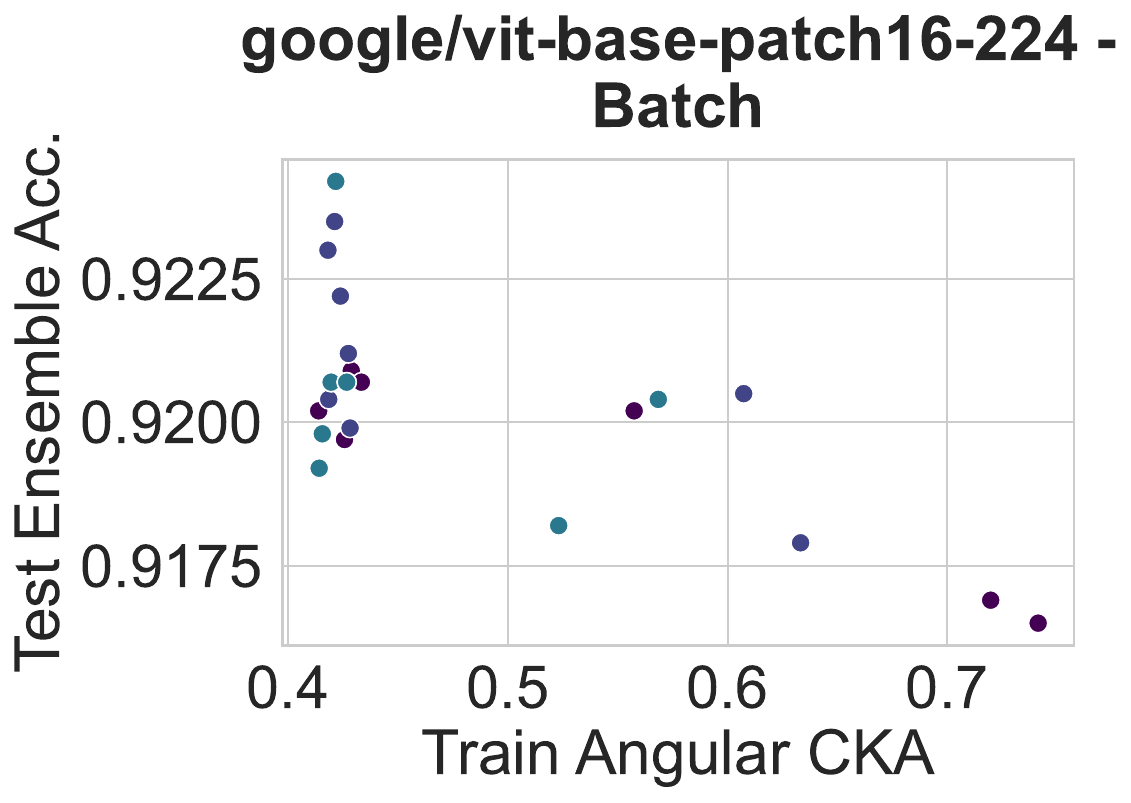}
    \includegraphics[height=0.22\linewidth]{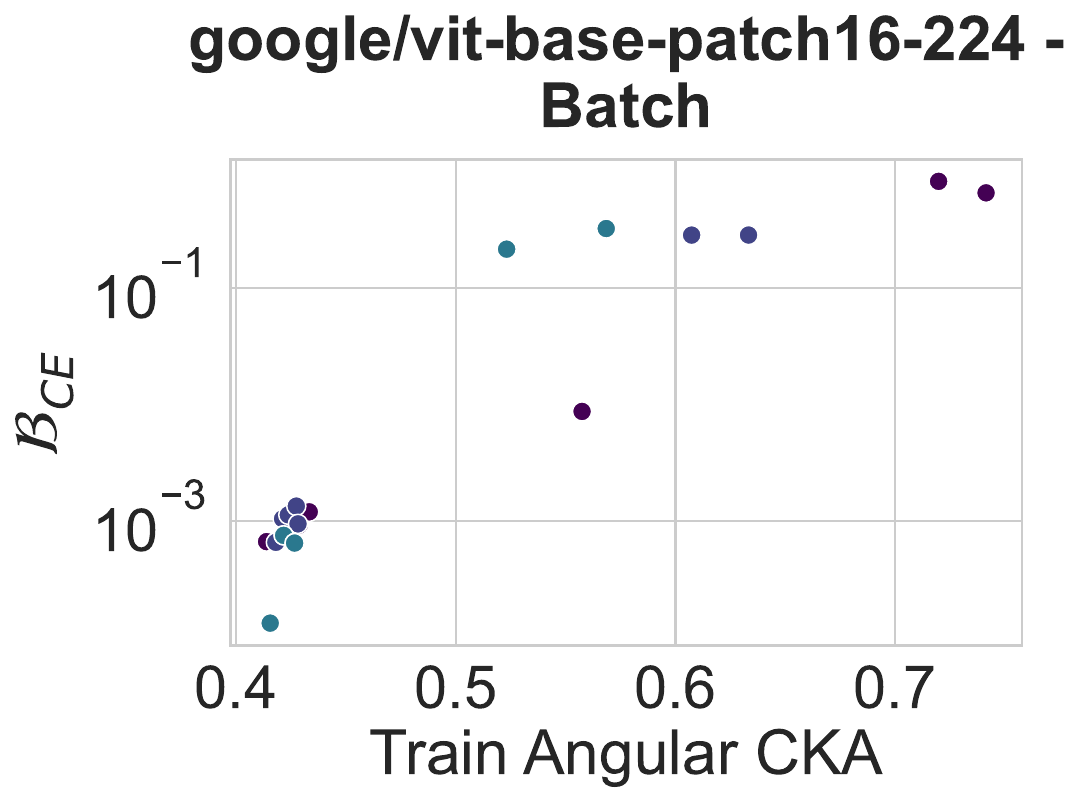}
}
\centerline{
    \includegraphics[width=0.4\linewidth]{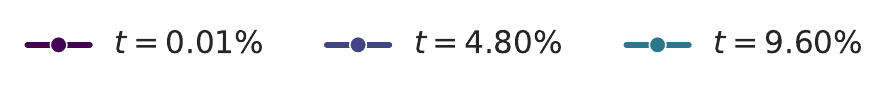}
}
\caption{\textbf{Left:} representational similarity distance measured via Angular CKA for ViT-base. \textbf{Middle:} test ensembling accuracy against Angular CKA. 
\textbf{Right:} training loss barrier against Angular CKA.
}
\label{ap:fig:vit-cka-l2}
\end{center}
\vskip -0.2in
\end{figure*}

\subsection{BERT Fine-Tuning}

\begin{figure*}[ht]
\vskip 0.1in
\begin{center}
\centerline{
    \includegraphics[height=0.22\linewidth]{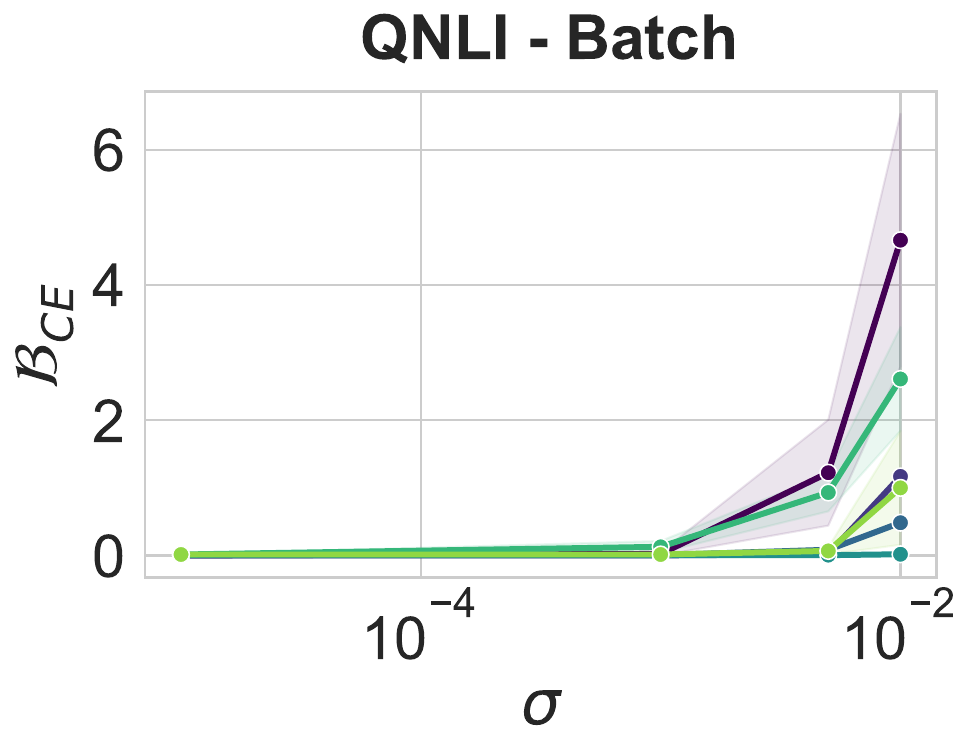}
    \includegraphics[height=0.22\linewidth]{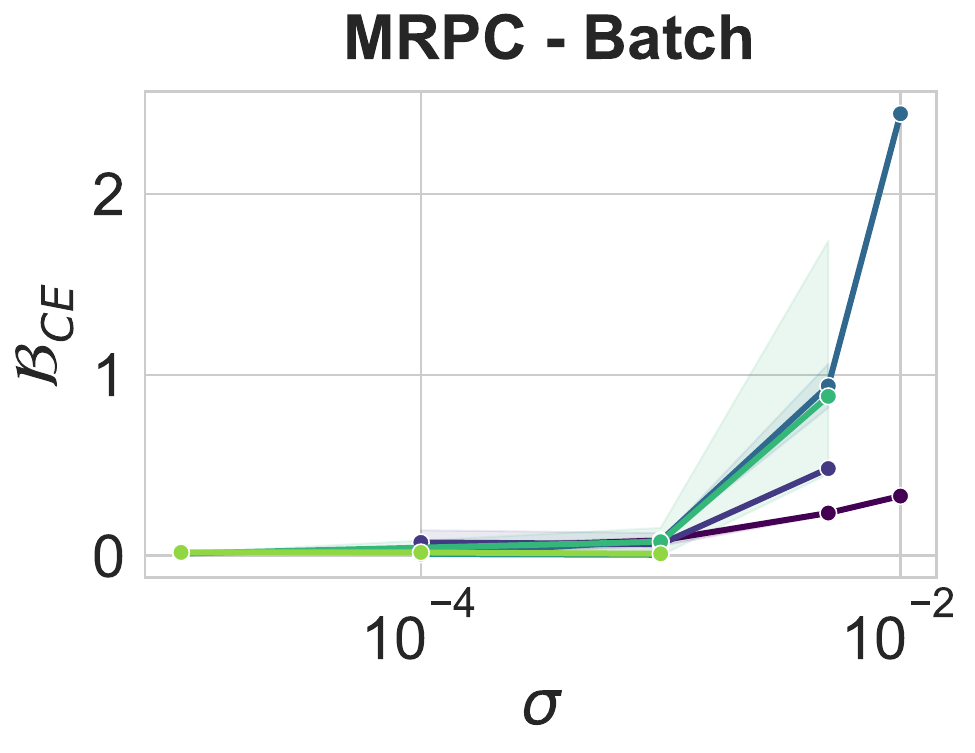}
}
\centerline{
    \includegraphics[height=0.22\linewidth]{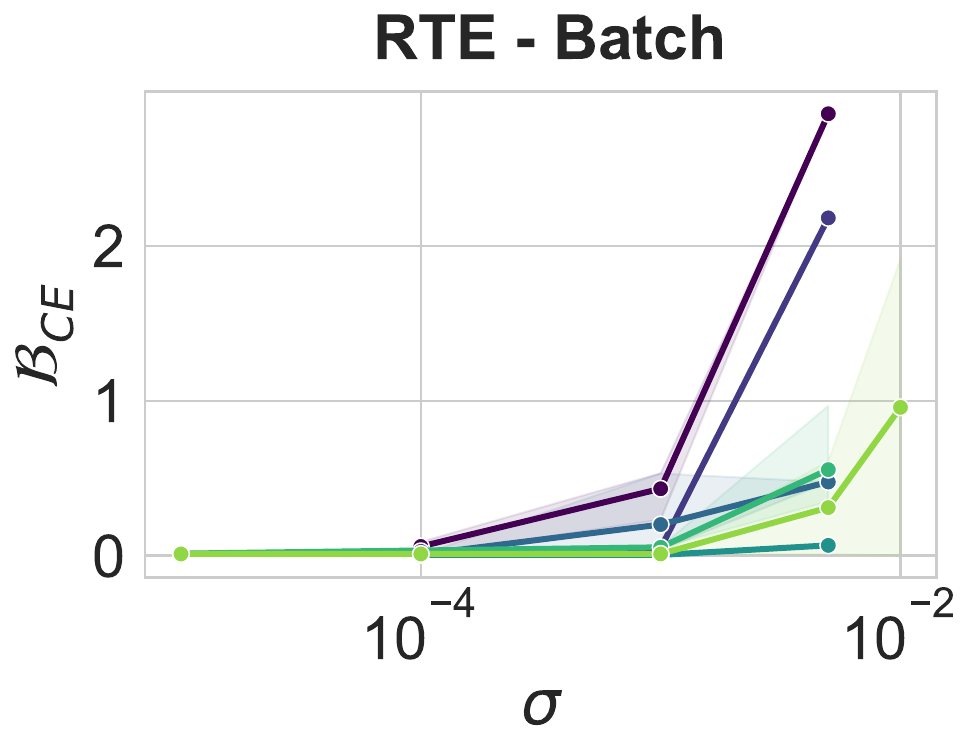}
    \includegraphics[height=0.22\linewidth]{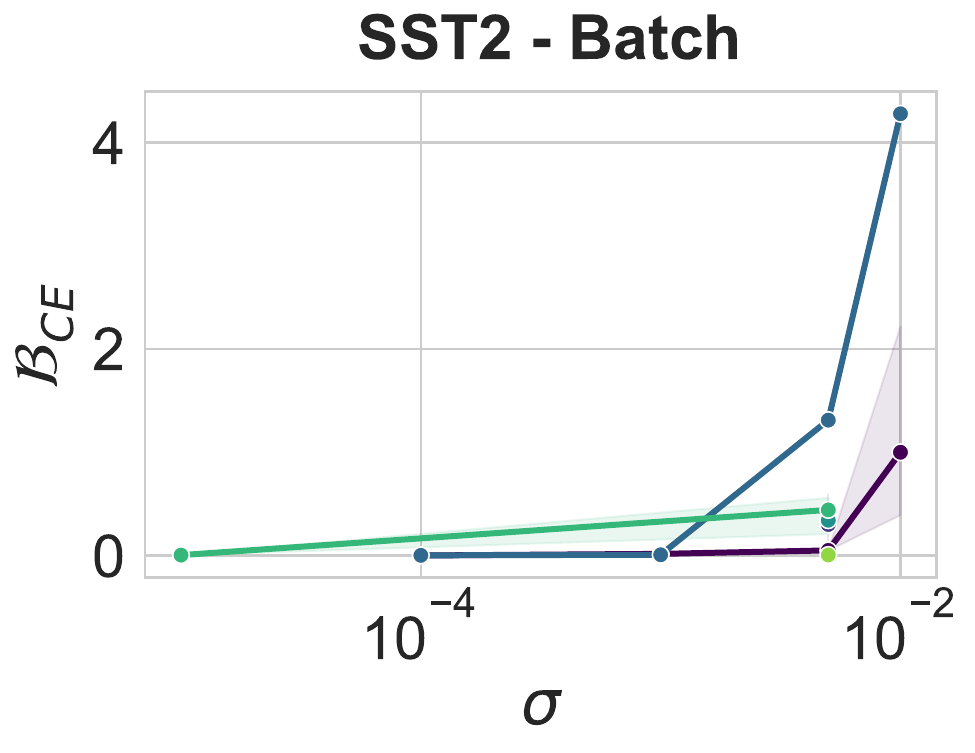}
    \includegraphics[height=0.22\linewidth]{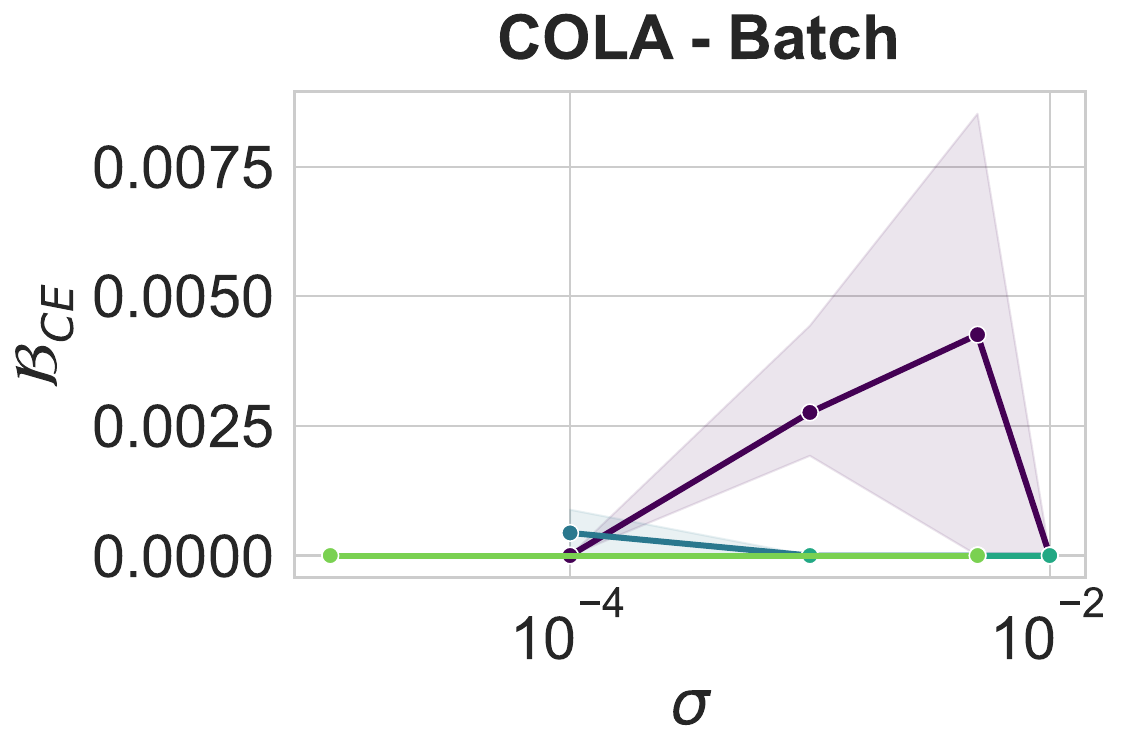}
}
\centerline{
    \includegraphics[width=0.6\linewidth]{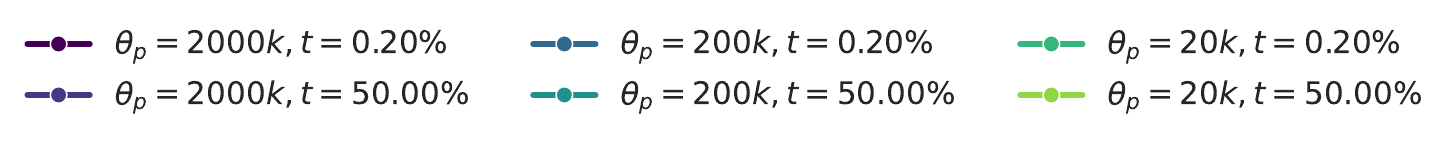}
    }
\caption{Same as \cref{fig:fine-tuning} (right), but with additional pre-training times, perturbation times, and tasks. Fine-tuning stability of Multi-BERT on QNLI and MRPC, starting from 20K, 200K, and 2000K checkpoints with different perturbation times. Tasks are QNLI (\textbf{top left}), MRPC (\textbf{top right}), RTE (\textbf{bottom left}), SST-2 (\textbf{bottom middle}), and COLA (\textbf{bottom right}).
}
\label{ap:fig:bert-fine-tuning:time}
\end{center}
\vskip -0.2in
\end{figure*}

\cref{ap:fig:bert-fine-tuning:time} plots additional fine-tuning tasks as described in \cref{ap:sec:finetuning-details}. 

\begin{figure*}[ht]
\vskip 0.1in
\begin{center}
\centerline{
    \includegraphics[height=0.22\linewidth]{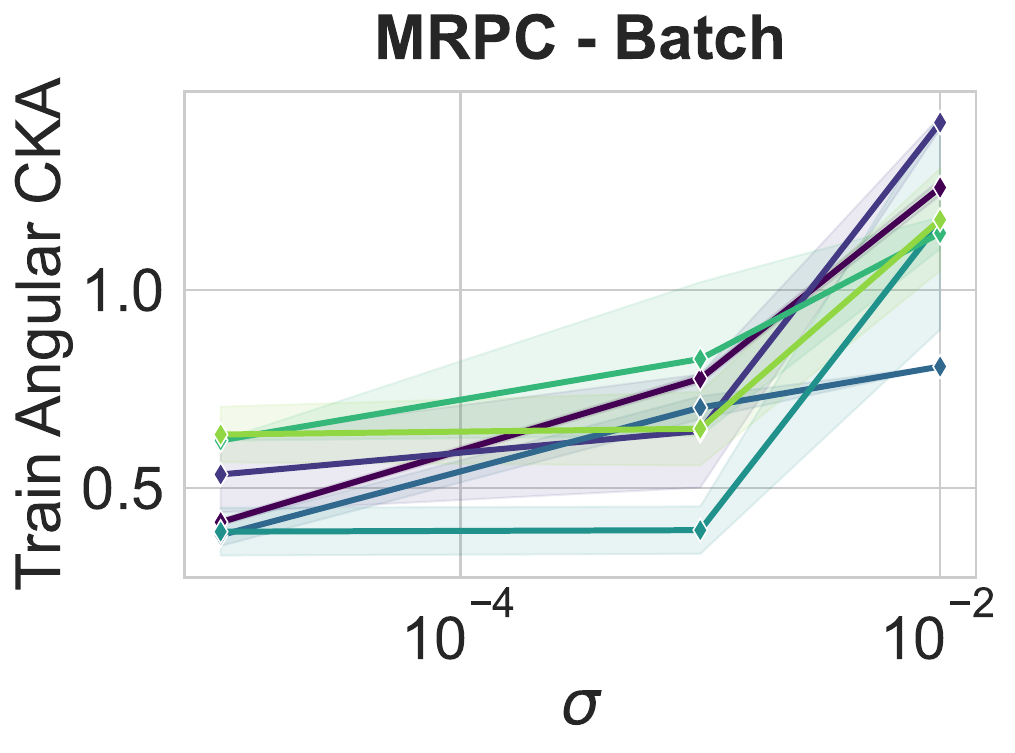}
    \includegraphics[height=0.22\linewidth]{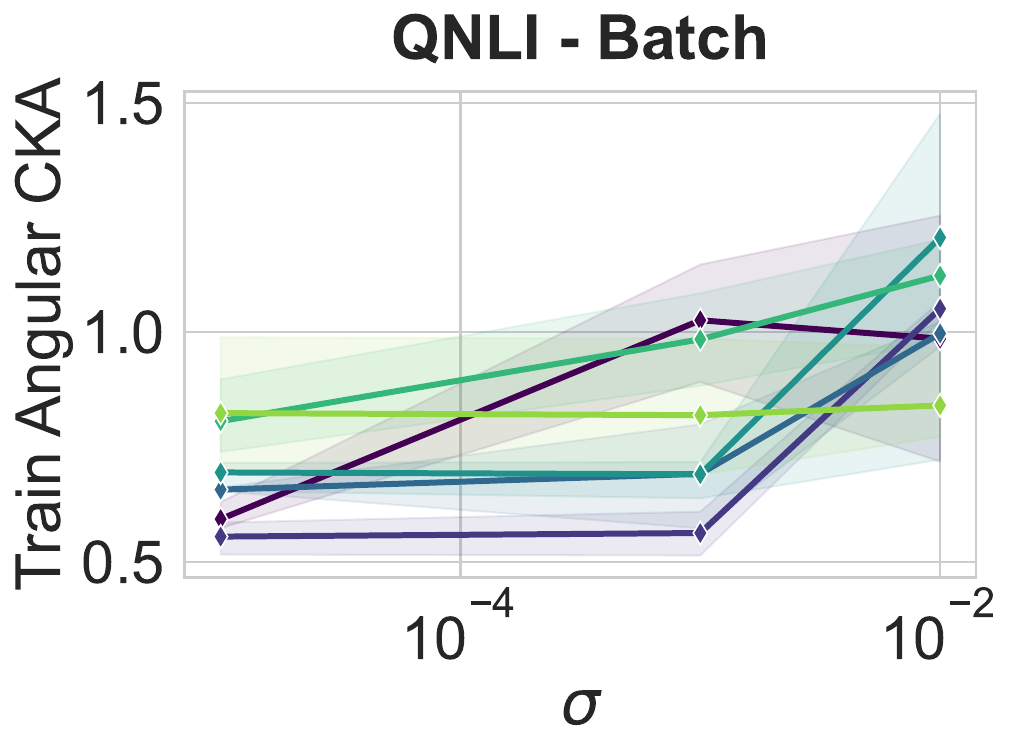}
}
\centerline{
    \includegraphics[height=0.22\linewidth]{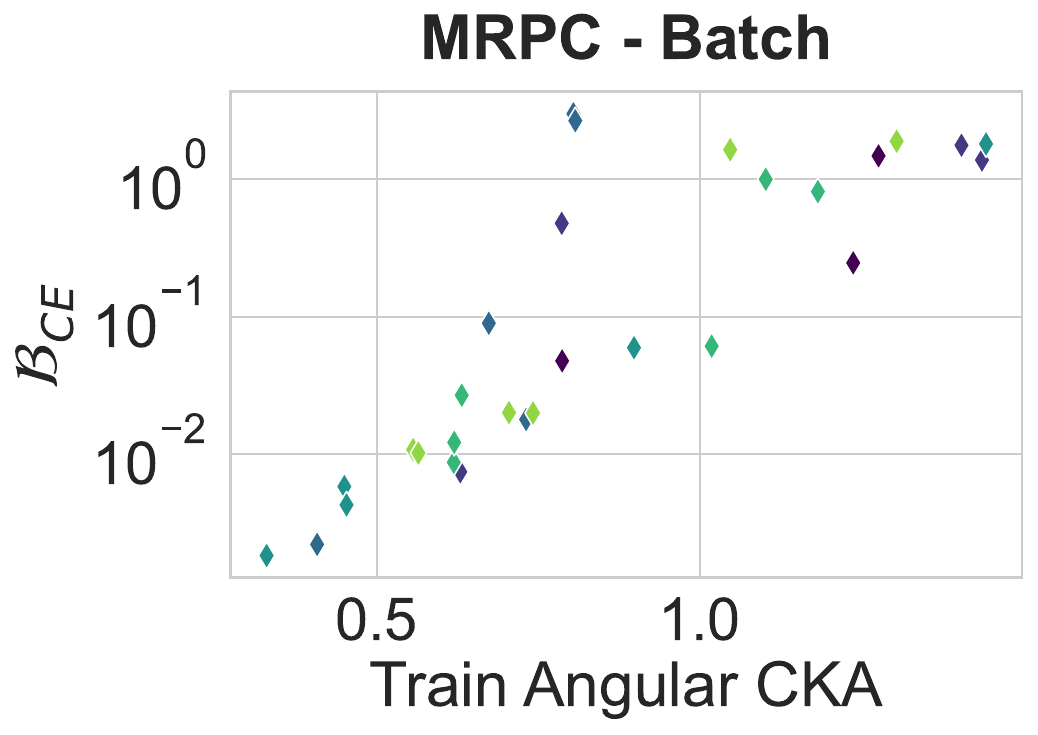}
    \includegraphics[height=0.22\linewidth]{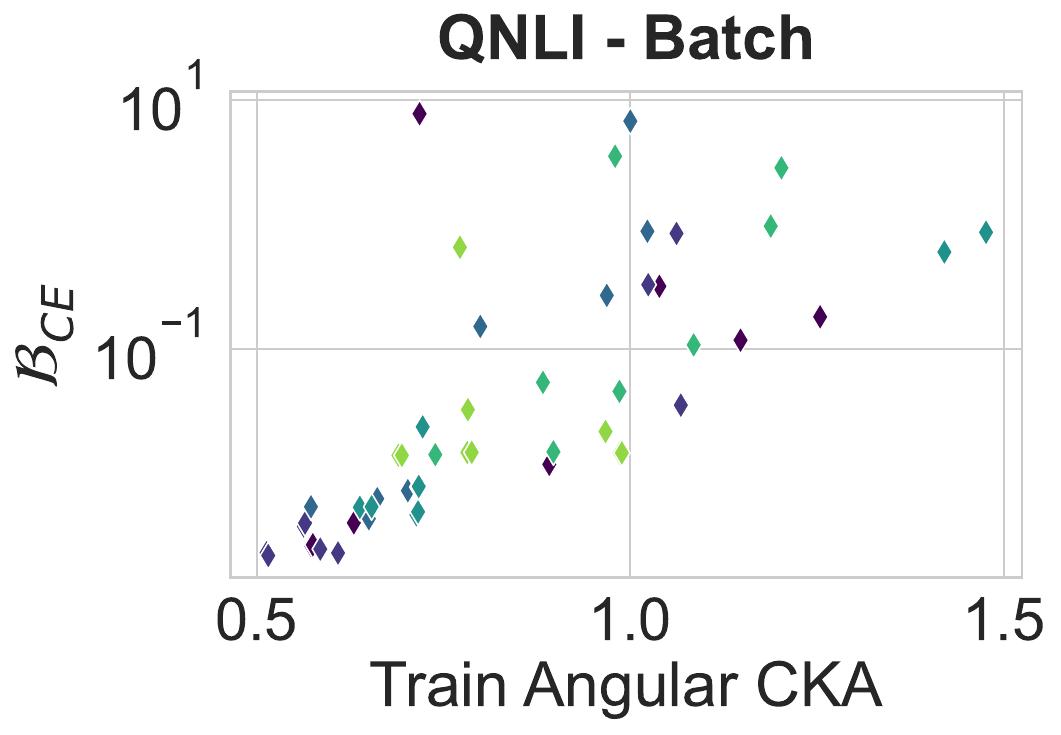}
}
\centerline{
    \includegraphics[height=0.06\linewidth]{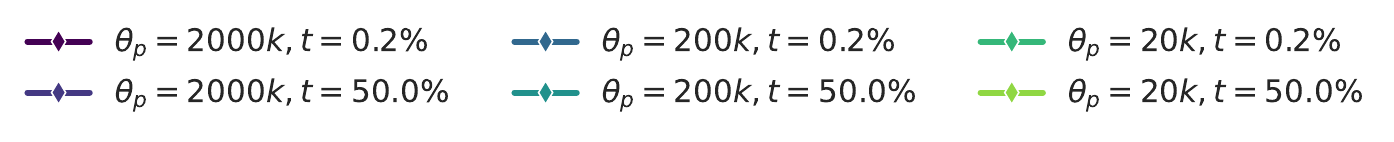}
}
\caption{\textbf{Top:} representational similarity distance measured via Angular CKA for MultiBERT on MRPC (\textbf{left}) and QNLI (\textbf{right}). \textbf{Bottom:} barriers vs. angular CKA on MRPC (\textbf{left}) and QNLI (\textbf{right}).}
\label{ap:fig:bert-cka}
\end{center}
\vskip -0.2in
\end{figure*}

\emph{Representational similarity for BERT models.}
We also provide Angular CKA plots for BERT on MRPC and QNLI datasets in \cref{ap:fig:bert-cka}.
\cref{ap:fig:bert-cka} shows that barriers are correlated with Angular CKA, indicating real functional differences between the networks.
This is consistent with our findings in vision models (\cref{fig:butterfly-cka,fig:butterfly-cka-additional-barriers,ap:fig:vit-cka-l2} right), unlike the correlation between barriers and $L^2$ divergence (\cref{ap:fig:bert-transfer-l2-barr}) which is not consistent between vision and language settings.

\subsection{OLMo Fine-Tuning}
\begin{figure}[ht]
\begin{center}
\vskip 0.1in
\centerline{
    \includegraphics[height=0.25\linewidth]{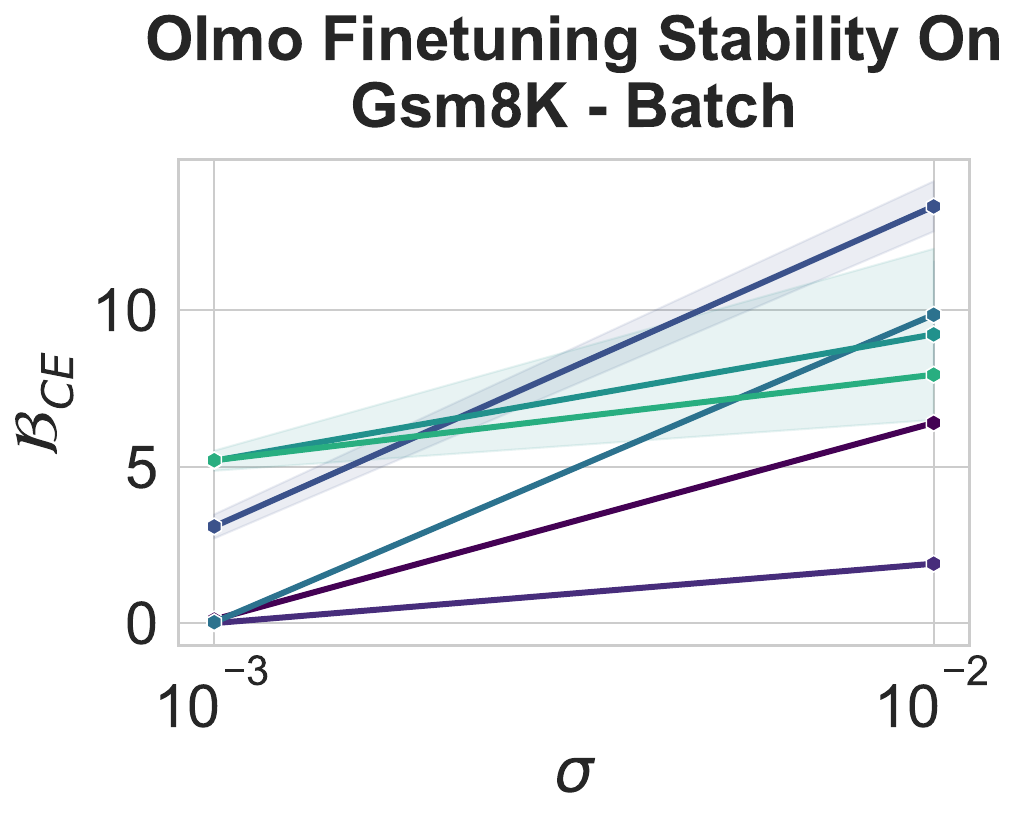}
    \includegraphics[height=0.25\linewidth]{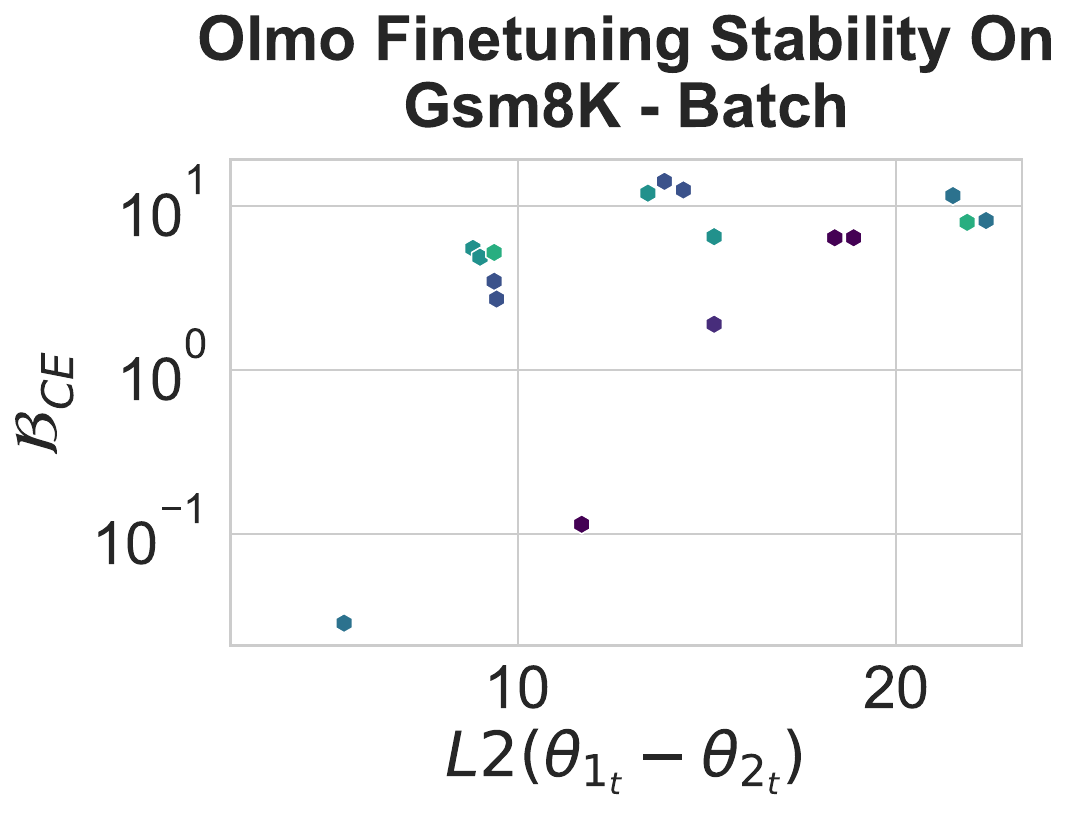}
}
\centerline{
    \includegraphics[width=0.8\linewidth]{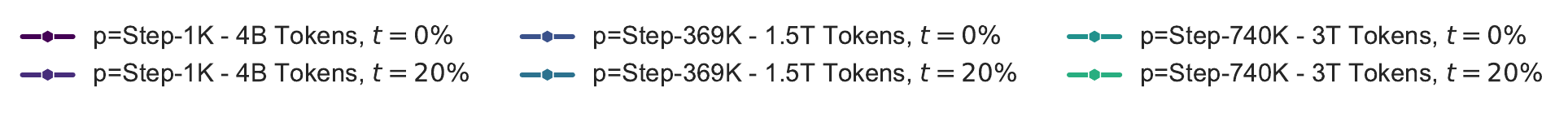}
}
\vskip -0.1in
\caption{
Stability of fine-tuning OLMo-1B on GSM8K mathematical reasoning tasks. We fine-tune OLMo-1B checkpoints from different pre-training stages (early, middle, and final checkpoints) on GSM8K with batch perturbations applied at various training steps. \textbf{Left:} Loss barriers (y-axis) plotted against perturbation magnitude (x-axis) for different checkpoint combinations and perturbation steps (colors). \textbf{Right:} Barriers vs. $L^2$ distance between the original and perturbed models. Consistent with our vision experiments, earlier perturbations and later pre-training checkpoints lead to higher loss barriers, demonstrating that fine-tuning stability patterns generalize from vision to language models.
}
\label{fig:olmo-fine-tuning:time}
\end{center}
\vskip -0.2in
\end{figure}

\cref{fig:olmo-fine-tuning:time} demonstrates that the trends we observe in terms of perturbation time and magnitude extend to decoder-only large language models. 
Consistent with our MultiBERT findings, we find that more pre-training does not necessarily lead to improved fine-tuning stability.

\subsection{$L^2$ Divergence and Barriers}
\begin{figure}[ht]
\vskip 0.1in
\begin{center}
\centerline{
    \includegraphics[height=0.25\linewidth]{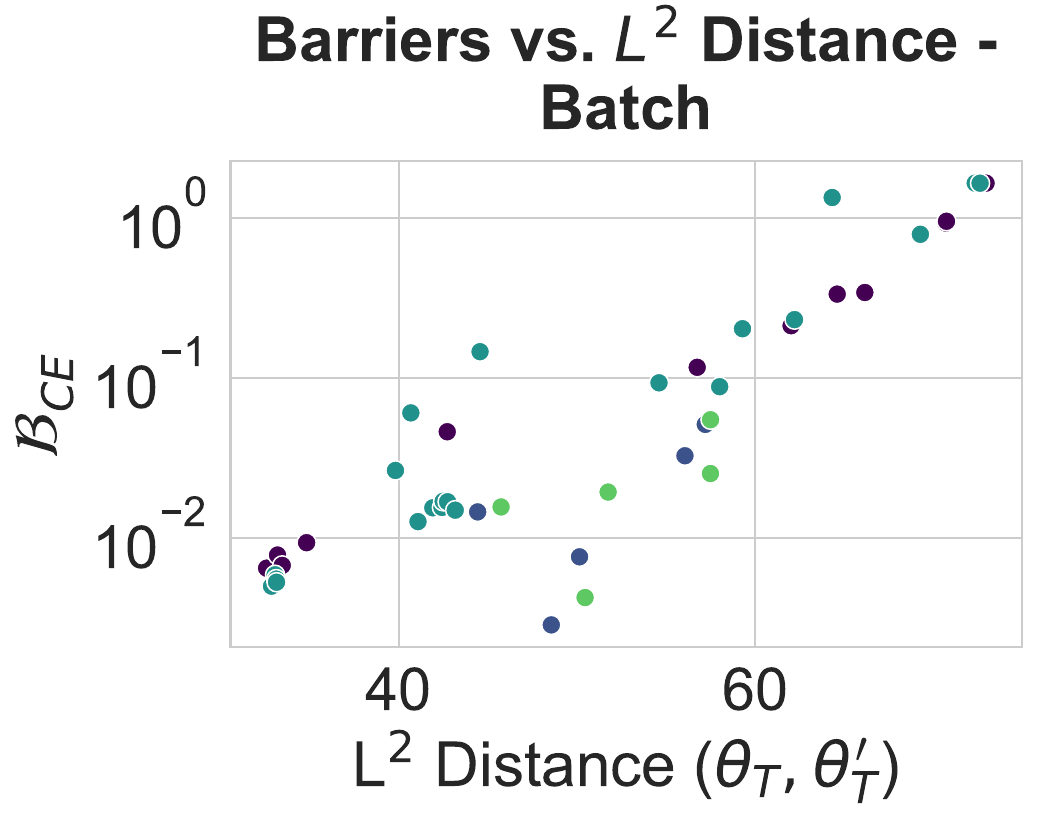}
    \includegraphics[height=0.25\linewidth]{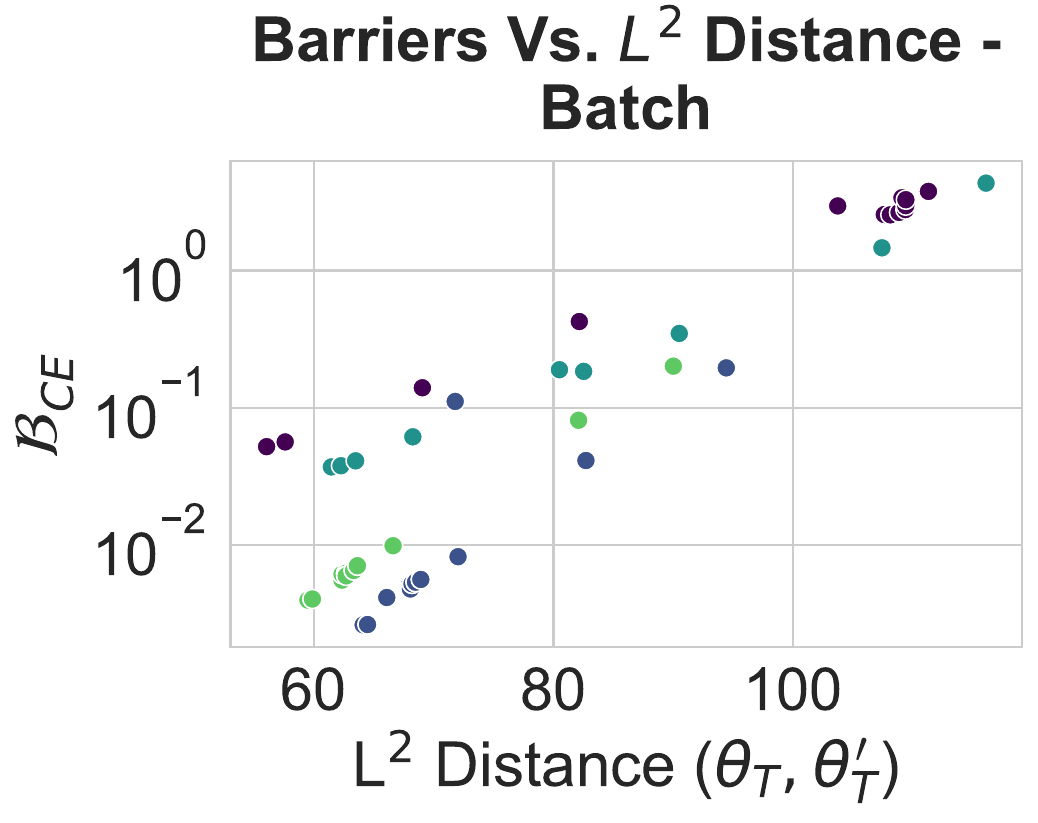}
    \includegraphics[height=0.25\linewidth]{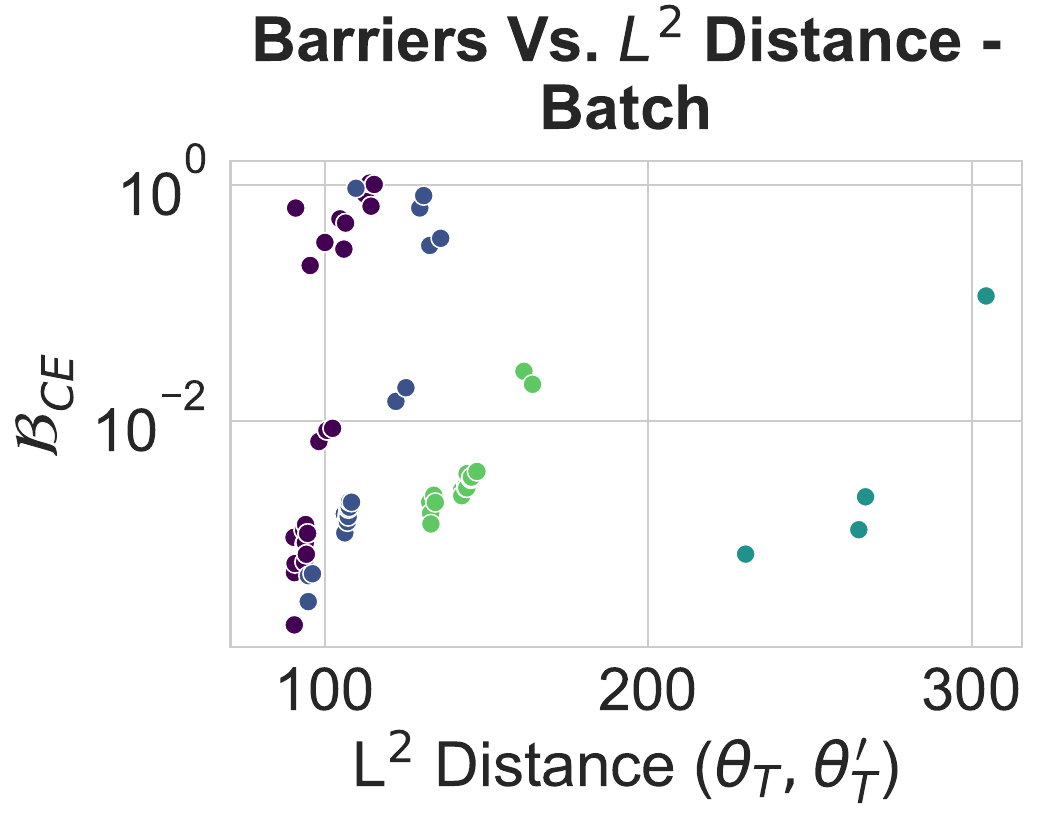}
}
\centerline{
    \includegraphics[width=0.5\columnwidth]{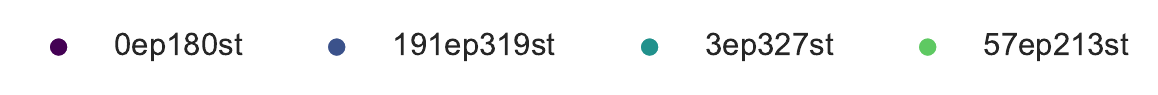}
    \includegraphics[width=0.5\linewidth]{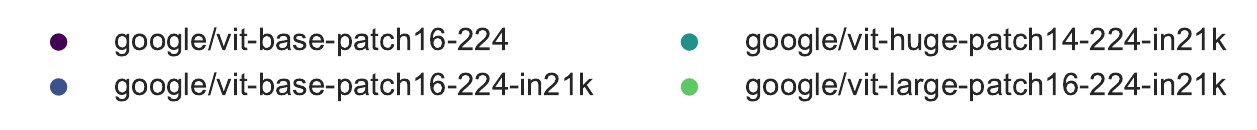}
}
\caption{
Train loss barriers vs. $L^2$ distance between the perturbed and original models at the end of training for fine-tuning vision models: ResNet-50 transferring from CIFAR-100 to CIFAR-10 (\textbf{left}), ResNet-50 transferring from CIFAR-10 to CIFAR-100 (\textbf{middle}), and ViT-base fine-tuned on CIFAR-100 (\textbf{right}). Note the legend for the colors is different in the rightmost plot.
}
\label{ap:fig:finetune-cifar-l2-barr}
\end{center}
\vskip -0.2in
\end{figure}

\begin{figure}[ht]
\vskip 0.1in
\begin{center}
\centerline{
    \includegraphics[width=0.25\linewidth]{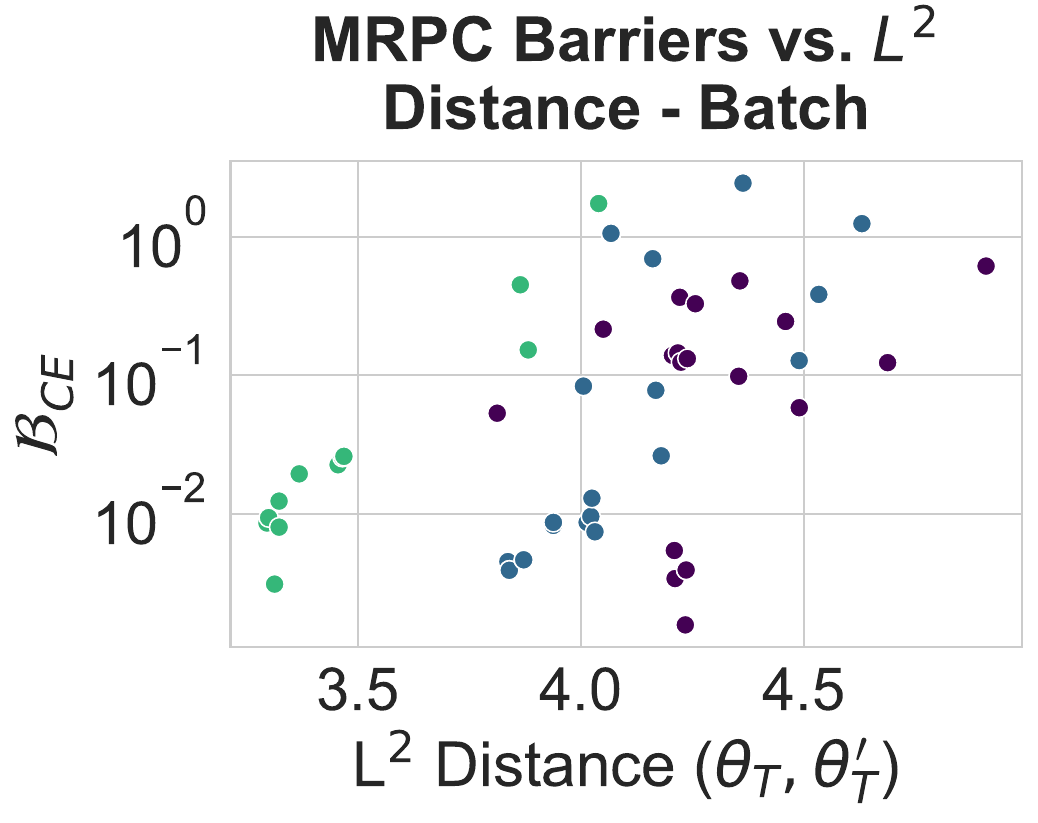}
    \includegraphics[width=0.25\linewidth]{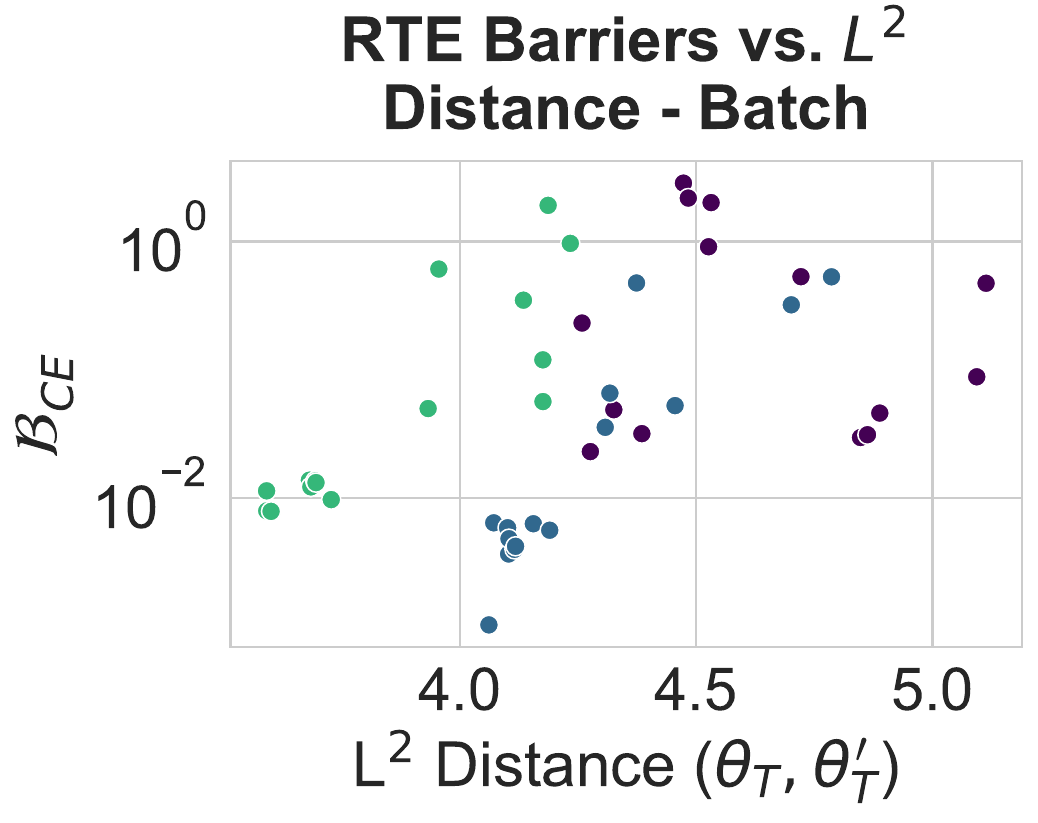}
    \includegraphics[width=0.25\linewidth]{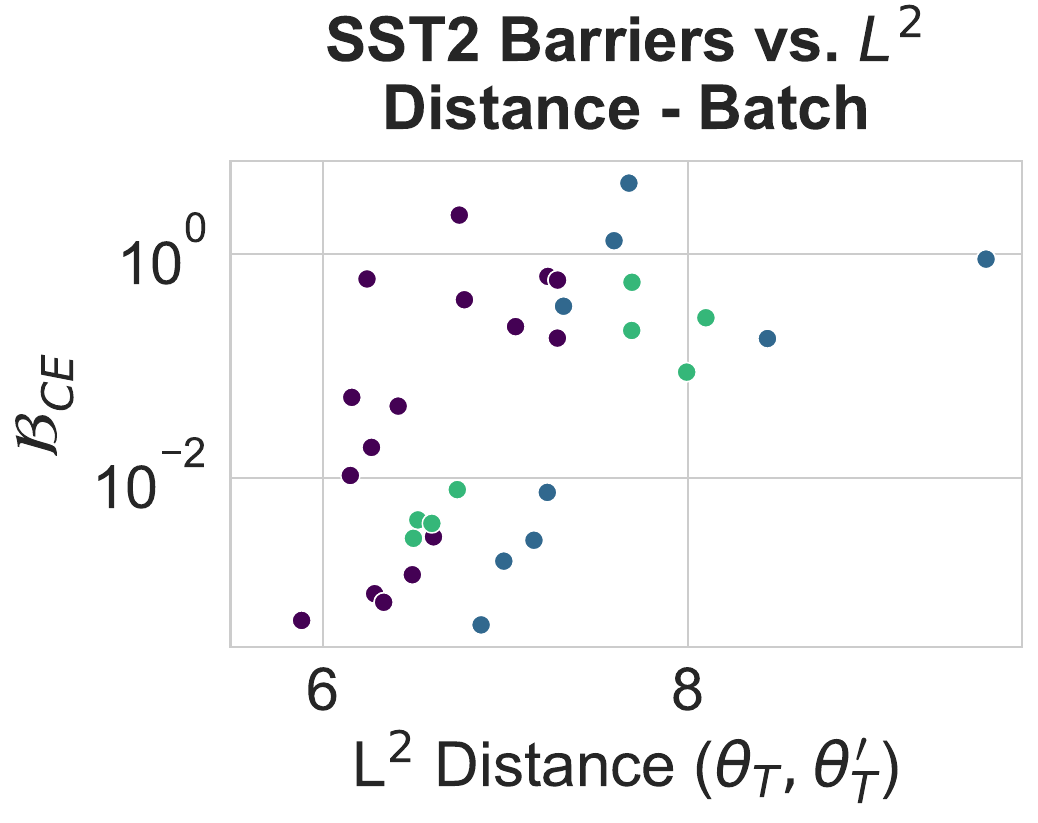}
    \includegraphics[width=0.25\linewidth]{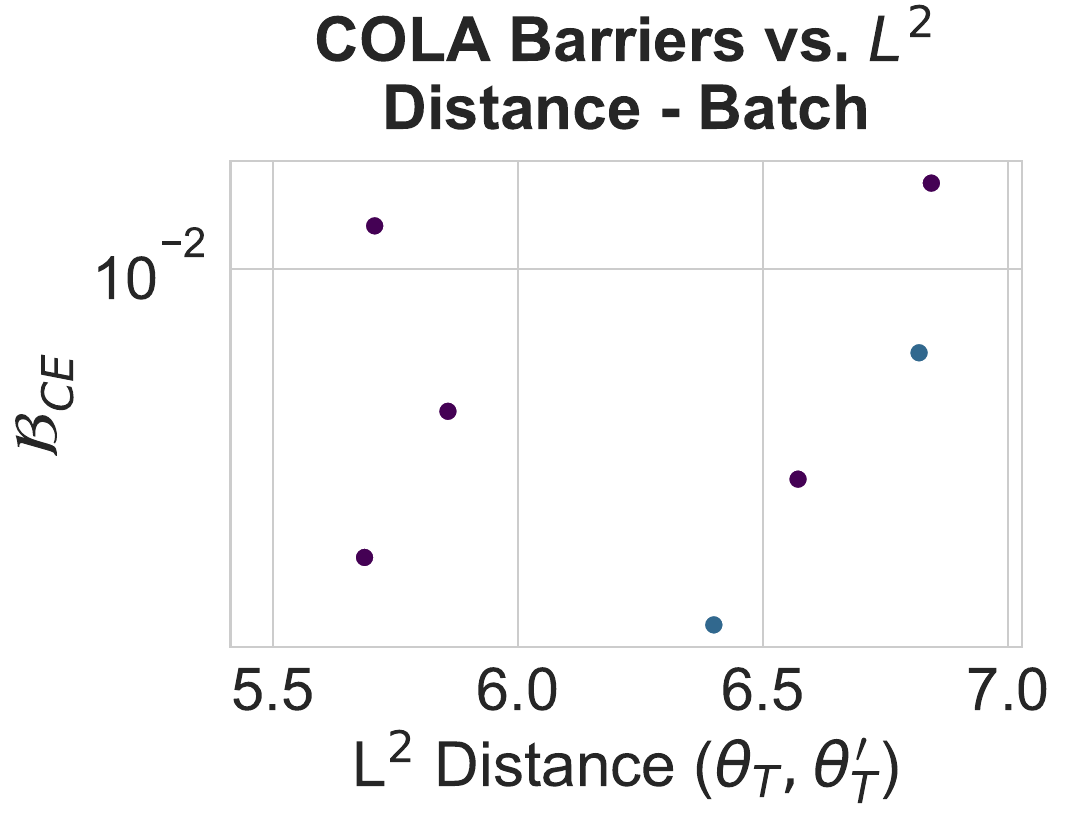}
}
\centerline{
    \includegraphics[width=0.3\columnwidth]{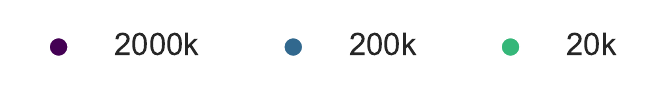}
}
\caption{
Train loss barriers vs. $L^2$ distance between the perturbed and original models at the end of training for MRPC, RTE, SST-2, and COLA.
}
\label{ap:fig:bert-transfer-l2-barr}
\end{center}
\vskip -0.2in
\end{figure}

Here, we examine the relationship between the barriers and the $L^2$ divergence between models at the end of training in greater detail.
\cref{ap:fig:finetune-cifar-l2-barr} (left, middle) shows that fine-tuning on vision tasks, such as transferring from CIFAR-100 to CIFAR-10 and vice versa, follows the trends presented in \cref{fig:l2-barriers} (left).
We see that this direct relationship is weak or non-existent when transferring ViTs from ImageNet to CIFAR-100 (\cref{ap:fig:finetune-cifar-l2-barr} right), as well as for the GLUE benchmark in our study (\cref{ap:fig:bert-transfer-l2-barr}), with QNLI and COLA exhibiting almost no correlation.
Interestingly, OLMo fine-tuned on GSM8K (\cref{fig:olmo-fine-tuning:time}, right) shows a clearer correlation between barriers and $L^2$ divergence, which more closely resembles our ResNet-20 results (\cref{fig:l2-barriers}, left) than BERT.

% L2 barr cifar

% l2 barr bert

\clearpage

\section{Additional Log-Scale Plots}

These plots are copies of main figure plots with log y-axes, and are included to display a clearer separation between smaller barriers.
%%% log scaled plots from main text

\begin{figure*}[htbp]
\vskip 0.1in
\begin{center}
\centerline{
    \includegraphics[height=0.23\linewidth]{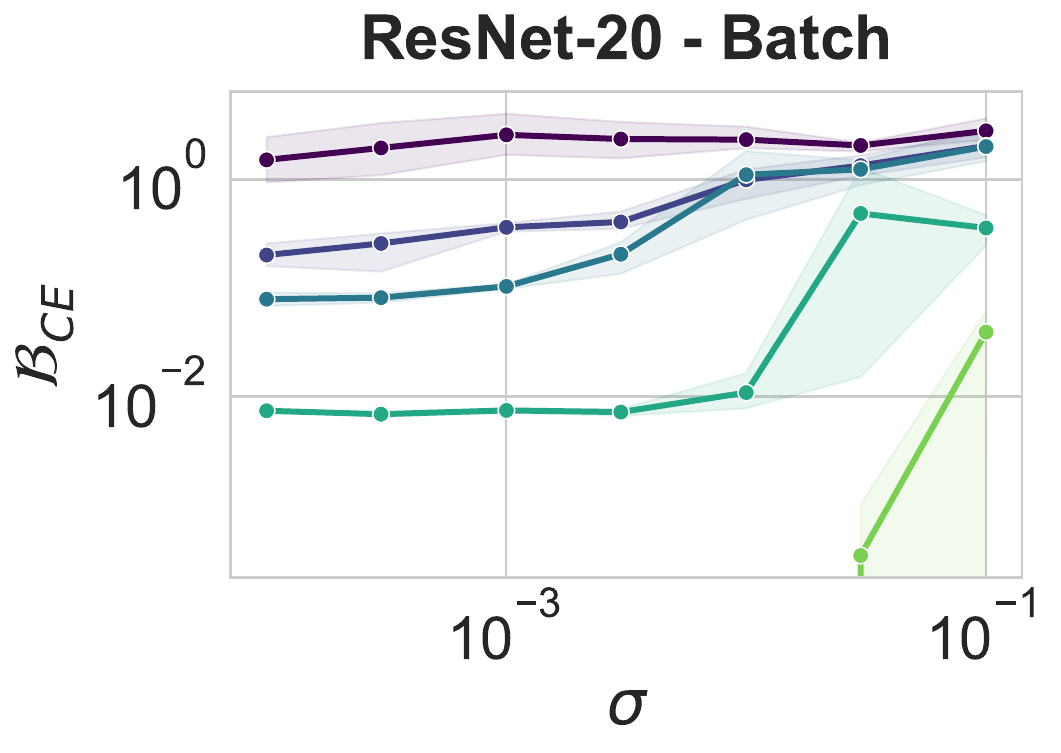}
    \includegraphics[height=0.23\linewidth]{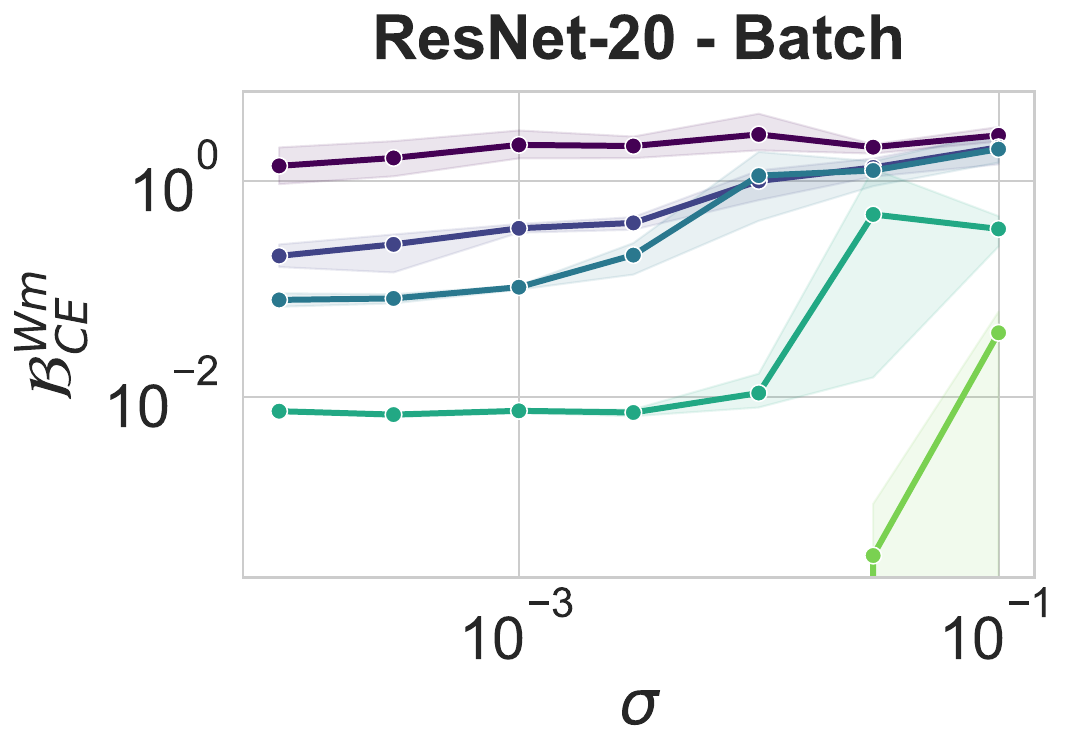}
    \includegraphics[height=0.23\linewidth]{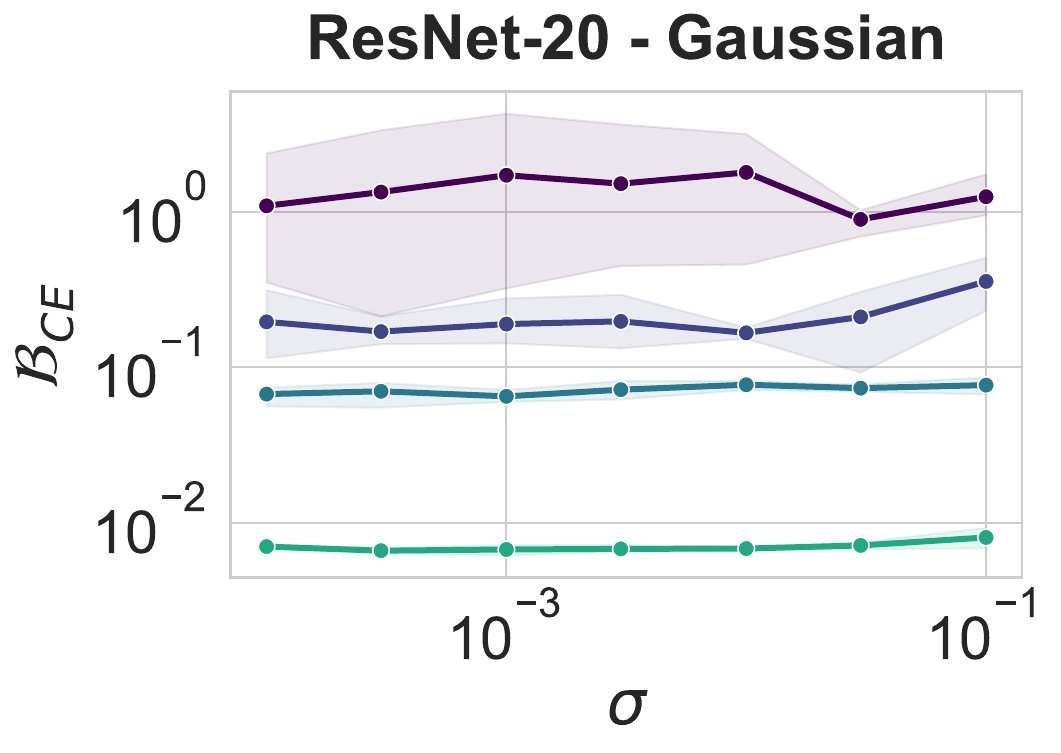}
}
\vskip -0.05in
\centerline{
    \includegraphics[width=0.6\linewidth]{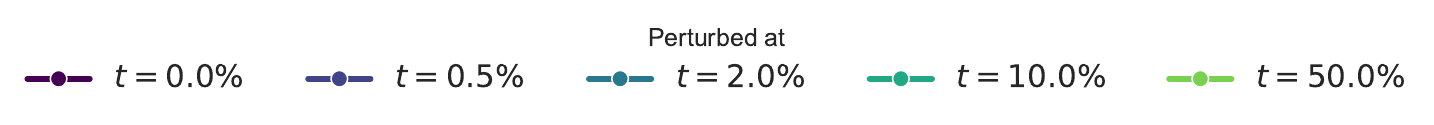}
}
\vskip -0.2in
\caption{
Same data as \cref{fig:butterfly-time} with y-axis in log-scale to improve the readability of smaller barriers.
Loss barriers on training data at the end of training (y-axis) are plotted against perturbation magnitude (x-axis) and perturbation step (color indicates fraction of total training time).  \textbf{Left:} barriers due to batch perturbation. \textbf{Middle:} batch perturbation barriers after accounting for permutations. \textbf{Right:} barriers due to Gaussian perturbation.
}
\label{fig:butterfly-time-log}
\end{center}
\vskip -0.2in
\end{figure*}

%%%% Warmup, shallow, combo

\begin{figure*}[ht]
\vskip 0.1in
\begin{center}
\centerline{
    \includegraphics[height=0.23\linewidth]{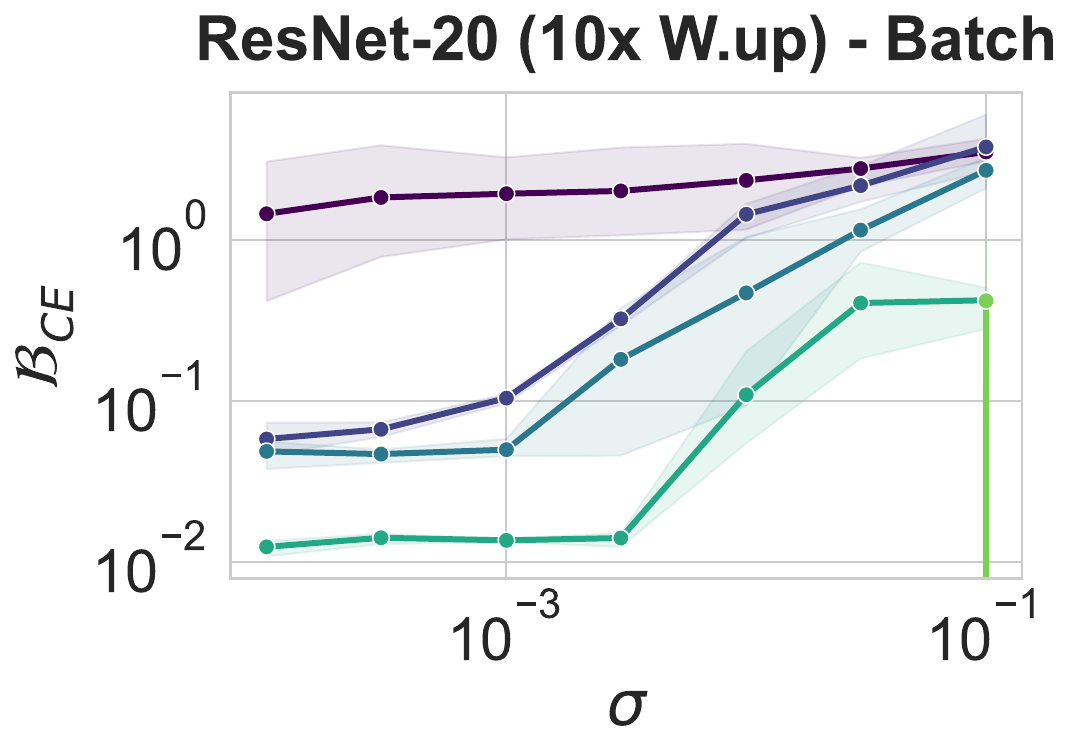}
    \includegraphics[height=0.23\linewidth]{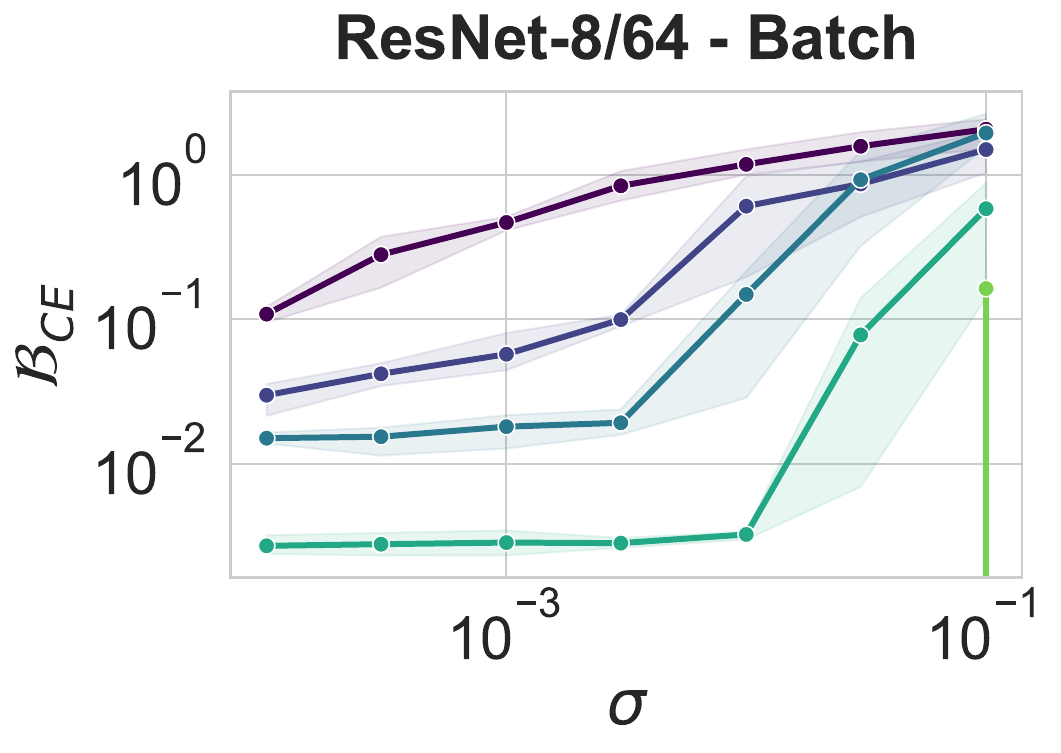}
    \includegraphics[height=0.23\linewidth]{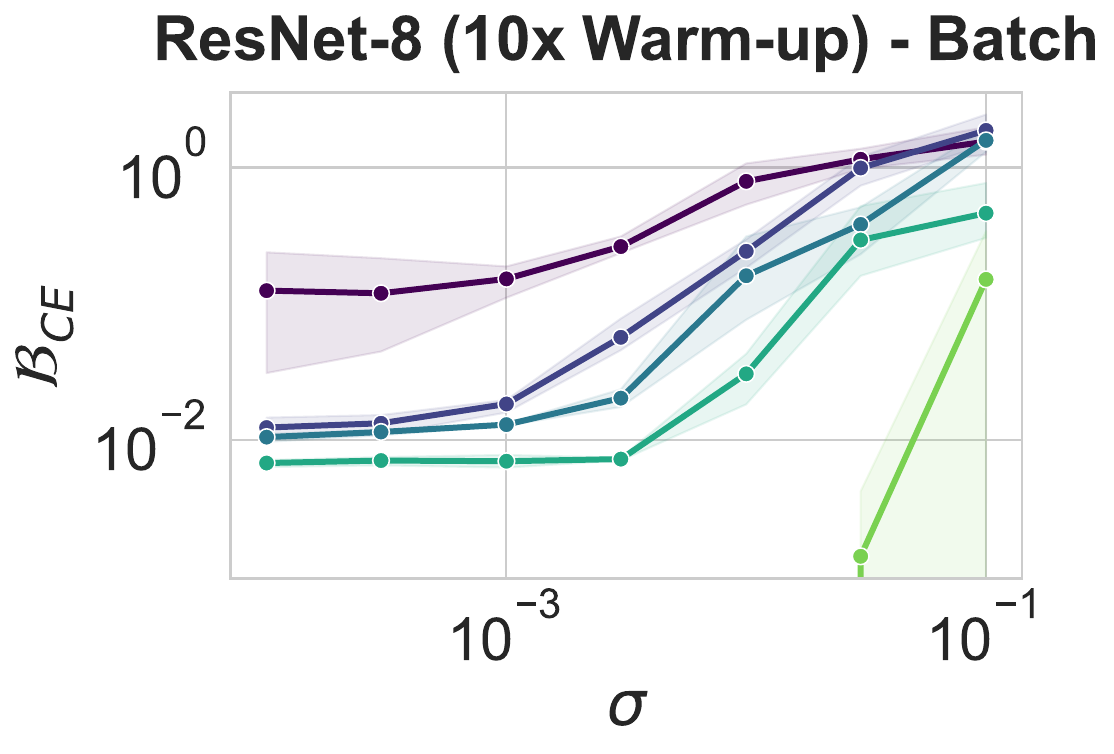}
}
\vskip -0.2in
\centerline{
    \includegraphics[width=0.6\linewidth]{figures/rebuttal/log/no-decay-sanity-batch-lmc-0-1-loss-weighted-barrier-log-legend.pdf}
}
\caption{
Same as \cref{fig:butterfly-warmup-arch-combo} but with y-axis in log-scale for models trained with $20\%$ warm-up time (\textbf{left}), a wider/shallower ResNet8 architecture (\textbf{middle}), and a combination of both settings (\textbf{right}).
}
\label{fig:butterfly-warmup-arch-combo-log}
\end{center}
\vskip -0.2in
\end{figure*}

\end{document}